\documentclass[a4paper,11pt]{article}

\usepackage[margin=1in]{geometry}

\usepackage{newtxtext,newtxmath}

\usepackage[
    colorlinks=true,
    linkcolor=blue,
    anchorcolor=blue,
    citecolor=blue
]{hyperref}

\usepackage[shortlabels]{enumitem}

\usepackage{xcolor}

\usepackage[english]{babel}
\usepackage{natbib}
\setcitestyle{authoryear,open={(},close={)}}

\usepackage{amsthm}

\usepackage{
  mathtools,
  thmtools,
  thm-restate,
  bbm,
  bm,
}

\newtheorem{theorem}{Theorem}[section]
\newtheorem{lemma}[theorem]{Lemma}
\newtheorem{corollary}[theorem]{Corollary}
\newtheorem{definition}[theorem]{Definition}

\newtheorem{algo}[theorem]{Algorithm}

\newenvironment{remark}[1][Remark]
  {
  \begin{proof}[\textnormal{\textbf{#1}}]}
  {\end{proof}}

\newcommand\locallabel[1]{\label{\currentprefix:#1}}
\newcommand\localref[1]{\ref{\currentprefix:#1}}

\newcommand{\rd}{\mathrm{d}}
\newcommand{\R}{\mathbb{R}}

\newcommand{\Id}{\mbf{I}}
\newcommand{\x}{{\mbf{x}}}

\newcommand{\inprod}[2]{\left\langle #1, #2\right\rangle}
\newcommand{\norm}[1]{\left\|#1\right\|}

\newcommand{\braces}[1]{ \left\{ #1 \right\} }

\newcommand{\inv}{^{-1}}
\newcommand{\trans}{^\top}

\newcommand{\indi}{\mathbbm{1}}
\newcommand{\E}{\mathop{\mathbb{E\/}}}

\DeclareMathOperator{\poly}{poly}

\newcommand{\mbf}{\bm}
\newcommand{\mcal}{\mathcal}

\newcommand{\D}{\mcal{D}}

\newcommand{\z}{{\mbf{z}}}
\newcommand{\A}{{\mbf{A}}}
\newcommand{\B}{{\mbf{B}}}

\newcommand{\f}{\mbf{f}}
\newcommand{\W}{\mbf{W}}

\newcommand{\Loss}{\mcal{L}}

\newcommand{\WA}{\W_\A}
\newcommand{\WB}{\W_\B}
\newcommand{\xA}{\x_\A}
\newcommand{\xB}{\x_\B}
\newcommand{\zA}{\z_\A}
\newcommand{\zB}{\z_\B}
\newcommand{\fA}{\f_\A}
\newcommand{\fB}{\f_\B}

\newcommand{\KA}{\mbf{K_A}}
\newcommand{\KB}{\mbf{K_B}}

\newcommand{\K}{\mbf{K}}
\newcommand{\Q}{\mbf{Q}}

\newcommand{\bSigma}{\mbf{\Sigma}}

\newcommand{\Z}{\mbf{Z}}
\newcommand{\bsigma}{\mbf{\sigma}}
\DeclareMathOperator{\diag}{diag}
\newcommand{\bkappa}{{\mbf{\kappa}}}

\newcommand{\KAB}{\K_{\A\B}}
\newcommand{\KBA}{\K_{\B\A}}

\newcommand{\C}{{\mbf{C}}}

\renewcommand{\D}{{\mbf{D}}}

\newcommand{\Term}{\textnormal{\texttt{T}}}

\DeclareMathOperator{\sgn}{sgn}

\newcommand{\ol}{\overline}

\newcommand{\bxi}{\mbf{\xi}}
\newcommand{\Axi}{\A_{\bxi}}
\newcommand{\Bxi}{\B_{\bxi}}
\newcommand{\KAxi}{\K_{\A, \bxi}}
\newcommand{\KBxi}{\K_{\B, \bxi}}

\newcommand{\AB}{\A\B}

\title{On the Importance of Contrastive Loss in Multimodal Learning}
\author{
  Yunwei Ren\thanks{This work was done while Y. Ren was at Carnegie Mellon University.} \\
  Princeton University \\ 
  \texttt{yunwei.ren@princeton.edu}
  \and 
  Yuanzhi Li \\ 
  Carnegie Mellon University \\ 
  \texttt{yuanzhil@andrew.cmu.edu}
}

\begin{document}

\maketitle

\begin{abstract}
  Recently, contrastive learning approaches (e.g., CLIP \cite[]{radford_learning_2021}) have received huge success in 
  multimodal learning, where the model tries to minimize the distance between the representations of different views 
  (e.g., image and its caption) of the same data point while keeping the representations of different data points away 
  from each other. However, from a theoretical perspective, it is unclear how contrastive learning can learn 
  the representations from different views efficiently, especially when the data is \emph{not} isotropic. 
  In this work, we analyze the training dynamics of a simple multimodal contrastive learning model and show that 
  contrastive pairs are important for the model to efficiently balance the learned representations. 
  In particular, we show that the positive pairs will drive the model to align the representations at the cost of
  increasing the condition number, while the negative pairs will reduce the condition number, keeping the learned 
  representations balanced.
\end{abstract}

\section{Introduction}

One of the exceptional abilities of humans is to associate data from different modalities (such as texts and 
images) together. For example, when we hear the words ``white dog'', we can immediately align it with the 
image of a dog with white color. Likewise, when we hear the loud sound of the engine, we can imagine an expensive sports 
car passing nearby.

Recently, in machine learning, multimodal learning methods -- training the model to align the data from different modules, 
has become an increasingly popular research direction, especially in deep learning (\cite{he_fine_grained_2017,%
stroud_learning_2020,radford_learning_2021,ramesh_zero_shot_2021,xu2021simple,jia_scaling_2021,wang2022cliptd}). 
Among them, the recent work CLIP (\cite{radford_learning_2021}) shows remarkable quality results on aligning the 
features of text and images.
The \emph{contrastive learning}-based method CLIP empirically outperforms many existing 
non-contrastive approaches (\cite{grill_bootsrap_2020,chen_exploring_2021,he_momentum_2020}). The main difference 
between the contrastive approach and other approaches is that the contrastive loss not only requires the learned representations 
from the same pair of data (i.e., positive pairs) to be positively aligned but also requires the data from different 
pairs (i.e., negative pairs) to be as negatively aligned as possible. In that paper, the authors also identify 
contrastive loss as the most critical part of CLIP.

Despite the empirical success of this contrastive learning-based method, from a theoretical perspective, the most 
fundamental questions are still largely open: In particular, how do contrastive pairs help in this new multimodal 
learning approach? How can the \textbf{non-convex} contrastive loss be efficiently minimized to learn features from 
both modules?

Unlike prior theoretical works on contrastive learning, which mostly focus on extracting features from one 
module (e.g., \cite{arora_theoretical_2019,jing_understanding_2022,pokle_contrasting_2022,tian_understanding_2021,%
wen_toward_2021}), one main technical challenge of analyzing contrastive learning in a multimodal setting is how the 
model can be trained to align the feature representations $\f_\A, \f_\B$ from modules $\A$ and $\B$ respectively. 
Due to the existence of negative pairs that emphasize negative correlations of $\f_\A$ and $\f_\B$, it is unclear 
that the model still has incentives to align the features from different modules.

In this paper, we make preliminary theoretical steps toward answering the fundamental theoretical questions of the 
importance of contrastive loss in multimodal learning. We assume the data from the two modules are of the form 
$\xA = \A\z_\A + \Axi\bxi_\A$ and $\xB = \B\z_\B + \Bxi\bxi_\B$, respectively, where $\z_\A,\z_\B$ are 
the latent signals, $\A, \B$ linear transformations from the signal to the observation, and $\Axi\bxi_\A, \Bxi\bxi_\B$ the noises. Similar
linear models have also been used in previous works (\cite{tian_understanding_2021,wen_toward_2021}) in the 
context of single-modal learning ($\A = \B$). 
The positive pair of the data share the same signal $\z_\A = \z_\B$ but has different noises $\bxi_\A, \bxi_\B$ and 
transformations $\A, \B$. The goal is to learn features $\f_\A, \f_\B$ that align positive pairs while keeping 
representations of negative pairs away from each other.

Under this setting, we make the following contributions:

\begin{enumerate}
  \item We consider the challenging (but more practical) setting where the features in $\A$ and $\B$ are 
    inhomogeneous, that is, the condition number of $\A$ and $\B$ can be $\omega(1)$. 
    Prior works (\cite{jing_understanding_2022,tian_understanding_2021,wen_toward_2021}) only consider cases where 
    $\A$ and $\B$ are column orthonormal matrices even in the simpler single-modal setting ($\A = \B$).

  \item We consider feature learners $\f_\A, \f_\B$ with normalization, meaning that $\f_\A, \f_\B$ are always 
    normalized to have (expected) norm one during training. Output normalization plays a critical role in the 
    practical success of contrastive learning and is also employed in CLIP, but it is rarely considered in theory 
    due to the additional complexity of the division by the norm. 

  \item We analyze the learning process of stochastic gradient descent from random initialization. We prove that 
    contrastive learning converges efficiently to a nearly optimal solution, which indeed aligns the feature 
    representations $\fA$ and $\fB$.

  \item We also demonstrate the importance of negative pairs by comparing with training only over the positive pairs:
    We prove that although the latter can also learn to align $\fA$ and $\fB$, the features learned by contrastive 
    learning with negative pairs is much more \emph{balanced}, meaning that $\fA, \fB$ can recover all the singular 
    vectors of $\A$ and $\B$ and normalize them. On the other hand, without negative pairs, the learned representation 
    is close to a rank one solution, meaning that $\fA, \fB$ will only focus on the top singular direction of 
    $\A$ and $\B$. 

\item We also perform simulations and more practical experiments to support our theory further. 
        
\end{enumerate}

\section{Related Works}

\paragraph{Multimodal learning}
Despite the empirical success of multimodal learning, there are very few theoretical results on this topic. 
The one most related to ours is \cite{huang_what_2021}, in which the authors show that, in certain cases, 
multimodal methods can provably perform better than single-modal models. However, the authors consider neither 
contrastive pairs nor the training dynamics.

\paragraph{Contrastive/Non-contrastive learning theory}
Another much richer line of research is about contrastive and non-contrastive methods in the context of single-modal 
self-supervised learning. 
Starting from \cite{arora_theoretical_2019}, many recent works have provided various explanations on why 
the representations learned with contrastive learning are useful in downstream tasks (\cite{chuang_debiased_2020,%
tosh_contrastive_2021,nozawa_understanding_2021,wang_chaos_2022,haochen_provable_2021,lee_predicting_2021,%
wang_understanding_2020}). These works mostly focus on the generalization aspect of the problem and do not consider  
training. Among them, \cite{wang_understanding_2020} also study the problem using the notions of 
alignment and uniformity, and demonstrate that balanced representations benefit downstream tasks. However, they do not 
provide guarantees on training. 
Another related line of research is about non-contrastive learning, where the necessity of negative 
examples is questioned. In this line of research, the optimization problem does get considered as
non-contrastive losses have trivial collapsed solutions. \cite{tian_understanding_2021} show that, 
under certain conditions, non-contrastive learning methods can learn non-collapsed solutions. 
\cite{jing_understanding_2022} show that, even with negative examples, contrastive learning can still suffer 
from another type of collapse, where the learned representations only expand a low-dimensional subspace of 
the embedding space. In \cite{pokle_contrasting_2022}, the authors show that non-contrastive losses have 
many non-collapsed bad minima which the training algorithm does not avoid. Another related work that considers optimization is \cite{wen_toward_2021}, in which the authors analyze the training dynamics 
of contrastive learning and show that, with random augmentation, neural networks can learn features that are 
suppressed by noises when no augmentations are used. Though these works do consider the optimization problem, 
they focus on the case where the features are uniform, and only \cite{wen_toward_2021} consider output 
normalization. We compare our results with the most relevant works in the next paragraph.

\paragraph{Comparison with \cite{jing_understanding_2022,tian_understanding_2021,wen_toward_2021}}
The dimensional collapse problem reported in \cite{jing_understanding_2022} is not a real issue in our 
setting since, in our case, the best the model can do is to recover the latent vector $\z$, up to some 
rotation. As a result, it is natural for the learned representations to span only a low-dimensional subspace 
of the embedding space $\R^m$. Here, the point of choosing $m \gg d$ is to make the optimization dynamics more 
regular, which is a common strategy in the study of over-parameterized neural networks. The main difference 
between our work and the analysis in \cite{tian_understanding_2021} and \cite{wen_toward_2021} is we do not 
assume the inputs are isotropic. The condition number can be as large as $\Theta(\log d)$ in our setting. 
When the condition number is $1$, one can imagine that thanks to the symmetry, all directions will be learned 
simultaneously. Therefore, we do not need negative examples to prevent collapse (\cite{tian_understanding_2021})
or the negative examples do not play an important role in analysis (\cite{wen_toward_2021}). On the other hand,
when the condition number is larger than $1$, we do need to use the negative examples to shrink the condition 
number, corresponding to the second stage of our analysis.

\section{Problem Setup}

Similar to previous theoretical works (e.g., \cite{lee_predicting_2021,wen_toward_2021}), we consider a linear 
data-generating model. Formally, we assume that the contrastive pairs $(\xA^+, \xB^-)$ are constructed as 
\begin{equation}
  \label{eq: main text: data-generating model}
  \xA^+ = \A\z^+ + \Axi\bxi_\A^+, \quad 
  \xB^- = \B\z^- + \Bxi\bxi_\B^-,  
\end{equation}
where $\z^\pm$, $\bxi_\A^-$, $\bxi_\B^-$ are independent random variables following the uniforms distributions 
over $\{ \pm 1 / \sqrt{d} \}^r$, $\{ \pm 1 / \sqrt{d} \}^{d-r}$ and $\{ \pm 1 / \sqrt{d} \}^{d-r}$, respectively,
and $\A, \B \in \R^{d \times r}$, $\Axi, \Bxi \in \R^{d \times (d - r)}$
are matrices with $\A\trans\A = \B\trans\B = \diag(\bsigma^2)$ and $\Axi\trans\Axi = \Bxi\trans\Bxi = 
\sigma_{\xi}^2 \Id_{d-r}$ for some $\bsigma \in \R_+^r$ and $\sigma_{\xi} \in \R_+$. In words, we first 
sample the latent vectors $\z^\pm \in \R^r$ independently and then encode them with $\A, \B$ to form the 
signal part of the input. After that, we sample the latent noises $\bxi_\A^+, \bxi_\B^- \in \R^{d-r}$, and encode 
them with $\Axi, \Bxi$ to form the noise part of the input. Finally, we combine the signal and noise parts 
to obtain $(\xA^+, \xB^-)$. We use the same latent vector to generate a positive pair $(\xA^+, \xB^+)$. That is,
$\xA^+ = \A\z^+ + \Axi\bxi_\A^+$ and $\xB^+ = \B\z^+ + \Bxi\bxi_\B^+$. Note that the latent noises here are 
still independent. Let $\sigma_{\max}^2$ and $\sigma_{\min}^2$ denote the maximum and minimum of 
$\sigma_1^2, \dots, \sigma_r^2, \sigma_\xi^2$, respectively. We assume that $(\sigma_{\max}^2/\sigma_{\min}^2)
\max\braces{ 1, (d - r) \sigma_\xi^2 / (r \sigma_{\min}^2) } \le c \log d$ for some small constant $c > 0$. 
Our results can be easily generalized to settings where the dimensions of $\xA$ and $\xB$ are different
since one can simply pad zeros at the end of each column of $\A$ and $\B$. 

One way of interpreting this model is to view each coordinate of the latent vector $\z$ as an indicator for the 
presence/absence of a certain object, and the corresponding columns in $\A$ and $\B$ as the visual and word 
embeddings of this object, respectively.

Now, we describe our learner model. We consider (normalized) linear feature learners. Define 
\[
  \f_\A(\xA)
  := \frac{\WA\trans\xA}{\sqrt{ \E_{\xA} \norm{\WA\trans\xA}^2 }},  \quad 
  \f_\B(\xB)
  := \frac{\WB\trans\xB}{\sqrt{ \E_{\xB} \norm{\WB\trans\xA}^2 }}, 
\]
where $\WA, \WB \in \R^{d \times m}$ are the trainable parameters. In words, we first map the inputs $(\xA, \xB)$
into the embedding space $\R^m$ using $\WA$ and $\WB$, and then apply batch normalization to the outputs. By 
saying the learned representations are aligned, we mean that $\fA$ and $\fB$ are close for positive pairs and far away 
from each other for negative pairs. Meanwhile, we say the learned representations are balanced if changing a small 
fraction of coordinates of $\z$ does not change the representation dramatically. See Section~\ref{sec: main results}
for formal definitions. 

One can easily verify that, in the population case, we have $\E_{\xA} \norm{\WA\trans\xA}^2 = \norm{\WA\trans\A}_F^2/d
+ \norm{\WA\trans\Axi}_F^2/d$. For notational simplicity, we write $\KA = \WA\trans\A$, $\KB = \WB\trans\B$, 
$\KAxi = \WA\trans\A_{\bxi}$, and $\KBxi = \WB\trans\B_{\bxi}$. These are the matrices that directly map latent vectors 
to their final representations. We also define $N_\A^2 = (\norm{\KA}_F^2 + \norm{\KAxi}_F^2) / d$ and
$N_\B^2 = (\norm{\KB}_F^2 + \norm{\KBxi}_F^2) / d$. Then our model can be rewritten as\footnote{See 
Section~\ref{sec: gf to gd} for discussions on the sample complexity.} 
\begin{equation}
  \label{eq: learner model}
  \f_\A(\xA) = \frac{\KA\z_\A + \KAxi\bxi_\A }{ N_\A }, \quad 
  \f_\B(\xB) = \frac{\KB\z_\B + \KBxi\bxi_\B }{ N_\B }. 
\end{equation}

We initialize each entry of $\WA$ and $\WB$ using iid Gaussian $\mcal{N}(0, 1/m)$. This scaling ensures the norm of 
outputs before normalization does not blow up as $m \to \infty$. 
We train our model using gradient descent over the following contrastive loss $\Loss$. First, we 
define\footnote{We use $\fA^+$ as a shorthand for $\f_\A(\xA^+)$, similarly for other combinations of $\A, \B$ and 
$\pm$.}
\begin{align*}
  S_\A(\xA^+, \xB^+)
  &= \frac{ \exp(\tau_t^2 \fA^+ \cdot \fB^+ ) }
    {\exp(\tau_t^2 \fA^+ \cdot \fB^+ ) + K \E_{\xB^-} \exp(\tau_t^2 \fA^+ \cdot \fB^- ) }, \\
  S_\B(\xA^+, \xB^+)
  &= \frac{ \exp(\tau_t^2 \fA^+ \cdot \fB^+ ) }
    {\exp(\tau_t^2 \fA^+ \cdot \fB^+ ) + K \E_{\xA^-} \exp(\tau_t^2 \fA^- \cdot \fB^+ ) },
\end{align*}
where $K$ is a positive constant controlling the strength of negative samples and $\tau_t \in (0, 1]$ is 
the inverse temperature. We use a time-dependent inverse temperature to separate the aligning and balancing phases of 
the training process. In practice, these two phases can interlace. 
We define the contrastive loss as 
\begin{equation}
  \label{eq: contrastive loss}
  \Loss 
  := \Loss_\A + \Loss_\B 
  := - \E \log S_\A(\xA^+, \xB^+) - \E \log S_\B(\xA^+, \xB^+).
\end{equation}
By the non-contrastive loss, we mean 
\begin{equation}
  \label{eq: non-constrative loss}
  \hat\Loss 
  := - \E \inprod{\fA^+}{\fB^+}. 
\end{equation}
Formally, the training algorithm is defined as follows. 
\begin{algo}[Training algorithm]
  \label{algo: training algorithm}
  Let $\tilde\Loss$ be $\Loss$ in the contrastive case and $\hat\Loss$ in the non-contrastive case. 
  At each step, we first sample a batch of positive/negative pairs $\{ (\xA^+, \xB^+, \xA^-, \xB^-) \}_{i=1}^N$, use 
  them to compute the empirical version of $\tilde\Loss$, and update the weight matrices using $\WA \gets 
  \WA - \tau_t^{-2} \eta \nabla_{\WA} \tilde\Loss $ and $\WB \gets \WB - \tau_t^{-2} \eta \nabla_{\WB} \tilde\Loss $. 

  In the non-contrastive case, we always use $\tau_t = 1$\footnote{Note that, in the non-contrastive case, 
  changing $\tau_t$ only changes the learning rate.}, 
  and we repeat the above update until gradient descent converges to an approximate 
  stationary point. In the contrastive case, we first use a small $\tau_t = 1 / \poly(d)$, run the process for 
  $T_1 = \poly(d)$ iterations, switch to $\tau_t = 1$, and run the process for another 
  $T_2 - T_1 = \poly(d)$ iterations. 
\end{algo}

\section{Main Results}
\label{sec: main results}

\newcommand{\Alignment}{\Gamma_{\textnormal{\texttt{Align}}}}
\newcommand{\Balance}{\Gamma_{\textnormal{\texttt{Balance}}}}

In words, our results say that though both contrastive and non-contrastive methods can align the representations, the 
representations learned by contrastive methods are more balanced. First, we define the meaning of alignment and balance 
as follows. Though our theoretical analysis focuses on the normalized linear case, where alignment and balance can be
measured using the distance $\norm{\KA/N_\A - \KB/N_\B}$ and the condition number of $\KA$ (or $\KB$), the following two 
definitions are architecture-agnostic.

\begin{definition}[Alignment]
  \label{defi: alignment}
  We define the alignment score as 
  \[
    \Alignment
    := \frac{1}{2} \E_{\xA^\pm, \xB^\pm} \braces{
        \indi\braces{ \norm{\fA^+ - \fB^+} < \norm{\fA^+ - \fB^-} }
        + \indi\braces{ \norm{\fA^+ - \fB^+} < \norm{\fA^- - \fB^+} }
      }. 
  \]
  Namely, $\Alignment$ is the accuracy of classifying whether the input pair $(\xA, \xB)$ is a positive pair. 
  We say that the learned representations are aligned if $\Alignment \approx 1$. 
\end{definition}

Note that the notion of alignment introduced here is stronger than matching the positive pairs, which can be achieved 
by simply mapping all inputs to one single embedding. In that case, $\Alignment$ will be $0$ (or $0.5$ if we 
choose to break ties randomly instead of using strict inequality).

\begin{definition}[Balance]
  \label{defi: balance}
  We define the balance score as 
  \[
    \Balance
    := \norm{\bSigma_f}_F^2 / \norm{\bSigma}_2^2 
    \quad\text{where}\quad
    \bSigma_f 
    := \E_{\xA} \braces{ \fA \fA\trans } . 
  \]
  Namely, $\Balance$ is the stable rank of the covariance matrix of the output embeddings. We say that 
  the learned representations are balanced if $\Balance \ge \alpha r$ for some $\alpha \approx 1$. 
\end{definition}

We use only $\fA$ in this definition because if the representations are well-aligned, $\fA\fA\trans$ and $\fB\fB\trans$ 
should be approximately the same.
The intuition behind the use of (stable) rank is that the more independent latent variables the model learns, 
the higher the rank of the representations needs to be. In \cite{jing_understanding_2022}, the authors also use rank 
to measure the degree of ``dimensional collapse''.
Unlike their argument, here we only require $\Balance$ to be at least $\alpha r$, instead of $\alpha m$, for the 
representations to be called balanced because even if we can recover the underlying latent vectors, the rank is still 
at most $r$. Hence, it does not make much sense to expect the learned representations to span the entire embedding 
space. Finally, note that a sufficient condition for $\Balance \ge \alpha r$ is that the ratio of the largest and $r$-largest 
singular values is at most $\sqrt{\alpha}$. In other words, after excluding those singular values that should be 
$0$, the condition number is approximately $1$. 

\paragraph{Why balance representations are important?} In our setting, one simple example of aligned but unbalanced representations is $\WA\trans = \diag(\mbf\nu)\A\inv$
and $\WB\trans = \diag(\mbf\nu)\B\inv$ with $\nu_1 = 1$ and $\nu_2 = \cdots = \nu_r = 1/\poly(d)$. This model maps 
inputs whose latent vector is $\z$ to $\diag(\mbf\mu)\z$, up to some normalization, for both modules, whence it 
has $\Alignment = 1$. Meanwhile, one can verify that this model has $\Balance / r \le 2/r \approx 0$. The problem of 
this model is that it overly emphasizes $z_1$ and is sensitive to small changes in $z_1$. This model can be made balanced 
by replacing $\diag(\mbf\mu)$ with $\Id_r$, in which case the model directly recovers the latent vector $\z$. 
As a result, all changes in $\z$ will be reflected in the final representation in a faithful way. 
Similar notions have also been studied in \cite{wang_understanding_2020} under the name ``uniformity'' in the 
context of self-supervised learning, and they also report that balanced representations lead to better performance
in downstream tasks, though, unlike our result, they do not provide guarantees on training. Still, this further
suggests that learning balanced representations is a reasonable and important goal. 

With these two definitions, we can now state our main results.  

\begin{theorem}
  \label{thm: main}
  Suppose that the network width $m = \poly(d)$ is sufficiently large, the learning rate $\eta = 1 / \poly(d)$
  is sufficiently small, and we generate sufficiently, but still polynomially, many samples at each step to 
  compute the loss\footnote{To make the proof cleaner, we write it in terms of gradient flow over population loss.
  See Section~\ref{sec: gf to gd} for discussions on the discretization of gradient flow. }.
  Suppose that, for some small constant $c > 0$,
  \[ 
    \frac{\sigma_{\max}^2}{\sigma_{\min}^2}
    \max\braces{ 1, \frac{ (d - r) \sigma_\xi^2 }{ r \sigma_{\min}^2 } } \le c \log d. 
  \] 
  \begin{enumerate}[(a)]
    \item \textbf{Non-contrastive loss.}
      There exists a $\bsigma \in \R^r$ satisfying the above assumptions such that, after $\poly(d)$ iterations, 
      Algorithm~\ref{algo: training algorithm} will converge to an approximate stationary point, at which the 
      learned representations are aligned but not balanced, that is, $\Alignment \approx 1$ but 
      $\Balance / r \approx 0$. 
    
    \item \textbf{Contrastive loss.}
      There exists $\tau_0^2 = 1 / \poly(d)$ and $T_1 = \poly(d)$ such that, for any valid $\bsigma$, 
      Algorithm~\ref{algo: training algorithm} will reach a point after $\poly(d)$ iterations at which the learned 
      representations are aligned and balanced, that is, $\Alignment \approx 1$ and $\Balance / r \approx 1$. 
  \end{enumerate}
\end{theorem}

We close this section with sufficient conditions for the learned representations to be aligned and balanced in our 
setting. 
First, in our specific setting, the most natural way for a model to achieve $\Alignment \approx 1$ is to learn 
$\norm{ \fA - \fB } \approx \norm{ \K_0 (\zA - \zB)  }$ for some $\K_0 \in \R^{m \times r}$ with 
$\K_0\trans\K_0 \succ 0$. In this case, the distance between positive pairs is always close to $0$ while the distance 
between negative pairs is positive. Recall that the output of our model is $\fA = (\KA\z_\A + \KAxi \bxi_\A) 
/ N_\A$ and $\fB = (\KB\z_\B + \KBxi \bxi_\B) / N_\B$. Hence, we have the following sufficient condition. 

\begin{lemma}[Sufficent condition for aligned representations]
  \label{lemma: sufficient condition for aligned representations}
  If $\norm{\KA}_F \gg \norm{\KAxi}_F$, $\norm{\KB}_F \gg \norm{\KBxi}_F$, $\KA \approx \KB$, and the condition number 
  of $\KA$ is upper bounded by $\poly(d)$, then the learned representations are aligned. 
\end{lemma}
The first two conditions imply the signal parts dominate the noise parts, the third condition gives 
the existence of $\K_0$, and the final condition makes sure the smallest singular value of $\K_0\trans\K_0$ 
is still at least $1 / \poly(d)$ after normalization. 

Now we consider the balance of the learned representations. Suppose that our representations are already 
aligned, in the sense of Lemma~\ref{lemma: sufficient condition for aligned representations}. Then we have 
\[
  \bSigma_f 
  = \E_{\xA} \braces{ \fA \fA\trans }
  = \E_{\z} \braces{ \frac{ \KA\z\z\trans\KA\trans }{ \norm{\KA}_F^2 / d }  }
  = \frac{ \KA\KA\trans }{ \norm{\KA}_F^2 }. 
\]
Recall the relationship between $\Balance$ and the effective condition number of $\bSigma_f$. We have the 
following sufficient condition. 

\begin{lemma}[Sufficient condition for balanced representations]
  \label{lemma: sufficient condition for balanced representations}
  Suppose that our model is aligned, in the sense of Lemma~\ref{lemma: sufficient condition for aligned 
  representations}. If the condition number of $\KA\trans\KA$ is close to $1$, then the learned representations
  are balanced. 
\end{lemma}

Note that, since the columns of $\A$ are not orthonormal, even at initialization, the condition number of 
$\KA\trans\KA$ is not close to $1$. In other words, a condition-number-reducing stage is necessary for the model to 
learn balanced representations.

\section{Training Dynamics and Proof Outline}

\begin{figure}[ht]
  \centering
  \includegraphics[width=0.95\textwidth]{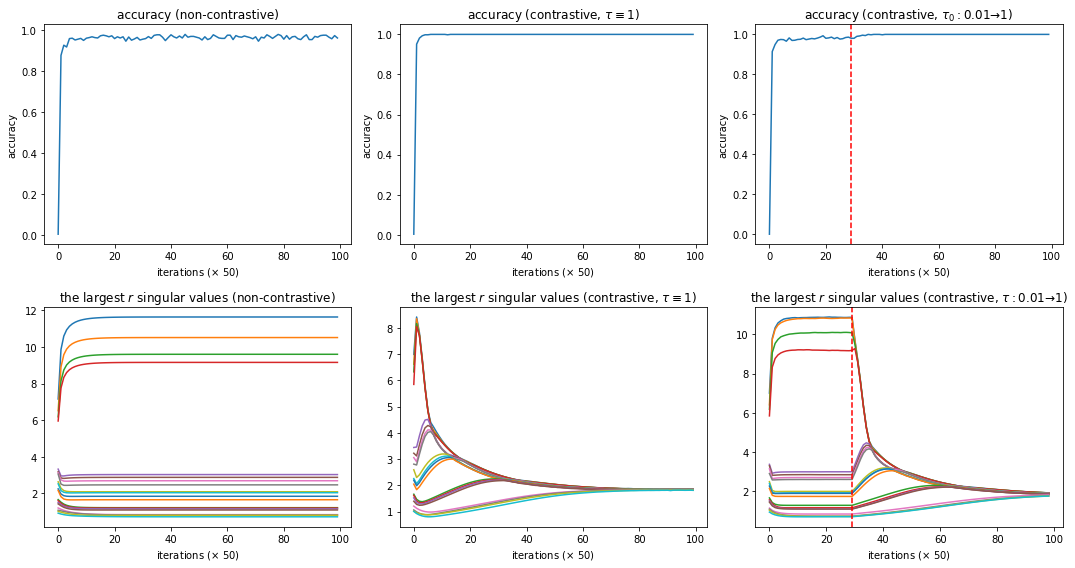}
  \caption{Simulation results.
    The first row reports the accuracies of different approaches on the classifying positive/negative pairs problem,
    and the second row reports the values of the largest $r$ singular values. From left to right, the columns 
    correspond to the non-contrastive method, the contrastive method with $\tau_t \equiv 1$ throughout the entire 
    process, and the contrastive method with $\tau_t$ switch from $0.01$ to $1$ at the vertical dashed line. One can 
    make several observations here. (a) All methods can quickly attain near $100\%$ accuracy. (b) Only contrastive 
    methods will reduce the condition number to approximately $1$. (c) Even when $\tau_t \equiv 1$, we still 
    have the stage-wise behavior, where the models align the representations in Stage~1 and then balance 
    the representations in Stage~2. 
  }
  \label{fig: simulation}
\end{figure}

Our proof is based on characterizing the dynamics of gradient descent over the contrastive loss. We first choose 
a small inverse temperature $\tau_t^2 = 1 / \poly(d)$ and run gradient descent for $\poly(d)$ iterations. This 
stage is called Stage~1. Then, we set $\tau_t^2 = 1$ and run gradient descent for another $\poly(d)$ iterations.
This stage is called Stage~2. Using different $\tau_t$ is mainly for technical purposes since this gives 
cleaner separation between stages. In our setting, similar stage-wise behavior can still be observed when using a uniform 
$\tau_t^2$. In practice, these two stages can interlace. See Figure~\ref{fig: simulation} for simulation results and 
Section~\ref{sec: main text: experiments} for empirical results. 

\subsection{The training dynamics}

Instead of tracking the parameters $\WA$ and $\WB$ directly, we will track $\KA, \KAxi, \KB, \KBxi$, the matrices 
that directly map latent signals and noises to the final representations. For the case 
with contrastive pairs, one can show that the dynamics of $\KA$ are governed by the following equation
\begin{equation}
  \label{eq: main text: d KA}
  \begin{aligned}
    \dot{\K}_{\A}
    &= \E_{\xA^+, \xB^+} \braces{ 
        \left( 2 - S_\A(\xA^+, \xB^+) - S_\B(\xA^+, \xB^+) \right)
        \left(
          \frac{ \fB^+(\z^+)\trans }{N_\A}
          - \inprod{\fA^+}{\fB^+}\frac{\KA}{ N_\A^2 d}
        \right)
      } \diag(\bsigma^2)
      \\
      &\quad
      - K \E_{\xA^+, \xB^\pm} \braces{ 
        \frac{ S_\A(\xA^+, \xB^+) \exp(\tau_t^2 \fA^+ \cdot \fB^-)}{\exp(\tau_t^2 \fA^+ \cdot \fB^+)}
        \left(
          \frac{ \fB^-(\z^+)\trans }{N_\A}
          - \inprod{\fA^+}{\fB^-}  \frac{\KA}{ N_\A^2 d }
        \right)
      } \diag(\bsigma^2) \\
      &\quad
      - K \E_{\xA^\pm, \xB^+} \braces{ 
        \frac{S_\B(\xA^+, \xB^+) \exp(\tau_t^2 \fA^- \cdot \fB^+)}{\exp(\tau_t^2 \fA^+ \cdot \fB^+)}
          \left(
            \frac{ \fB^+(\z^-)\trans }{N_\A}
            - \inprod{\fA^-}{\fB^+} \frac{ \KA }{ N_\A^2 d }
          \right)
      } \diag(\bsigma^2).
  \end{aligned}
\end{equation}
See Lemma~\ref{lemma: d KA, d KAxi} for the calculation. Intuitively, the first term comes from the positive pairs,
and the second and third terms are from the negative pairs. Within each term, the second part (the one containing 
$\inprod{\fA}{\fB}$) comes from the normalization.
The equations for the other $\K$-matrices can be derived similarly. We rescale the gradients by $1/\tau_t^2$ so that 
$\frac{\rd}{\rd t} \KA$ does not shrink with $\tau_t^2$. For the non-contrastive case, 
the equation is 
\begin{equation}
  \label{eq: main text: d KA (without contrastive pairs)}
  \frac{\rd}{\rd t} \KA 
  = \E_{\xA^+, \xB^+} \braces{ 
      \frac{ \fB^+(\z^+)\trans }{N_\A}
      - \inprod{\fA^+}{\fB^+}\frac{\KA}{ N_\A^2 d}
    } \diag(\bsigma^2). 
\end{equation}
Note that the RHS resembles the first term of \eqref{eq: main text: d KA}. This is not a coincidence. We will 
later establish the approximate equivalence between the non-contrastive approach and the contrastive approach with a 
small inverse temperature $\tau_t^2$ (cf.~Section~\ref{sec: main text: stage 1} and Lemma~\ref{lemma: non-contrastive 
equiv stage 1}).

\subsection{The infinite-width dynamics}

The overall proof strategy is to first characterize the dynamics of the infinite-width limit, which is much simpler 
compared to \eqref{eq: main text: d KA}, and then control the error introduced by discretizing the infinite-width 
network using polynomially many neurons. This discretization is one of the main technical challenges of the proof. 
In general, in order to track the infinite-width dynamics, an exponentially large network is needed 
(\cite{mei_mean_2018}).

The basic idea behind a $\poly(d)$-width discretization is to Taylor expand the dynamics around the infinite-width trajectory to factor out the first-order error 
terms and show that, either they drive the process toward the infinite-width trajectory or the error growth 
introduced by them is much slower than the convergence rate. 

Here, for ease of presentation, we will focus on the noiseless infinite-width dynamics and, in particular, the 
evolution of the condition number. In the appendix, we do handle the noisy finite-width case rigorously. 
First, recall that we use iid Gaussians to initialize the entries of $\WA$ and $\WB$. 
Hence, in the $m \to \infty$ limit, different columns of $\KA$ and $\KB$ are orthogonal to each other, at least at 
initialization. To see this, note that for any $p, q \in [r]$, 
\begin{align*}
  \inprod{[\KA]_p}{[\KB]_q} 
  = \sum_{k=1}^m [\WA\trans\A]_{k, p} [\WB\trans\B]_{k, q}
  &= \sum_{k=1}^m \sum_{i=1}^d [\WA]_{i, k} A_{i, p} \sum_{j=1}^d [\WB]_{j, k} B_{j, q} \\
  &= \sum_{i, j \in [d]} A_{i, p} B_{j, q} \sum_{k=1}^m [\WA]_{i, k} [\WB]_{j, k}. 
\end{align*}
Since the entries of $\WA$ and $\WB$ are initialized with iid $\mcal{N}(0, 1/m)$, we have, as $m \to \infty$, 
\[
  \sum_{k=1}^m [\WA]_{i, k} [\WB]_{j, k}
  \to \E_{w_A, w_B \sim \mcal{N}(0, 1)} [w_A w_B] 
  = 0. 
\]
Similarly, at initialization and $m \to \infty$, for any $p \ne q$, we have
\begin{align*}
  \inprod{[\KA]_p}{[\KA]_q} \to 0, \qquad
  \norm{[\KA]_p}^2  
  \to \sum_{i, j \in [d]} A_{i, p} A_{j, p} \E_{w \sim \mcal{N}(0, 1)} w^2 
  = \norm{[\A]_p}^2 = \sigma_p^2, 
\end{align*}
and similarly for $\inprod{[\KB]_p}{[\KB]_q}$ and $\norm{[\KB]_p}^2$. In other words, at $t = 0$ and $m \to \infty$, we have 
\begin{equation}
  \label{eq: infinite-width, orthogonality}
  \KA\trans\KA = \KB\trans\KB = \diag(\bkappa^2) 
  \quad\text{and}\quad 
  \KA\trans\KB = \diag(\hat\bkappa^2), 
\end{equation}
for some $\bkappa, \hat\bkappa \in \R^r$. Note that the dynamics depend only on the norms and the inner products 
$\inprod{\fA}{\fB}$. As a result, at least at initialization, it suffices to consider $\bkappa$ and $\hat\bkappa$. Moreover, 
one can show that thanks to the symmetry, as long as \eqref{eq: infinite-width, orthogonality} holds at initialization, it 
will remain true throughout the entire training procedure. This implies that, in order to characterize the (infinite-width) 
dynamics of $\KA, \KB$, it suffices to look at $\bkappa^2$ and $\hat\bkappa^2$. 
One can show that, in this noiseless infinite-width limit, the dynamics of $\bkappa^2$ and $\hat\bkappa^2$
are given by (cf.~Lemma~\ref{lemma: infinte-width: d kappa}) 
\begin{equation}
  \label{eq: main text: infinite-width dyanmics}
  \begin{aligned}
    \frac{\rd}{\rd t} \kappa_p^2 
    &= 4 \left( 1 - \tilde{S} \right)
      \left(
        \frac{\hat\kappa_p^2}{\norm{\bkappa}^2 }
        - \frac{ \norm{\hat\bkappa}^2 }{\norm{\bkappa}^2}
          \frac{\kappa_p^2}{ \norm{\bkappa}^2 }
      \right) 
      \sigma_p^2 
      - 4 \left( 1 - \tilde{S}  \right)
      \left(
        \frac{ \hat\kappa_p^2  }{\norm{\bkappa}^2} T_p
        - \frac{ \kappa_p^2 }{\norm{\bkappa}^2} \tilde{T}
      \right)
      \sigma_p^2, \\
    \frac{\rd}{\rd t} \hat\kappa_p^2
    &= 4 \left( 1 - \tilde{S} \right)
      \left(
        \frac{\kappa_p^2}{\norm{\bkappa}^2 }
        - \frac{ \norm{\hat\bkappa}^2 }{\norm{\bkappa}^2}
          \frac{\hat\kappa_p^2}{ \norm{\bkappa}^2 }
      \right) 
      \sigma_p^2 
      - 4 \left( 1 - \tilde{S}  \right)
      \left(
        \frac{\kappa_p^2}{\norm{\bkappa}^2} T_p
        - \frac{\hat\kappa_p^2}{\norm{\bkappa}^2} \tilde{T}
      \right)
      \sigma_p^2, 
  \end{aligned}
\end{equation}
where $\tilde{S}$ is a $\Theta(1)$ quantity depending on $\hat\bkappa$ and $\bkappa$,
$T_p = \tanh( \tau_t^2 \hat\kappa_p^2 / \norm{\bkappa}^2 )$, and $\tilde{T} = \sum_{k=1}^r 
( \hat\kappa_p^2 / \norm{\bkappa}^2 ) T_p $. The first term comes from the first term of 
\eqref{eq: main text: d KA} and the second term from the second and third terms of \eqref{eq: main text: d KA}. 
Finally, note that when $\kappa_p^2 \approx \hat\kappa_p^2$ for all $p \in [r]$, the model is aligned, and when all 
$\kappa_p^2$ are roughly the same, the model is balanced.

\subsection{Stage 1}
\label{sec: main text: stage 1}

In this subsection, we describe the dynamics of gradient flow in Stage~1 and explain how we control the 
growth of the errors and condition number. As mentioned earlier, for ease of presentation, we will use the 
infinite-width dynamics \eqref{eq: main text: infinite-width dyanmics} instead of the finite-width ones 
\eqref{eq: main text: d KA}. 

\paragraph{Equivalence of Stage~1 and non-contrastive methods}
Recall that we use a small $\tau_t^2$ in Stage~1. As a result, $T_p \approx 0$ for all $p \in [r]$, whence $\tilde{T}$ is 
also approximately $0$. Therefore, the second terms of \eqref{eq: main text: infinite-width dyanmics} are approximately 
$0$. In other words, only the positive pairs matter. Meanwhile, one can check that, in the infinite-width limit, 
\eqref{eq: main text: d KA (without contrastive pairs)} corresponds to the first term of \eqref{eq: main text: 
infinite-width dyanmics}, up to some multiplicative factor. This gives the equivalence of the dynamics of Stage~1 and 
the non-contrastive method. See Lemma~\ref{lemma: non-contrastive equiv stage 1} for a more formal proof in the 
finite-width setting. 

Now, we consider the contrastive loss. The main result of Stage~1 is as follows. 
\begin{lemma}[Informal version of Lemma~\ref{lemma: stage 1: main}]
  Under the assumptions of Theorem~\ref{thm: main}, the finite-width dynamics closely track the infinite-width 
  one throughout Stage~1, which takes at most $\poly(d)$ iterations, and, at the end of Stage~1, 
  we have, for any $p, q \in [r]$ and $s \in [d - r]$, 
  \begin{equation}
    \label{eq: main text: stage 1 ending state}
    \hat\kappa_p^2 / \kappa_p^2 \approx 1, \quad 
    \norm{[\KAxi]_s}^2 / \kappa_p^2 \approx 0, \quad 
    \kappa_p^2 / \kappa_q^2 
    \le O\left( \sqrt{d} \right).
  \end{equation}
  Moreover, there exists a $\bsigma \in \R^r$ such that $\max_{p, q} \kappa_p^2 / \kappa_q^2 = 
  \Omega(\sqrt{d})$ at the end of Stage~1. 
\end{lemma}

In words, \eqref{eq: main text: stage 1 ending state} says, at the end of Stage~1, $\KA \approx \KB$ in the 
relative sense, the noise-signal ratio is small, and the condition number is bounded by $O(\sqrt{d})$. 
Moreover, this bound can be achieved by some $\bsigma$, implying that with only non-contrastive pairs, the final 
condition number can be $\Theta(\sqrt{d})$. By Lemma~\ref{lemma: sufficient condition for aligned representations}, the first two conditions of \eqref{eq: main text: stage 1 ending state} imply that the learned representations are aligned. 

The proof of this lemma can be found in Section~\ref{sec: stage 1}. We discuss the high-level idea using the infinite-width 
dynamics here. Basically, we couple the convergence of $\hat\kappa_p^2 / \kappa_p^2$ (and the noise-signal ratio)
with the growth of the condition number (the discretization error). The main tool we use is the following comparative 
version of Gronwall's lemma. 

\begin{lemma}
  \label{lemma: gronwall}
  Let $A_t$ be a positive process. Let $X_t$ and $Y_t$ be defined as 
  $
    \dot{X}_t \le - A_t X_t, 
    \dot{Y}_t \le \beta A_t X_t Y_t, 
  $
  with $X_0, Y_0, \beta$ being positive. Then, for any $T \ge 0$, we have $Y_T \le Y_0 \exp(\beta X_0)$. 
\end{lemma}

Here, $X_t$ represents the progress we have made and $Y_t$ the error we wish to control. In our case, $X_t$ represents the 
maximum between $1 - \hat\kappa_p^2 / \kappa_p^2$ and the noise-signal ratio, and $Y_t$ represents the discretization 
error and condition number. This lemma says that, if the error growth rate decreases as we make more progress, then by 
coupling these two processes, we can make sure the error does not blow up. Note that the exponent of the RHS is $\beta X_0$,
which is independent of time $t$. Hence, as long as it is $O(\log d)$, the final error $Y_T$ is $O(\poly(d))$. 

To be more specific, we have, in the (noiseless) infinite-width limit, 
\begin{align*}
  \frac{\rd}{\rd t} \frac{\hat\kappa_p^2}{\kappa_p^2}
  &\approx
    \frac{4 \left( 1 - \tilde{S} \right) \sigma_p^2}{\norm{\bkappa}^2 }
    \left( 1 + \frac{\hat\kappa_p^2}{\kappa_p^2} \right) 
    \left( 1 - \frac{\hat\kappa_p^2}{\kappa_p^2} \right) 
  \ge \frac{4 \left( 1 - \tilde{S} \right) \sigma_p^2}{\norm{\bkappa}^2 }
    \left( 1 - \frac{\hat\kappa_p^2}{\kappa_p^2} \right), \\
  \frac{\rd}{\rd t} \frac{\kappa_p^2}{\kappa_q^2}
  &\approx
      \frac{ 4 \left( 1 - \tilde{S} \right) \sigma_p^2 }{ \norm{\bkappa}^2 }
      \left(
        \frac{ \hat\kappa_p^2 }{ \kappa_p^2 } 
        - \frac{ \norm{\hat\bkappa}^2 }{\norm{\bkappa}^2}
      \right) 
      \frac{ \kappa_p^2 }{ \kappa_q^2 }
    - 
    \frac{ 4 \left( 1 - \tilde{S} \right) \sigma_q^2 }{ \norm{\bkappa}^2 }
      \left(
        \frac{ \hat\kappa_q^2 }{ \kappa_q^2 }
        - \frac{ \norm{\hat\bkappa}^2 }{\norm{\bkappa}^2}
      \right) 
    \frac{\kappa_p^2}{\kappa_q^2} \\
  &\le 
    \frac{ 8 \left( 1 - \tilde{S} \right) \max\braces{ \sigma_p^2, \sigma_q^2 } }{ \norm{\bkappa}^2 }
      \left(
        \left| \frac{ \hat\kappa_p^2 }{ \kappa_p^2 } - 1 \right|
        + \left| 1 - \frac{ \norm{\hat\bkappa}^2 }{\norm{\bkappa}^2} \right|
      \right) 
      \frac{ \kappa_p^2 }{ \kappa_q^2 },
\end{align*}
where the approximation comes from the second terms of \eqref{eq: main text: infinite-width dyanmics} being approximately $0$. 
Set $A_t = 4 (1 - \tilde{S}) \sigma_{\min}^2 / \norm{\bkappa}^2$. Recall $\sigma_{\max}^2 / \sigma_{\min}^2 \le c \log d$
for some small constant $c > 0$. Hence, we can choose $\beta = O(\log d)$ and apply Lemma~\ref{lemma: gronwall} to conclude 
that $\hat\kappa_p^2 / \kappa_p^2$ will become inverse polynomially small while the condition number can be bounded by some 
value that is at most polynomially large and depends on the initialization but not the time (cf.~the first and third plots
on the second row of Fig.~\ref{fig: simulation}).

\subsection{Stage 2}

In Stage~2, $\tau_t^2$ is no longer $o(1)$, and now the second terms of \eqref{eq: main text: infinite-width dyanmics} 
come into play. We show that, in this stage, the model will reduce the condition number of $\KA$ to approximately $1$.
By Lemma~\ref{lemma: sufficient condition for balanced representations}, this implies that the learned representations 
are balanced. Formally, we have the following lemma. 

\begin{lemma}[Informal version of Lemma~\ref{lemma: stage 2: main}]
  Under the assumptions of Theorem~\ref{thm: main}, the finite-width dynamics still closely track the infinite-width one 
  throughout Stage~2, which again takes at most $\poly(d)$ iterations, and, at the end of Stage~2, we have, 
  for any $p, q \in [r]$ and $s \in [d - r]$,
  \begin{equation}
    \label{eq: main text: stage 2 ending state}
    \hat\kappa_p^2 / \kappa_p^2 \approx 1, \quad 
    \norm{[\KAxi]_s}^2 / \kappa_p^2 \approx 0, \quad 
    \kappa_p^2 / \kappa_q^2 \approx 1.
  \end{equation}
\end{lemma}

The proof of this lemma can be found in Section~\ref{sec: stage 2}. As in the previous subsection, here we discuss the 
proof strategy using the (noiseless) infinite-width limit. Maintaining $\hat\kappa_p^2 / \kappa_p^2 \approx 1$ and the 
noise-signal ratio being close to $0$ in this stage is similar to Stage~1, and we will not repeat the proof here. Instead, 
we will simply assume that $\hat\kappa_p^2 = \kappa_p^2$ and $\norm{[\KAxi]_s}^2 / \kappa_p^2 = 0$. Under this assumption, 
the first terms of \eqref{eq: main text: infinite-width dyanmics} vanish, and we have 
\[ 
    \frac{\rd}{\rd t} \kappa_p^2 
    = - 4 \left( 1 - \tilde{S}  \right)
      \left(
        \frac{ \hat\kappa_p^2  }{\norm{\bkappa}^2} T_p
        - \frac{ \kappa_p^2 }{\norm{\bkappa}^2} \tilde{T}
      \right)
      \sigma_p^2 
    = - 4 \left( 1 - \tilde{S}  \right) \frac{ \kappa_p^2  }{\norm{\bkappa}^2} 
      \left( T_p - \tilde{T} \right) \sigma_p^2. 
\]
Recall that $T_p = \tanh( \kappa_p^2 / \norm{\bkappa}^2 )$ and $\tilde{T}$ is a weighted average of these 
$T_p$'s. Hence, the $T_p - \tilde{T}$ term on the RHS will push $\kappa_p^2$ toward the average. To obtain an estimation 
for the convergence rate, we consider the ratio and compute 
\[ 
  \frac{\rd}{\rd t} \frac{\kappa_p^2}{\kappa_q^2}
  = 
    - \frac{ 4 (1 - \tilde{S}) }{ \norm{\bkappa}^2 } 
      \left(
        \left( T_p - \tilde{T} \right) \sigma_p^2
        - \left( T_q - \tilde{T} \right) \sigma_q^2
      \right)
      \frac{\kappa_p^2}{\kappa_q^2}. 
\] 
Suppose that $\kappa_p^2$ is the largest and $\kappa_q^2$ is the smallest singular value. Then, we have 
$T_p - \tilde{T} \ge 0$ and $T_q - \tilde{T} \le 0$. As a result, 
\[ 
  \frac{\rd}{\rd t} \frac{\kappa_p^2}{\kappa_q^2}
  \lesssim 
    - \frac{ 4 (1 - \tilde{S}) \sigma_{\min}^2}{ \norm{\bkappa}^2 } 
      \left( T_p -  T_q \right)
      \frac{\kappa_p^2}{\kappa_q^2}. 
\]
When the condition number is larger than, say, $2$, $T_p - T_q$ is $\Omega(1)$, and the above implies a linear convergence 
rate. See Appendix~\ref{sec: stage 2: convergence rate} for details on the convergence rate.

\section{Experimental results}
\label{sec: main text: experiments}

\begin{figure}[htbp]
  \centering
  \includegraphics[height=3.6cm]{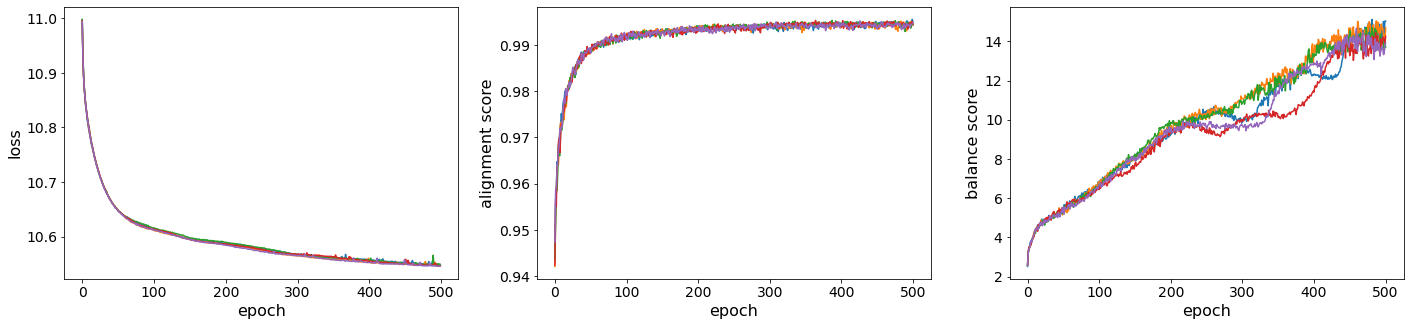} \\
  \includegraphics[height=3.7cm]{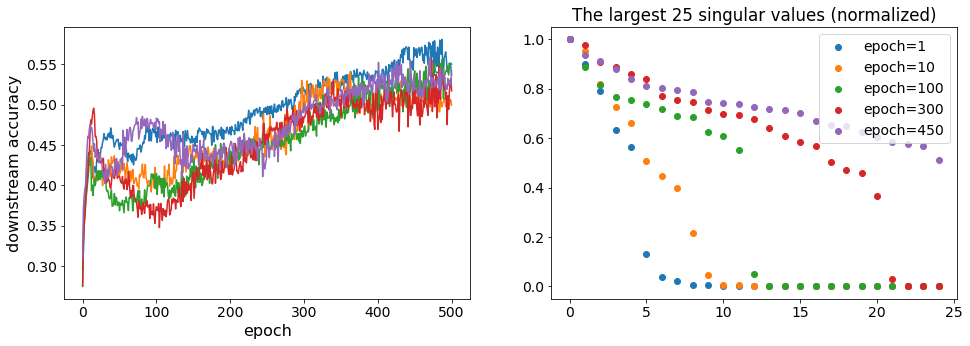}
  \caption{Results of the MSCOCO experiments. 
    The top row plots report the training loss, alignment scores, and balance scores during training, respectively. 
    The bottom row plots report the downstream accuracies and the largest $25$ singular value of $\Sigma_f$ at 
    different epochs, normalized so that the largest one has value $1$. One can see that the alignment score quickly 
    reaches near $100\%$, and the balance score, as well as the downstream accuracy, increases gradually during 
    training, which matches our theoretical analysis. 
  }
  \label{fig: mscoco}
\end{figure}

Besides the simulation results reported in Figure~\ref{fig: simulation}, we also conduct experiments on the 
MSCOCO-2014 dataset \citep{lin_microsoft_2014} using more practical models. See Figure~\ref{fig: mscoco} for the results. 
For the text part, we use a pre-trained RoBERTa model \citep{liu_roberta_2019}, followed by a 3-layer fully-connected network
with batch norm between layers. For the image part, we use a pre-trained ResNet101 \citep{he_deep_2015}, followed by 
the same layers. In both parts, the width of the fully-connected layers and the output dimension are $768$. 
We freeze the pre-trained parts of the model and only train the fully-connected parts. 

We measure the quality of the learned representation using its zero-shot performance on the MSCOCO-2014 validation set. 
Unlike common image classification datasets, images in the MSCOCO dataset usually have multiple labels, each corresponding
to an object that appears in the image, and there are $80$ categories in total. We regard a prediction to be correct if 
it matches one label. The zero-shot classification is done in the same way as in \cite{radford_learning_2021}. Namely, 
we compute the image embedding and the embeddings of prompts ``This is a [LABEL\_NAME]'', and use the prompt with the 
highest correlation with the image embedding as the prediction.

\section{Conclusion and Discussion}

In this work, we study the role of contrastive pairs in multimodal learning, and show that contrastive pairs 
are important for the model to learn representations that are both \textbf{aligned and balanced}. Our work extends previous results in several 
directions: First, we consider the more complicated multimodal learning problem. Meanwhile, our data generating 
model is inhomogeneous, and we show that in this case, contrastive method will learn a balanced model, and without 
contrastive pairs, it will collapse to an approximately rank-$1$ solution.
We also include output normalization in our analysis, a technique that is widely used in practice but is still 
under-studied in theory. 

However, despite the complexity of the analysis, our model is still mostly linear, which is very different from the 
models used in practice. Also, for the results on non-contrastive methods, we do not consider more advanced 
training techniques such as \cite{grill_bootsrap_2020} and \cite{chen_exploring_2021}. We leave the analysis of 
these more practical techniques for future work.

\bibliography{references}

\newpage
\appendix

\section{Gradient Calculation}

In this section, we compute the gradients and the equations governing some related quantities. We first 
consider the general finite-width case and then the infinite-width case, in which we have simple formulas for 
many quantities of interests. We postpone all proofs to the end of each subsection. 

\subsection{The finite-width case}

First, we prove the following auxiliary lemma. 

\begin{lemma}
  \label{lemma: grad exp(fA fB)}
  For any $\xA^+$ and $\xB^-$, we have
  \[ 
    \nabla_{\WA} \exp(\tau_t^2 \fA^+ \cdot \fB^-) 
    = \frac{ \tau_t^2 \exp(\tau_t^2 \fA^+ \cdot \fB^-) }{ N_\A }
      \left(
        \xA^+(\fB^-)\trans
        - \inprod{\fA^+}{\fB^-} 
          \frac{\left(\A\KA\trans + \A_{\bxi}\KAxi\trans \right) / d}{ N_\A }
      \right).
  \] 
\end{lemma}

Then we compute the gradients. 

\begin{lemma}
  \label{lemma: grad Loss}
  We have
  \begin{align*}
    \nabla_{\WA} \Loss
    &= - \tau_t^2 
      \E_{\xA^+, \xB^+} \braces{ 
        \left( 2 - S_\A(\xA^+, \xB^+) - S_\B(\xA^+, \xB^+) \right)
        \left(
          \frac{\xA^+(\fB^+)\trans}{N_\A}
          - \inprod{\fA^+}{\fB^+} 
            \frac{\left(\A\KA\trans + \A_{\bxi}\KAxi\trans \right)}{ N_\A^2 d }
        \right)
      } \\
      &\qquad
      + \frac{K \tau_t^2 }{N_\A}
      \E_{\xA^+, \xB^\pm} \braces{ 
        \frac{ S_\A(\xA^+, \xB^+) \exp(\tau_t^2 \fA^+ \cdot \fB^-)}{\exp(\tau_t^2 \fA^+ \cdot \fB^+)}
        \left(
          \xA^+(\fB^-)\trans
          - \inprod{\fA^+}{\fB^-} 
            \frac{\left(\A\KA\trans + \A_{\bxi}\KAxi\trans \right) / \sqrt{d}}
                 { \sqrt{\norm{\KA}_F^2 + \norm{\KAxi}_F^2} }
        \right)
      } \\
      &\qquad
      + \frac{ K \tau_t^2 }{ N_\A }
      \E_{\xA^\pm, \xB^+} \braces{ 
        \frac{S_\B(\xA^+, \xB^+) \exp(\tau_t^2 \fA^- \cdot \fB^+)}{\exp(\tau_t^2 \fA^+ \cdot \fB^+)}
          \left(
            \xA^-(\fB^+)\trans
            - \inprod{\fA^-}{\fB^+} 
              \frac{\left(\A\KA\trans + \A_{\bxi}\KAxi\trans \right) / \sqrt{d}}
                   { \sqrt{\norm{\KA}_F^2 + \norm{\KAxi}_F^2} }
          \right)
      }.
  \end{align*}
  The formula for $\nabla_{\WB} \Loss$ can be obtained by interchanging the roles of $\A$ and $\B$.
\end{lemma}

Instead of tracking $\WA$ and $\WB$ directly, we are going to track $\KA$, $\KB$, $\KAxi$ and 
$\KBxi$. Their dynamics are governed by the following equations, which are direct corollaries of 
Lemma~\ref{lemma: grad Loss}. 

\begin{lemma}
  \label{lemma: d KA, d KAxi}
  We have
  \begin{align*}
    & \frac{\rd}{\rd t} \KA \\
    =\;& \tau_t^2
      \E_{\xA^+, \xB^+} \braces{ 
        \left( 2 - S_\A(\xA^+, \xB^+) - S_\B(\xA^+, \xB^+) \right)
        \left(
          \frac{ (\KB\z^+ + \KBxi\bxi_\B) (\z^+)\trans }{N_\A N_\B}
          - \inprod{\fA^+}{\fB^+}\frac{\KA}{ N_\A^2 d}
        \right)
      } \diag(\bsigma^2)
      \\
      &\qquad
      - K \tau_t^2
      \E_{\xA^+, \xB^\pm} \braces{ 
        \frac{ S_\A(\xA^+, \xB^+) \exp(\tau_t^2 \fA^+ \cdot \fB^-)}{\exp(\tau_t^2 \fA^+ \cdot \fB^+)}
        \left(
          \frac{ \fB^-(\z^+)\trans }{N_\A}
          - \inprod{\fA^+}{\fB^-}  \frac{\KA}{ N_\A^2 d }
        \right)
      } \diag(\bsigma^2) \\
      &\qquad
      - K \tau_t^2 
      \E_{\xA^\pm, \xB^+} \braces{ 
        \frac{S_\B(\xA^+, \xB^+) \exp(\tau_t^2 \fA^- \cdot \fB^+)}{\exp(\tau_t^2 \fA^+ \cdot \fB^+)}
          \left(
            \frac{ \fB^+(\z^-)\trans }{N_\A}
            - \inprod{\fA^-}{\fB^+} \frac{ \KA }{ N_\A^2 d }
          \right)
      } \diag(\bsigma^2),
  \end{align*}
  and 
  \begin{align*}
    \frac{\rd}{\rd t} \KAxi
    &= \frac{\tau_t^2 }{N_\A}
      \E_{\xA^+, \xB^+} \braces{ 
        \left( 2 - S_\A(\xA^+, \xB^+) - S_\B(\xA^+, \xB^+) \right)
        \left(
          \fB^+(\bxi_\A^+)\trans 
          - \frac{ \inprod{\fA^+}{\fB^+}  \KAxi / \sqrt{d}}
                 { \sqrt{\norm{\KA}_F^2 + \norm{\KAxi}_F^2} }
        \right)
      } \sigma_\xi^2 \\
      &\qquad
      - \frac{K \tau_t^2 }{N_\A}
      \E_{\xA^+, \xB^\pm} \braces{ 
        \frac{ S_\A(\xA^+, \xB^+) \exp(\tau_t^2 \fA^+ \cdot \fB^-)}{\exp(\tau_t^2 \fA^+ \cdot \fB^+)}
        \left(
          \fB^-(\bxi_\A^+)\trans
          - \frac{ \inprod{\fA^+}{\fB^-}  \KAxi / \sqrt{d}}
                 { \sqrt{\norm{\KA}_F^2 + \norm{\KAxi}_F^2} }
        \right)
      } \sigma_\xi^2 \\
      &\qquad
      - \frac{ K \tau_t^2 }{ N_\A }
      \E_{\xA^\pm, \xB^+} \braces{ 
        \frac{S_\B(\xA^+, \xB^+) \exp(\tau_t^2 \fA^- \cdot \fB^+)}{\exp(\tau_t^2 \fA^+ \cdot \fB^+)}
          \left(
            \fB^+(\bxi_\A^-)\trans
            - \frac{ \inprod{\fA^-}{\fB^+} \KAxi / \sqrt{d}}
                   { \sqrt{\norm{\KA}_F^2 + \norm{\KAxi}_F^2} }
          \right)
      } \sigma_\xi^2.
  \end{align*}
\end{lemma}

We can rewrite the above result as follows. 
\begin{corollary}
  \label{cor: dynamics, Q}
  Define 
  \begin{align*}
    Q_0
    &:= 
      - \E_{\xA^+, \xB^+} \braces{ 
        \left( 2 - S_\A(\xA^+, \xB^+) - S_\B(\xA^+, \xB^+) \right)
        \inprod{\fA^+}{\fB^+} 
      } \\
      &\qquad
      + K 
      \E_{\xA^+, \xB^\pm} \braces{ 
        \frac{ S_\A(\xA^+, \xB^+) \exp(\tau_t^2 \fA^+ \cdot \fB^-)}{\exp(\tau_t^2 \fA^+ \cdot \fB^+)}
        \inprod{\fA^+}{\fB^-} 
      }  \\ 
      &\qquad
      + K 
      \E_{\xA^\pm, \xB^+} \braces{ 
        \frac{S_\B(\xA^+, \xB^+) \exp(\tau_t^2 \fA^- \cdot \fB^+)}{\exp(\tau_t^2 \fA^+ \cdot \fB^+)}
        \inprod{\fA^-}{\fB^+} 
      },
  \end{align*}
  and 
  \begin{align*}
    \Q_1
    &:= \E \braces{
        \left( 2 - S_\A(\xA^+, \xB^+) - S_\B(\xA^+, \xB^+) \right)
        \z^+(\z^+)\trans d
      } \\
      &\qquad
      - K \E \braces{
        \frac{ S_\A(\xA^+, \xB^+) \exp(\tau_t^2 \fA^+ \cdot \fB^-)}{\exp(\tau_t^2 \fA^+ \cdot \fB^+)}
        \z^- (\z^+)\trans d
      } \\
      &\qquad
      - K \E \braces{
        \frac{S_\B(\xA^+, \xB^+) \exp(\tau_t^2 \fA^- \cdot \fB^+)}{\exp(\tau_t^2 \fA^+ \cdot \fB^+)}
        \z^+ (\z^-)\trans d
      }, \\
    \Q_{1, \xi_\B}
    &:= \E \braces{
        \left( 2 - S_\A(\xA^+, \xB^+) - S_\B(\xA^+, \xB^+) \right)
        \bxi_\B^+(\z^+)\trans d
      } \\
      &\qquad
      - K \E \braces{
        \frac{ S_\A(\xA^+, \xB^+) \exp(\tau_t^2 \fA^+ \cdot \fB^-)}{\exp(\tau_t^2 \fA^+ \cdot \fB^+)}
        \bxi_\B^- (\z^+)\trans d
      } \\
      &\qquad
      - K \E \braces{
        \frac{S_\B(\xA^+, \xB^+) \exp(\tau_t^2 \fA^- \cdot \fB^+)}{\exp(\tau_t^2 \fA^+ \cdot \fB^+)}
        \bxi_\B^+ (\z^-)\trans d
      }, \\
    \Q_{1, \xi_\A}
    &:= \E \braces{
        \left( 2 - S_\A(\xA^+, \xB^+) - S_\B(\xA^+, \xB^+) \right)
        \bxi_\A^+(\z^+)\trans d
      } \\
      &\qquad
      - K \E \braces{
        \frac{ S_\A(\xA^+, \xB^+) \exp(\tau_t^2 \fA^+ \cdot \fB^-)}{\exp(\tau_t^2 \fA^+ \cdot \fB^+)}
        \bxi_\A^+ (\z^-)\trans d
      }  \\
      &\qquad
      - K \E \braces{
        \frac{S_\B(\xA^+, \xB^+) \exp(\tau_t^2 \fA^- \cdot \fB^+)}{\exp(\tau_t^2 \fA^+ \cdot \fB^+)}
        \bxi_\A^- (\z^+)\trans d
      }, \\
    \Q_2
    &:= \E \braces{
        \left( 2 - S_\A(\xA^+, \xB^+) - S_\B(\xA^+, \xB^+) \right)
        \bxi_\B^+ (\bxi_\A^+)\trans d
      } \\
      &\qquad
      - K \E \braces{
        \frac{ S_\A(\xA^+, \xB^+) \exp(\tau_t^2 \fA^+ \cdot \fB^-)}{\exp(\tau_t^2 \fA^+ \cdot \fB^+)}
        \bxi_\B^- (\bxi_\A^+)\trans d 
      } \\
      &\qquad
      - K \E \braces{ 
        \frac{S_\B(\xA^+, \xB^+) \exp(\tau_t^2 \fA^- \cdot \fB^+)}{\exp(\tau_t^2 \fA^+ \cdot \fB^+)}
        \bxi_\B^+(\bxi_\A^-)\trans d
      }.
  \end{align*}
  We have
  \begin{align*}
    \frac{\rd}{\rd t} \KA 
    &= \frac{\KB}{N_\A N_\B d} \Q_1 \diag(\bsigma^2) 
      + \frac{\KBxi}{N_\A N_\B d} \Q_{1, \xi_\B} \diag(\bsigma^2) 
      + \frac{\KA}{N_\A^2 d} Q_0 \diag(\bsigma^2), \\
    \frac{\rd}{\rd t} \KB 
    &= \frac{\KA}{N_\A N_\B d} \Q_1\trans \diag(\bsigma^2) 
      + \frac{\KAxi}{N_\A N_\B d} \Q_{1, \xi_\A} \diag(\bsigma^2) 
      + \frac{\KB}{N_\B^2 d} Q_0 \diag(\bsigma^2), \\
    \frac{\rd}{\rd t} \KAxi
    &= \frac{ \KB }{ N_\A N_\B d} \Q_{1, \bxi_\A}\trans \sigma_\xi^2
      + \frac{\KBxi}{N_\A N_\B d} \Q_2 \sigma_\xi^2
      + \frac{\KAxi}{N_\A^2 d} Q_0 \sigma_\xi^2, \\
    \frac{\rd}{\rd t} \KBxi
    &= \frac{ \KA }{ N_\A N_\B d } \Q_{1, \bxi_\B}\trans \sigma_\xi^2
      + \frac{ \KAxi }{ N_\A N_\B d } \Q_2\trans \sigma_\xi^2 
      + \frac{\KBxi}{N_\B^2 d} Q_0 \sigma_\xi^2. 
  \end{align*}
\end{corollary}

We are interested in each column of $\KA$ and $\KB$, whose dynamics are given by the next lemma. 
The next lemma also decomposes the dynamics along the radial and tangent directions. 
\begin{lemma}
  \label{lemma: d norm and d bar}
  For any $p \in [r]$ and $q \in [d-r]$, we have 
  \begin{align*}
    \frac{\rd}{\rd t} \norm{[\KA]_p}^2
    &= 2 \frac{\inprod{[\KA]_p}{[\KB\Q_1]_p}}{N_\A N_\B d} \sigma_p^2
      + 2 \frac{\inprod{[\KA]_p}{[\KBxi\Q_{1, \xi_\B}]_p}}{N_\A N_\B d}  \sigma_p^2
      + 2 \frac{\norm{[\KA]_p}^2}{N_\A^2 d} Q_0 \sigma_p^2, \\
    \frac{\rd}{\rd t} \norm{[\KB]_p}^2
    &= 2 \frac{ \inprod{[\KB]_p}{[\KA\Q_1\trans]_p}}{N_\A N_\B d} \sigma_p^2
      + 2 \frac{ \inprod{[\KB]_p}{[\KAxi\Q_{1, \xi_\A}]_p}}{N_\A N_\B d}  \sigma_p^2
      + 2 \frac{\norm{[\KB]_p}^2}{N_\B^2 d} Q_0 \sigma_p^2, \\
    \frac{\rd}{\rd t} \norm{[\KAxi]_q}^2
    &= 2 \frac{ \inprod{[\KAxi]_q}{[\KB \Q_{1, \bxi_\A}\trans]_q} }{ N_\A N_\B d}  \sigma_\xi^2
      + 2 \frac{ \inprod{[\KAxi]_q}{[\KBxi \Q_2]_q} }{N_\A N_\B d} \sigma_\xi^2
      + 2 \frac{\norm{[\KAxi]_q}_F^2}{N_\A^2 d} Q_0 \sigma_\xi^2, \\
    \frac{\rd}{\rd t} \norm{[\KBxi]_q}^2
    &= 2 \frac{ \inprod{[\KBxi]_q}{[\KA \Q_{1, \bxi_\B}\trans]_q} }{ N_\A N_\B d } \sigma_\xi^2
      + 2 \frac{ \inprod{[\KBxi]_q}{[\KAxi \Q_2\trans]_q} }{ N_\A N_\B d } \sigma_\xi^2 
      + 2 \frac{\norm{[\KBxi]_q}^2}{N_\A^2 d} Q_0 \sigma_\xi^2,
  \end{align*}
  and 
  \begin{align*}
    \frac{\rd}{\rd t} \ol{ [\KA]_p }
    &= \left( \Id - \ol{ [\KA]_p } \left( \ol{ [\KA]_p } \right)\trans  \right)
      \left(
        \frac{[\KB \Q_1]_p }{ \norm{[\KA]_p} }
        + \frac{[\KBxi \Q_{1, \xi_\B}]_p}{\norm{[\KA]_p} }
      \right)
      \frac{\sigma_p^2 }{N_\A N_\B d }, \\
    \frac{\rd}{\rd t} \ol{ [\KB]_p }
    &= \left( \Id - \ol{ [\KB]_p } \left( \ol{ [\KB]_p } \right)\trans  \right)
      \left(
        \frac{[\KA \Q_1\trans]_p }{ \norm{[\KB]_p} }
        + \frac{[\KAxi \Q_{1, \xi_\A}]_p }{ \norm{[\KB]_p} }
      \right)
      \frac{\sigma_p^2 }{N_\A N_\B d }, \\
    \frac{\rd}{\rd t} \ol{ [\KAxi]_q }
    &= \left( \Id - \ol{ [\KAxi]_q } \left( \ol{ [\KAxi]_q } \right)\trans \right)
      \left(
        \frac{ [\KB \Q_{1, \bxi_\A}\trans]_q }{ \norm{[\KAxi]_q} }
        + \frac{ [\KBxi \Q_2]_q }{ \norm{[\KAxi]_q} }
      \right) 
      \frac{\sigma_\xi^2 }{N_\A N_\B d }, \\
    \frac{\rd}{\rd t} \ol{ [\KBxi]_q }
    &= \left( \Id - \ol{ [\KBxi]_q } \left( \ol{ [\KBxi]_q } \right)\trans \right)
      \left(
        \frac{ [\KA \Q_{1, \bxi_\B}\trans]_q }{ \norm{[\KBxi]_q} }
        + \frac{ [\KAxi \Q_2\trans]_q }{ \norm{[\KBxi]_q} }
      \right) 
      \frac{\sigma_\xi^2 }{N_\A N_\B d }. 
  \end{align*}
\end{lemma}

Finally, as a simple corollary of Lemma~\ref{lemma: grad exp(fA fB)}, we have the following result on 
the gradients of the non-contrastive loss. 
\begin{lemma}
  \label{lemma: grad non-contrastive}
  For the non-contrastive loss \eqref{eq: non-constrative loss}, we have 
  \begin{align*}
    \nabla_{\WA} \hat{\Loss}
    &= \E\braces{
        \frac{ \xA^+(\fB^+)\trans }{N_\A}
        - \inprod{\fA^+}{\fB^+} 
          \frac{\left(\A\KA\trans + \A_{\bxi}\KAxi\trans \right) }{ N_\A^2 d }
      }
  \end{align*}
  As a corollary, we have, in this case, 
  \begin{align*}
    \frac{\rd}{\rd t} \KA 
    &= \E\braces{
        \frac{ \fB^+(\z^+)\trans}{N_\A} 
        - \inprod{\fA^+}{\fB^+} \frac{\KA}{ N_\A^2 d } 
      } \diag(\bsigma^2), \\
    \frac{\rd}{\rd t} \KAxi
    &= \E\braces{
        \frac{ \fB^+(\bxi_\A^+)\trans }{N_\A}
        - \inprod{\fA^+}{\fB^+} \frac{\KAxi}{ N_\A^2 d }
      } \sigma_\xi^2.  
  \end{align*}
\end{lemma}

\subsubsection*{Omitted proof of this subsection}

\begin{proof}[Proof of Lemma~\ref{lemma: grad exp(fA fB)}]
  We compute 
  \begin{align*}
    \nabla_{\WA} \exp(\tau_t^2 \fA^+ \cdot \fB^-)
    &= \tau_t^2 \exp(\tau_t^2 \fA^+ \cdot \fB^-) \nabla_{\WA} \inprod{\fA^+}{\fB^-} \\
    &= \tau_t^2 \exp(\tau_t^2 \fA^+ \cdot \fB^-) 
      \nabla_{\WA} 
      \frac{ \inprod{ \WA\trans\xA^+}{\fB^-} }
           { \sqrt{\norm{\KA}_F^2 + \norm{\KAxi}_F^2} / \sqrt{d} } \\
    &= \tau_t^2 \exp(\tau_t^2 \fA^+ \cdot \fB^-) 
      \frac{ \nabla_{\WA} \inprod{ \WA\trans\xA^+}{\fB^-} }
           { \sqrt{\norm{\KA}_F^2 + \norm{\KAxi}_F^2} / \sqrt{d} } \\
      &\qquad
      - \tau_t^2 \exp(\tau_t^2 \fA^+ \cdot \fB^-) \inprod{\fA^+}{\fB^-} 
      \frac{ \nabla_{\WA} \sqrt{\norm{\KA}_F^2 + \norm{\KAxi}_F^2} }
           { \sqrt{\norm{\KA}_F^2 + \norm{\KAxi}_F^2} }.
  \end{align*}
  For the first term, we have $\nabla_{\WA} \inprod{\WA\trans\xA^+}{\fB^-} = \xA^+(\fB^-)\trans$. 
  For the second term, we have 
  \begin{align*}
    \frac{ \nabla_{\WA} \sqrt{\norm{\KA}_F^2 + \norm{\KAxi}_F^2} }
         { \sqrt{\norm{\KA}_F^2 + \norm{\KAxi}_F^2} }
    = \frac{ \nabla_{\WA} \norm{\KA}_F^2 + \nabla_{\WA} \norm{\KAxi}_F^2 }
           { 2 (\norm{\KA}_F^2 + \norm{\KAxi}_F^2) }
    = \frac{ \A\KA\trans + \A_{\bxi}\KAxi\trans }
           { \norm{\KA}_F^2 + \norm{\KAxi}_F^2 }.
  \end{align*}
  Hence, 
  \begin{align*}
    \nabla_{\WA} \exp(\tau_t^2 \fA^+ \cdot \fB^-)
    &= \tau_t^2 \exp(\tau_t^2 \fA^+ \cdot \fB^-) 
      \frac{ \xA^+(\fB^-)\trans }{ \sqrt{\norm{\KA}_F^2 + \norm{\KAxi}_F^2} / \sqrt{d} } \\
      &\qquad
      - \tau_t^2 \exp(\tau_t^2 \fA^+ \cdot \fB^-) \inprod{\fA^+}{\fB^-} 
      \frac{ \A\KA\trans + \A_{\bxi}\KAxi\trans }
           { \norm{\KA}_F^2 + \norm{\KAxi}_F^2 } \\
    &= \frac{ \tau_t^2 \exp(\tau_t^2 \fA^+ \cdot \fB^-) }
            { \sqrt{\norm{\KA}_F^2 + \norm{\KAxi}_F^2} / \sqrt{d} }
      \left(
        \xA^+(\fB^-)\trans
        - \inprod{\fA^+}{\fB^-} 
          \frac{\left(\A\KA\trans + \A_{\bxi}\KAxi\trans \right) / \sqrt{d}}
               { \sqrt{\norm{\KA}_F^2 + \norm{\KAxi}_F^2} }
      \right).
  \end{align*}
\end{proof}

\begin{proof}[Proof of Lemma~\ref{lemma: grad Loss}]
  We compute 
  \begin{align*}
    \nabla_{\WA} \Loss_{\A}
    &= - \E \frac{ \nabla_{\WA} S_{\A}(\xA^+, \xB^+) }{ S_{\A}(\xA^+, \xB^+) } \\
    &= - \E \braces{ 
        \frac{1}{ S_{\A}(\xA^+, \xB^+) } 
        \nabla_{\WA} 
        \frac{ \exp(\tau_t^2 \fA^+ \cdot \fB^+ ) }
             {\exp(\tau_t^2 \fA^+ \cdot \fB^+ ) + K \E_{\z^-} \exp(\tau_t^2 \fA^+ \cdot \fB^- ) }
      } \\
    &= - \E \braces{ 
        \frac{1}{ S_{\A}(\xA^+, \xB^+) } 
        \frac{ \nabla_{\WA}  \exp(\tau_t^2 \fA^+ \cdot \fB^+ ) }
             {\exp(\tau_t^2 \fA^+ \cdot \fB^+ ) + K \E_{\z^-} \exp(\tau_t^2 \fA^+ \cdot \fB^- ) }
      } \\
      &\qquad
      + \E \braces{ 
        \frac{ 
          \nabla_{\WA}  \exp(\tau_t^2 \fA^+ \cdot \fB^+ ) 
          + K \E_{\z^-} \nabla_{\WA}  \exp(\tau_t^2 \fA^+ \cdot \fB^- ) 
        }{\exp(\tau_t^2 \fA^+ \cdot \fB^+ ) + K \E_{\z^-} \exp(\tau_t^2 \fA^+ \cdot \fB^- ) }
      }.
  \end{align*}
  By Lemma~\ref{lemma: grad exp(fA fB)}, the first term is 
  \begin{align*}
    & - \E \braces{ 
        \frac{1}{ S_{\A}(\xA^+, \xB^+) } 
        \frac{ \nabla_{\WA}  \exp(\tau_t^2 \fA^+ \cdot \fB^+ ) }
             {\exp(\tau_t^2 \fA^+ \cdot \fB^+ ) + K \E_{\z^-} \exp(\tau_t^2 \fA^+ \cdot \fB^- ) }
      } \\
    =\;& 
      - \frac{\tau_t^2 }{ N_\A }
      \E_{\xA^+, \xB^+} 
      \braces{ 
          \xA^+(\fB^+)\trans
          - \inprod{\fA^+}{\fB^+} 
            \frac{\left(\A\KA\trans + \A_{\bxi}\KAxi\trans \right) / \sqrt{d}}
                 { \sqrt{\norm{\KA}_F^2 + \norm{\KAxi}_F^2} }
      },
  \end{align*}
  and the second term is 
  \begin{align*}
    & \E \braces{ 
        \frac{ 
          \nabla_{\WA}  \exp(\tau_t^2 \fA^+ \cdot \fB^+ ) 
          + K \E_{\xB^-} \nabla_{\WA}  \exp(\tau_t^2 \fA^+ \cdot \fB^- ) 
        }{\exp(\tau_t^2 \fA^+ \cdot \fB^+ ) + K \E_{\z^-} \exp(\tau_t^2 \fA^+ \cdot \fB^- ) }
      } \\
    =\;& 
      \frac{\tau_t^2}{N_\A}
      \E \braces{  
        S_\A(\xA^+, \xB^+)
        \left(
          \xA^+(\fB^+)\trans
          - \inprod{\fA^+}{\fB^+} 
            \frac{\left(\A\KA\trans + \A_{\bxi}\KAxi\trans \right) / \sqrt{d}}
                 { \sqrt{\norm{\KA}_F^2 + \norm{\KAxi}_F^2} }
        \right)
      } \\
      &\qquad
      + \frac{K \tau_t^2 }{N_\A}
      \E \braces{ 
        \frac{ S_\A(\xA^+, \xB^+) \exp(\tau_t^2 \fA^+ \cdot \fB^-)}{\exp(\tau_t^2 \fA^+ \cdot \fB^+)}
        \left(
          \xA^+(\fB^-)\trans
          - \inprod{\fA^+}{\fB^-} 
            \frac{\left(\A\KA\trans + \A_{\bxi}\KAxi\trans \right) / \sqrt{d}}
                 { \sqrt{\norm{\KA}_F^2 + \norm{\KAxi}_F^2} }
        \right)
      }.
  \end{align*}
  Thus, 
  \begin{align*}
    \nabla_{\WA} \Loss_\A 
    &= - \frac{\tau_t^2 }{N_\A}
      \E_{\xA^+, \xB^+} \braces{ 
        \left( 1 -  S_\A(\xA^+, \xB^+) \right)
        \left(
          \xA^+(\fB^+)\trans
          - \inprod{\fA^+}{\fB^+} 
            \frac{\left(\A\KA\trans + \A_{\bxi}\KAxi\trans \right) / \sqrt{d}}
                 { \sqrt{\norm{\KA}_F^2 + \norm{\KAxi}_F^2} }
        \right)
      } \\
      &\qquad
      + \frac{K \tau_t^2 }{N_\A}
      \E \braces{ 
        \frac{ S_\A(\xA^+, \xB^+) \exp(\tau_t^2 \fA^+ \cdot \fB^-)}{\exp(\tau_t^2 \fA^+ \cdot \fB^+)}
        \left(
          \xA^+(\fB^-)\trans
          - \inprod{\fA^+}{\fB^-} 
            \frac{\left(\A\KA\trans + \A_{\bxi}\KAxi\trans \right) / \sqrt{d}}
                 { \sqrt{\norm{\KA}_F^2 + \norm{\KAxi}_F^2} }
        \right)
      }.
  \end{align*}
  Then, for $\nabla_{\WA}\Loss_\B$, we compute 
  \begin{align*}
    \nabla_{\WA} \Loss_\B
    &= - \E \frac{\nabla_{\WA} S_\B(\xA^+, \xB^+)}{S_\B(\xA^+, \xB^+)} \\
    &= - \E \braces{
        \frac{1}{S_\B(\xA^+, \xB^+)} 
        \frac{ \nabla_{\WA} \exp(\tau_t^2 \fA^+ \cdot \fB^+ ) }
             {\exp(\tau_t^2 \fA^+ \cdot \fB^+ ) + K \E_{\xA^-} \exp(\tau_t^2 \fA^- \cdot \fB^+ ) }
      } \\
      &\qquad
      + \E \braces{
        \frac{ \nabla_{\WA} \exp(\tau_t^2 \fA^+ \cdot \fB^+ ) + K \nabla_{\WA} \E_{\xA^-} \exp(\tau_t^2 \fA^- \cdot \fB^+ )  }
             {\exp(\tau_t^2 \fA^+ \cdot \fB^+ ) + K \E_{\xA^-} \exp(\tau_t^2 \fA^- \cdot \fB^+ ) }
      }.
  \end{align*}
  Again, by Lemma~\ref{lemma: grad exp(fA fB)}, the first term is 
  \begin{align*}
    & - \E \braces{
        \frac{1}{S_\B(\xA^+, \xB^+)} 
        \frac{ \nabla_{\WA} \exp(\tau_t^2 \fA^+ \cdot \fB^+ ) }
             {\exp(\tau_t^2 \fA^+ \cdot \fB^+ ) + K \E_{\xA^-} \exp(\tau_t^2 \fA^- \cdot \fB^+ ) }
      } \\
    =\;& - \frac{ \tau_t^2 }{ N_\A }
      \E \braces{
        \xA^+(\fB^+)\trans
        - \inprod{\fA^+}{\fB^+} 
          \frac{\left(\A\KA\trans + \A_{\bxi}\KAxi\trans \right) / \sqrt{d}}
               { \sqrt{\norm{\KA}_F^2 + \norm{\KAxi}_F^2} }
      },
  \end{align*}
  and the second term is 
  \begin{align*}
    & \E \braces{
        \frac{ \nabla_{\WA} \exp(\tau_t^2 \fA^+ \cdot \fB^+ ) + K \nabla_{\WA} \E_{\xA^-} \exp(\tau_t^2 \fA^- \cdot \fB^+ )  }
             {\exp(\tau_t^2 \fA^+ \cdot \fB^+ ) + K \E_{\xA^-} \exp(\tau_t^2 \fA^- \cdot \fB^+ ) }
      } \\
    =\;& \frac{ \tau_t^2 }{ N_\A }
      \E \braces{
        S_{\B}(\xA^+, \xB^+)
        \left(
          \xA^+(\fB^+)\trans
          - \inprod{\fA^+}{\fB^+} 
            \frac{\left(\A\KA\trans + \A_{\bxi}\KAxi\trans \right) / \sqrt{d}}
                 { \sqrt{\norm{\KA}_F^2 + \norm{\KAxi}_F^2} }
        \right)
      } \\
      &\qquad
      + \frac{ K \tau_t^2 }{ N_\A }
      \E \braces{
        \frac{S_\B(\xA^+, \xB^+) \exp(\tau_t^2 \fA^- \cdot \fB^+)}{\exp(\tau_t^2 \fA^+ \cdot \fB^+)}
          \left(
            \xA^-(\fB^+)\trans
            - \inprod{\fA^-}{\fB^+} 
              \frac{\left(\A\KA\trans + \A_{\bxi}\KAxi\trans \right) / \sqrt{d}}
                   { \sqrt{\norm{\KA}_F^2 + \norm{\KAxi}_F^2} }
          \right)
      }.
  \end{align*}
  Thus, 
  \begin{align*}
    \nabla_{\WA} \Loss_\B
    &= - \frac{ \tau_t^2 }{ N_\A }
      \E \braces{
        (1 - S_\B(\xA^+, \xB^+))
        \left(
          \xA^+(\fB^+)\trans
          - \inprod{\fA^+}{\fB^+} 
            \frac{\left(\A\KA\trans + \A_{\bxi}\KAxi\trans \right) / \sqrt{d}}
                 { \sqrt{\norm{\KA}_F^2 + \norm{\KAxi}_F^2} }
        \right)
      } \\
      &\qquad
      + \frac{ K \tau_t^2 }{ N_\A }
      \E \braces{
        \frac{S_\B(\xA^+, \xB^+) \exp(\tau_t^2 \fA^- \cdot \fB^+)}{\exp(\tau_t^2 \fA^+ \cdot \fB^+)}
          \left(
            \xA^-(\fB^+)\trans
            - \inprod{\fA^-}{\fB^+} 
              \frac{\left(\A\KA\trans + \A_{\bxi}\KAxi\trans \right) / \sqrt{d}}
                   { \sqrt{\norm{\KA}_F^2 + \norm{\KAxi}_F^2} }
          \right)
      }.
  \end{align*}
  Combine these formulas together and we complete the proof. 
\end{proof}

\subsection{The infinite-width case}

Now we consider the noiseless infinite-width dynamics. The results of this subsection will not be used in the 
proof. It mainly serves as a way to give intuitions on how the dynamics look. As we have discussed in the 
main text, in this noiseless infinite-width case, it suffices to track $\kappa_p^2 = \norm{[\KA]_p}^2$ and 
$\hat\kappa_p^2 = \inprod{[\KA]_p}{[\KB]_p}$. 

\begin{lemma}
  \label{lemma: infinte-width: d kappa}
  In the noiseless infinite-width case, we have 
  \begin{align*}
    \frac{\rd}{\rd t} \kappa_p^2 
    &= 4 \left( 1 - \tilde{S} \right)
      \left(
        \frac{\hat\kappa_p^2}{\norm{\bkappa}^2 }
        - \frac{ \norm{\hat\bkappa}^2 }{\norm{\bkappa}^2}
          \frac{\kappa_p^2}{ \norm{\bkappa}^2 }
      \right) 
      \sigma_p^2 
      - 4 \left( 1 - \tilde{S}  \right)
      \left(
        \frac{ \hat\kappa_p^2  }{\norm{\bkappa}^2} T_p
        - \frac{ \kappa_p^2 }{\norm{\bkappa}^2} \tilde{T}
      \right)
      \sigma_p^2, \\
    \frac{\rd}{\rd t} \hat\kappa_p^2
    &= 4 \left( 1 - \tilde{S} \right)
      \left(
        \frac{\kappa_p^2}{\norm{\bkappa}^2 }
        - \frac{ \norm{\hat\bkappa}^2 }{\norm{\bkappa}^2}
          \frac{\hat\kappa_p^2}{ \norm{\bkappa}^2 }
      \right) 
      \sigma_p^2 
      - 4 \left( 1 - \tilde{S}  \right)
      \left(
        \frac{\kappa_p^2}{\norm{\bkappa}^2} T_p
        - \frac{\hat\kappa_p^2}{\norm{\bkappa}^2} \tilde{T}
      \right)
      \sigma_p^2, 
  \end{align*}
\end{lemma}
\begin{proof}
  First, note hat 
  \begin{align*}
    \inprod{\fA^+}{\fB^-}
    = \frac{ \inprod{\KBA\z^+}{\z^-} }{ \norm{\bkappa}^2 / d }
    = \frac{ \inprod{\z^+}{\z^-}_{\hat\bkappa^2} }{ \norm{\bkappa}^2 / d }. 
  \end{align*}
  This implies that (a) $\inprod{\fA^+}{\fB^+}$ does not depend on the actual value of $\z^+$, and (b) if we flip 
  the signs of $z^\pm_p$ simultaneously, then the value of $\inprod{\fA^+}{\fB^-}$ remain unchanged. To compute 
  $S_\A$, we then need to take expectation over $\z^-$. We compute 
  \[
    \E_{\z^-} \exp\left( \tau_t^2 \fA^+ \cdot \fB^- \right)
    = \prod_{k=1}^r \E_{z^-_k} \exp\left( \tau_t^2 \frac{\hat\kappa_k^2 z^+_k z^-_k}{\norm{\bkappa}^2 / d} \right)
    = \prod_{k=1}^r \cosh\left( \frac{\tau_t^2 \hat\kappa_k^2}{\norm{\bkappa}^2 } \right) 
    =: Z_c. 
  \]
  Again, it does not depend on the actual value of $\z^+$. One can conduct similar calculation for $S_\B$ and, 
  consequently, we have $S_\A \equiv S_\B \equiv \tilde{S}$ for some $\tilde{S}$ that depends on $\bkappa$ and 
  $\hat\bkappa$ but not on $\z^+$. Then, we can rewrite \eqref{eq: main text: d KA} as 
  \begin{align*}
    \frac{\rd}{\rd t} \KA
    &= 2 \left( 1 - \tilde{S} \right)
      \E_{\xA^+, \xB^+} \braces{ 
        \frac{ \fB^+(\z^+)\trans }{N_\A}
        - \inprod{\fA^+}{\fB^+}\frac{\KA}{ N_\A^2 d}
      } \diag(\bsigma^2)
      \\
      &\qquad
      - 2 K \tilde{S}
      \E_{\xA^+, \xB^\pm} \braces{ 
        \frac{ \exp(\tau_t^2 \fA^+ \cdot \fB^-)}{\exp(\tau_t^2 \fA^+ \cdot \fB^+)}
        \left(
          \frac{ \fB^-(\z^+)\trans }{N_\A}
          - \inprod{\fA^+}{\fB^-}  \frac{\KA}{ N_\A^2 d }
        \right)
      } \diag(\bsigma^2) \\ 
    &= 2 \left( 1 - \tilde{S} \right)
      \left(
        \frac{\KB}{\norm{\bkappa}^2 }
        - \frac{ \norm{\hat\bkappa}^2 }{\norm{\bkappa}^2}
          \frac{\KA}{ \norm{\bkappa}^2 }
      \right) 
      \diag(\bsigma^2)
      \\
      &\qquad
      - \frac{ 2 K \tilde{S} }{ \exp\left( \frac{ \tau_t^2 \norm{\hat\bkappa}^2 }{ \norm{\bkappa}^2  } \right) }
      \E_{\xA^+, \xB^\pm} \braces{ 
        \exp\left( 
          \frac{ \tau_t^2 \inprod{\z^+}{\z^-}_{\hat\bkappa^2} d }{ \norm{\bkappa}^2 }
        \right)
        \left(
          \frac{ \KB\z^-(\z^+)\trans }{ \norm{\bkappa}^2 / d }
          - \frac{ \inprod{\z^+}{\z^-}_{\hat\bkappa^2} }{ \norm{\bkappa}^2 / d}
          \frac{\KA}{ \norm{\bkappa}^2 }
        \right)
      } \diag(\bsigma^2). 
  \end{align*}
  Note that 
  \begin{align*}
    & \E_{\z^\pm}\braces{
        \exp\left( 
          \frac{ \tau_t^2 \inprod{\z^+}{\z^-}_{\hat\bkappa^2} d }{ \norm{\bkappa}^2 }
        \right) z^-_p z^+_q d
      } \\
    =\;& \indi\{p = q\}
      \E_{z^\pm_p}\braces{
        \exp\left( 
          \frac{ \tau_t^2 \hat\kappa_k^2 z^+_k z^-_k d }{ \norm{\bkappa}^2 }
        \right) z^-_p z^+_p d
      }
      \prod_{k \ne p} 
      \E_{z^\pm_k}\braces{
        \exp\left( 
          \frac{ \tau_t^2 \hat\kappa_k^2 z^+_k z^-_k d }{ \norm{\bkappa}^2 }
        \right) 
      } \\
    =\;& \indi\{p = q\}
      \sinh \left( \frac{ \tau_t^2 \hat\kappa_k^2 }{ \norm{\bkappa}^2 } \right)
      \prod_{k \ne p} 
      \cosh\left( 
        \frac{ \tau_t^2 \hat\kappa_k^2 }{ \norm{\bkappa}^2 }
      \right)  \\
    =\;& \indi\{p = q\} Z_c T_p  . 
  \end{align*}
  Meanwhile, note that 
  \begin{align*}
    \E_{\xA^+, \xB^\pm} \braces{ 
      \exp\left( 
        \frac{ \tau_t^2 \inprod{\z^+}{\z^-}_{\hat\bkappa^2} d }{ \norm{\bkappa}^2 }
      \right)
      \frac{ \inprod{\z^+}{\z^-}_{\hat\bkappa^2} }{ \norm{\bkappa}^2 / d}
    }
    &= \sum_{k=1}^r 
      \frac{ \hat\kappa_k^2 }{ \norm{\bkappa}^2 / d}
      \E_{\z^\pm} \braces{ 
        \exp\left( 
          \frac{ \tau_t^2 \inprod{\z^+}{\z^-}_{\hat\bkappa^2} d }{ \norm{\bkappa}^2 }
        \right)
        z^+_k z^-_k
      } \\
    &= Z_c \sum_{k=1}^r \frac{ \hat\kappa_k^2 }{ \norm{\bkappa}^2 } T_k \\
    &=: Z_c \tilde{T}. 
  \end{align*}
  Thus, 
  \begin{align*}
    \frac{\rd}{\rd t} \KA
    &= 2 \left( 1 - \tilde{S} \right)
      \left(
        \frac{\KB}{\norm{\bkappa}^2 }
        - \frac{ \norm{\hat\bkappa}^2 }{\norm{\bkappa}^2}
          \frac{\KA}{ \norm{\bkappa}^2 }
      \right) 
      \diag(\bsigma^2) \\
      &\qquad
      - \frac{ 2 K \tilde{S} Z_c }{ \exp\left( \frac{ \tau_t^2 \norm{\hat\bkappa}^2 }{ \norm{\bkappa}^2  } \right) }
      \left(
        \frac{ \KB  }{\norm{\bkappa}^2} \diag\left([T_k]_{k \in [r]} \right)
        - \frac{ \KA  }{\norm{\bkappa}^2} \tilde{T}
      \right)
      \diag(\bsigma^2) \\
    &= 2 \left( 1 - \tilde{S} \right)
      \left(
        \frac{\KB}{\norm{\bkappa}^2 }
        - \frac{ \norm{\hat\bkappa}^2 }{\norm{\bkappa}^2}
          \frac{\KA}{ \norm{\bkappa}^2 }
      \right) 
      \diag(\bsigma^2)  \\
      &\qquad
      - 2 \left( 1 - \tilde{S}  \right)
      \left(
        \frac{ \KB  }{\norm{\bkappa}^2} \diag\left([T_k]_{k \in [r]} \right)
        - \frac{ \KA  }{\norm{\bkappa}^2} \tilde{T}
      \right)
      \diag(\bsigma^2). 
  \end{align*}
  As a corollary, we have 
  \begin{align*}
    \frac{\rd}{\rd t} [\KA]_p
    &= 2 \left( 1 - \tilde{S} \right)
      \left(
        \frac{[\KB]_p}{\norm{\bkappa}^2 }
        - \frac{ \norm{\hat\bkappa}^2 }{\norm{\bkappa}^2}
          \frac{[\KA]_p}{ \norm{\bkappa}^2 }
      \right) 
      \sigma_p^2 
      - 2 \left( 1 - \tilde{S}  \right)
      \left(
        \frac{ [\KB]_p  }{\norm{\bkappa}^2} T_p
        - \frac{ [\KA]_p  }{\norm{\bkappa}^2} \tilde{T}
      \right)
      \sigma_p^2 .
  \end{align*}
  Hence, 
  \begin{align*}
    \frac{\rd}{\rd t} \kappa_p^2 
    &= 2 \inprod{[\KA]_p}{\frac{\rd}{\rd t} [\KA]_p} \\
    &= 4 \left( 1 - \tilde{S} \right)
      \left(
        \frac{\hat\kappa_p^2}{\norm{\bkappa}^2 }
        - \frac{ \norm{\hat\bkappa}^2 }{\norm{\bkappa}^2}
          \frac{\kappa_p^2}{ \norm{\bkappa}^2 }
      \right) 
      \sigma_p^2 
      - 2 \left( 1 - \tilde{S}  \right)
      \left(
        \frac{ \hat\kappa_p^2  }{\norm{\bkappa}^2} T_p
        - \frac{ \kappa_p^2 }{\norm{\bkappa}^2} \tilde{T}
      \right)
      \sigma_p^2 .
  \end{align*}
  By symmetry, for $\KB$, we have 
  \begin{align*}
    \frac{\rd}{\rd t} [\KB]_p
    &= 2 \left( 1 - \tilde{S} \right)
      \left(
        \frac{[\KA]_p}{\norm{\bkappa}^2 }
        - \frac{ \norm{\hat\bkappa}^2 }{\norm{\bkappa}^2}
          \frac{[\KB]_p}{ \norm{\bkappa}^2 }
      \right) 
      \sigma_p^2 
      - 2 \left( 1 - \tilde{S}  \right)
      \left(
        \frac{ [\KA]_p  }{\norm{\bkappa}^2} T_p
        - \frac{ [\KB]_p  }{\norm{\bkappa}^2} \tilde{T}
      \right)
      \sigma_p^2 .
  \end{align*}
  Then, we can compute 
  \begin{align*}
    \frac{\rd}{\rd t} \hat\kappa_p^2
    &= 4 \left( 1 - \tilde{S} \right)
      \left(
        \frac{\kappa_p^2}{\norm{\bkappa}^2 }
        - \frac{ \norm{\hat\bkappa}^2 }{\norm{\bkappa}^2}
          \frac{\hat\kappa_p^2}{ \norm{\bkappa}^2 }
      \right) 
      \sigma_p^2 
      - 4 \left( 1 - \tilde{S}  \right)
      \left(
        \frac{\kappa_p^2}{\norm{\bkappa}^2} T_p
        - \frac{\hat\kappa_p^2}{\norm{\bkappa}^2} \tilde{T}
      \right)
      \sigma_p^2. 
  \end{align*}
\end{proof}

\section{Stage 1}
\label{sec: stage 1}
In this section, we show that the following hold:
\begin{enumerate}[(a)]
  \item $\KA \approx \KB$ after Stage~1. 
  \item The noise-signal ratio is small after Stage~1. 
  \item The condition number is $O(\sqrt{d})$ in Stage~1. 
  \item The finite-width trajectory is always close to the infinite-width one in Stage~1. 
\end{enumerate}
We formalize the main results of Stage~1 below. 

\begin{lemma}[Stage 1]
  \label{lemma: stage 1: main}
  Under the assumption of Theorem~\ref{thm: main}. 
  We can choose a sufficiently (polynomially) large $m$ and a sufficiently (inverse polynomially) small $\tau_t^2$
  that may depend on the $\delta$'s that appear in this lemma so that the following statement holds. 

  Let $T_1$ be the earliest time all the following hold: 
  \begin{align*}
    & \inprod{ \ol{[\KA]_p} }{ \ol{[\KB]_p} }
      \ge 1 - \delta_-, 
    & \forall p \in [r], \\
    & \frac{ \norm{[\KAxi]_q} }{ \norm{[\KA]_p} }
      \le \delta_{N/S},
    & \forall p \in [r], q \in [d-r],
  \end{align*}
  where $\delta_-, \delta_{N/S} \in 1 / \poly(d)$ are two given parameters. We have $T_1 \le \poly(d)$. 
  Moreover, at any time $t \in [0, T_1]$, we have $\kappa_0 := \max_{p, q \in [r]} \norm{[\KA]_p} / \norm{[\KA]_q} 
  \le O(\sqrt{d})$ and 
  \begin{equation}
    \label{eq: stage 1: finite-width approx infinite-width}
    \begin{aligned}
      & \norm{[\KA]_p}^2 = (1 \pm \delta_{\A/\B}) \norm{[\KB]_p}^2, \, 
        \norm{[\KAxi]_q}^2 = (1 \pm \delta_{\A/\B}) \norm{[\KBxi]_q}^2, 
      & \forall p \in [r], q \in [d-r], \\
      & \left| 1 - \frac{\norm{[\KAxi]_p}}{\norm{[\KAxi]_q}} \right|
        \le \delta_{\xi, \kappa_0},
      & \forall p, q \in [d - r], \\
      & \max\braces{
          \left| \inprod{ \ol{[\K_{\C}]_p} }{ \ol{[\K_{\D}]_q} } \right|
          \,:\,
          \C, \D \in \braces{ \A, \B }
        }
        \le \delta_{\AB, \perp}, 
      & \forall p \ne q \in [r],  \\
      & \max\braces{
          \left| \inprod{ \ol{[\K_{\C}]_p} }{ \ol{[\K_{\D, \bxi}]_q} } \right|,
          \left| \inprod{ \ol{[\KAxi]_s} }{ \ol{[\KBxi]_q} } \right|
          \,:\,
          \C, \D \in \braces{ \A, \B }
        }
        \le \delta_{\xi, \perp} ,
      & \forall p \in [r], q, s \in [d-r],
    \end{aligned}
  \end{equation}
  where $\delta_{\A/\B}, \delta_{\xi, \kappa_0}, \delta_{\AB, \perp}, \delta_{\xi, \perp} \in 
  1 / \poly(d)$ are given parameters. 
\end{lemma}
Basically, the conditions in \eqref{eq: stage 1: finite-width approx infinite-width} mean that the norm of 
the corresponding columns are roughly the same, and the columns of all these matrices are approximately orthogonal 
to each other. Both of them are true in the infinite-width limit, and by some standard concentration argument, 
one can make all these errors to be arbitrarily inverse-polynomially small at initialization. 
Note that, as a simple corollary of \eqref{eq: stage 1: finite-width approx infinite-width}, we have 
\[
  N_\A = N_\B \left( 1 \pm \sqrt{ \delta_{\A/\B} } \right). 
\]
For notational simplicity, we also define 
\[
  \rho_- 
  = \max_{p \in [r]}\braces{ 1 - \inprod{ \ol{[\KA]_p} }{ \ol{[\KB]_p} } }, 
  \quad 
  \rho_{N/S}
  = \max_{\substack{p \in [r] \\ q \in [d - r]}} \frac{\norm{[\KAxi]_q}}{\norm{[\KA]_p}}. 
\]
The main tool we use to control the condition number and the discretization error is the following nonlinear 
version of Gronwall's lemma. 

\begin{lemma}
  \label{lemma: stage 1: gronwall}
  Let $A_t$ be a positive process. Let $X_t$ and $Y_t$ be defined as 
  \[
    \dot{X}_t \le - A_t X_t, \quad
    \dot{Y}_t \le \alpha A_t X_t Y_t, 
  \]
  with $X_0, Y_0, \alpha$ being positive. Then, for any $T \ge 0$, we have $Y_T \le Y_0 \exp(\alpha X_0)$. 
\end{lemma}
\begin{remark}
  Here, $X_t$ represents the progress we have made and $Y_t$ the error. In our case, $X_t$ is the maximum 
  between $1 - \inprod{\ol{[\KA]_p}}{\ol{[\KB]_p}}$ and the noise-signal ratio, and $Y_t$ the discretization 
  error. This lemma says that, if the error growth rate depends on the progress, then by coupling these two 
  processes, we can make sure the error does not blow up. The point of this lemma is that, with coupling,
  we do not need a very tight estimation on the convergence time nor the error growth rate. 
\end{remark}
\begin{proof}
  The solution to this ODE system is given by 
  \[
    X_T = X_0 \exp\left( - \int_0^T A_t \,\rd t \right), \quad 
    Y_T = Y_0 \exp\left( \alpha X_0 \int_0^T A_t \exp\left( - \int_0^t A_s \,\rd s \right) \, \rd t \right). 
  \]
  Note that 
  \[
    \int_0^T A_t \exp\left( - \int_0^t A_s \,\rd s \right) \, \rd t
    = - \int_0^T \,\rd \exp\left( - \int_0^t A_s \,\rd s \right)
    = 1 - \exp\left( - \int_0^T A_t \,\rd t \right). 
  \]
  Hence, 
  \[
    Y_T 
    = Y_0 \exp\left( \alpha X_0 \left( 1 - \exp\left( - \int_0^T A_t \,\rd t \right) \right) \right)
    \le Y_0 \exp\left( \alpha X_0 \right). 
  \]
\end{proof}

The organization of this section is as follows. 
In Section~\ref{sec: stage 1: Q}, we derive estimations for the $\Q$-matrices defined 
in Corollary~\ref{cor: dynamics, Q} and use them to simplify the equations governing the training dynamics. 
In Section~\ref{sec: stage 1: convergence, condition number}, we estimate the rate at which 
$1 - \inprod{\ol{[\KA]_p}}{\ol{[\KB]_p}}$ and the noise-signal ratio converge to $0$ and the growth rate of the 
condition number. In Section~\ref{sec: stage 1: discretization}, we estimate the growth rate of the 
discretization error. Then, in Section~\ref{sec: stage 1: proof of the main lemma}, we prove 
Lemma~\ref{lemma: stage 1: main}, the main lemma of Stage~1. Finally, we prove the negative result
for non-contrastive learning in Section~\ref{sec: stage 1: negative results}.

\subsection{Estimations for $Q$ and the dynamics}
\label{sec: stage 1: Q}

Thanks to Lemma~\ref{lemma: d norm and d bar}, in order to analyze the dynamics, it suffices to estimate 
the $\Q$-matrices. In this subsection, we derive estimations for them and use these estimations to simplify 
the equations in Lemma~\ref{lemma: d norm and d bar}. Recall that, in Stage~1, $\tau_t^2$ is small. Hence, 
$\exp(\tau_t^2 \inprod{\fA^+}{\fB^-}) = 1 \pm O(\tau_t^2)$. With this approximation, one can derive the 
following estimation for $S_\A$ and $S_\B$. 

\begin{lemma}[Estimations for $S$]
  \label{lemma: stage 1: estimations for S}
  In Stage~1, we have 
  \[
    S_\A 
    = \frac{1}{1 + K} 
      \pm O_z\left( \tau_t^2 \right)
      \pm O\left(\tau_t^2 d \delta_{\xi, \perp} \rho_{N/S} \right),
  \]
  and the same is also true for $S_\B$. Here, $O_z(\tau_t^2)$ means an $O(\tau_t^2)$ quantity that can depend on $z$ but not
  on $\bxi_\A$ or $\bxi_\B$.  
\end{lemma}

The proofs of this lemma and all following lemmas are deferred to the end of this subsection. Note that we derive 
a slightly finer estimation for the noise-related part. This additional $\rho_{N/S}$ will be used cancel with 
terms like $\norm{[\KA]_p} / \norm{[\KAxi]_q}$, at the cost of a $\kappa_0$ factor, in later analysis. We 
emphasize here that $O_z$ does not depend on the noises so that later we can argue 
$\E_{\bxi} \braces{ O_z(\tau_t^2) \bxi } = 0$. With this lemma, we now derive estimations for 
$\Q_1$, $\Q_{1, \bxi}$, $\Q_2$ and $Q_0$, respectively. 

\begin{lemma}[Estimations for $\Q_1$]
  \label{lemma: stage 1: estimations for Q1}
  In Stage~1, we have 
  \[
    \Q_1
    = \frac{2K}{1 + K} \Id_d \pm O\left( d \tau_t^2 \right) . 
  \]
\end{lemma}

\begin{lemma}[Estimations for $\Q_{1, \bxi}$ and $\Q_2$]
  \label{lemma: stage 1: estimations for Q1xi Q2}
  In Stage~1, we have 
  \[
    \max\braces{
      \norm{\Q_{1, \xi_\A}}_F, \norm{\Q_{1, \xi_\B}}_F, \norm{\Q_2}_F 
    }
    = \pm O\left( \tau_t^2 d^2 \rho_{N/S} \delta_{\xi, \perp} \right). 
  \] 
\end{lemma}

\begin{lemma}[Estimations for $Q_0$]
  \label{lemma: stage 1: estimations for Q0}
  In Stage~1, we have 
  \[
    Q_0 
    = - \frac{2 K}{1 + K} \frac{\inprod{\KA}{\KB}}{N_\A N_\B d} 
      \pm O\left( d \tau_t^2 \right). 
  \]
\end{lemma}

The proof of Lemma~\ref{lemma: stage 1: estimations for Q0} is essentially the same as the proof of 
Lemma~\ref{lemma: stage 1: estimations for Q1} so we omit it. With these three lemmas, we can now simplify 
Lemma~\ref{lemma: d norm and d bar} as follows. 

\begin{corollary}
  \label{cor: stage 1: d norm}
  In Stage~1, for any $p \in [r]$ and $q \in [d - r]$, we have 
  \begin{align*}
    \frac{\rd}{\rd t} \norm{[\KA]_p}^2
    &= \frac{4 K}{1 + K} 
      \frac{\sigma_p^2}{N_\A N_\B d} 
      \left(
        \frac{ \inprod{[\KA]_p}{[\KB]_p} }{ \norm{[\KA]_p}^2 }
        - \frac{\inprod{\KA}{\KB}}{N_\A N_\B d} 
      \right) 
      \norm{[\KA]_p}^2
      \\
      &\qquad
      \pm O\left( 
        \frac{ \sigma_p^2   }{N_\A N_\B d} 
        \left( \sqrt{\delta_{\A/\B}} + \tau_t^2 d \right)
        \kappa_p^2
      \right), \\
    \frac{\rd}{\rd t} \norm{[\KAxi]_q}^2
    &= - \frac{4 K}{1 + K}  
      \frac{\sigma_\xi^2}{N_\A N_\B d} 
      \frac{\inprod{\KA}{\KB}}{N_\A N_\B d} 
      \norm{[\KAxi]_q}^2 
      \pm O\left( 
          \frac{\sigma_\xi^2}{N_\A N_\B d} 
          \left( d \tau_t^2 + \sqrt{\delta_{\A/\B}} \right) 
          \norm{[\KAxi]_q}^2
        \right), \\
  \end{align*}
  The formulas for $\KB$ and $\KBxi$ can be obtained by interchanging the roles of $\A$ and $\B$. 
\end{corollary}

\begin{corollary}
  \label{cor: stage 1: d bar}
  In Stage~1, we have 
  \begin{align*}
    \frac{\rd}{\rd t} \ol{ [\KA]_p }
    &= \frac{2K}{1 + K} \frac{\sigma_p^2 }{N_\A N_\B d }
      \left( \Id - \ol{ [\KA]_p } \left( \ol{ [\KA]_p } \right)\trans  \right)
      \ol{[\KB]_p}
      \pm O\left(
        \frac{\sigma_p^2 }{N_\A N_\B d }
        \left( \tau_t^2 d^2 \kappa_0 + d \sqrt{\delta_{\A/\B}} \right)
      \right), \\
    \frac{\rd}{\rd t} \ol{[\KAxi]_q} 
    &= \pm O\left( 
        \frac{\sigma_\xi^2 }{N_\A N_\B d } 
        \tau_t^2 d^3 \kappa_0  \delta_{\xi, \perp}
      \right).
  \end{align*}
  Interchange the roles of $\A$ and $\B$ and one can obtain the formulas for $\KB$ and $\KBxi$. 
\end{corollary}

Note that the above results also imply the following lemma. 
\begin{lemma}
  \label{lemma: non-contrastive equiv stage 1}
  The dynamics of the non-contrastive method and the Stage~1 dynamics are equivalent, up to a multiplicative 
  constant and some higher order terms.
\end{lemma}

\subsubsection*{Omitted proofs of this subsection}

\begin{proof}[Proof of Lemma~\ref{lemma: stage 1: estimations for S}]
  First, we write 
  \[
    \inprod{\fA^+}{\fB^-}
    = \frac{ \inprod{\KA\z^+}{\KB\z^-} }{N_\A N_\B}
      + \frac{
          \inprod{\KA\z^+}{\KBxi\bxi_\B^-} + \inprod{\KAxi\bxi_\A^+}{\KB\z^-}
        }{N_\A N_\B}
      + \frac{ \inprod{\KAxi\bxi_\A^+}{\KBxi\bxi_\B^-} }{N_\A N_\B}. 
  \]
  For the second term, we write 
  \[
    \frac{ \inprod{\KA\z^+}{\KBxi\bxi_\B^-} }{N_\A N_\B}
    = \sum_{i \in [r], j \in [d-r]} 
      \frac{ \norm{[\KA]_i}^2 }{N_\A N_\B d} 
      \frac{\norm{[\KBxi]_j} }{\norm{[\KA]_i}}
      \inprod{\ol{[\KA]_i}}{\ol{[\KBxi]_j}}  z^+_i \xi_{\B, j}^- d.   
  \]
  For each summand, we have $\norm{[\KBxi]_j} / \norm{[\KA]_i} \le O(\rho_{N/S})$, 
  $\inprod{\ol{[\KA]_i}}{\ol{[\KBxi]_j}} = \pm O(\delta_{\xi, \perp})$, and $|z^+_i \xi_{\B, j}^- d| \le 1$. 
  Hence, $\frac{ \inprod{\KA\z^+}{\KBxi\bxi_\B^-} }{N_\A N_\B} = \pm O( d \delta_{\xi, \perp} \rho_{N/S} )$. 
  The same is also true for $\frac{ \inprod{\KAxi\bxi_\A^+}{\KB\z^-} }{ N_\A N_\B d }$ and the third term. 
  Therefore, 
  \[
    \inprod{\fA^+}{\fB^-}
    = \frac{ \inprod{\KA\z^+}{\KB\z^-} }{N_\A N_\B}
      \pm O\left( d \delta_{\xi, \perp} \rho_{N/S} \right). 
  \]
  Then, we compute 
  \begin{align*}
    \exp\left( \tau_t^2 \inprod{\fA^+}{\fB^-} \right)
    &= \exp\left( \tau_t^2 \frac{ \inprod{\KA\z^+}{\KB\z^-} }{N_\A N_\B} \right)
      \left( 1 \pm O\left(\tau_t^2 d \delta_{\xi, \perp} \rho_{N/S} \right) \right) \\
    &= 1 \pm O_z\left( \tau_t^2 \right)
      \pm O\left(\tau_t^2 d \delta_{\xi, \perp} \rho_{N/S} \right). 
  \end{align*}
  Thus, 
  \[
    S_\A 
    = \frac{1}{1 + K} 
      \pm O_z\left( \tau_t^2 \right)
      \pm O\left(\tau_t^2 d \delta_{\xi, \perp} \rho_{N/S} \right). 
  \]
  The proof for $S_\B$ is essentially the same. 
\end{proof}

\begin{proof}[Proof of Lemma~\ref{lemma: stage 1: estimations for Q1}]
  Recall that 
  \begin{align*}
    \Q_1
    &:= \E \braces{
        \left( 2 - S_\A(\xA^+, \xB^+) - S_\B(\xA^+, \xB^+) \right)
        \z^+(\z^+)\trans d
      } \\
      &\qquad
      - K \E \braces{
        \frac{ S_\A(\xA^+, \xB^+) \exp(\tau_t^2 \fA^+ \cdot \fB^-)}{\exp(\tau_t^2 \fA^+ \cdot \fB^+)}
        \z^- (\z^+)\trans d
      } \\ 
      &\qquad
      - K \E \braces{
        \frac{S_\B(\xA^+, \xB^+) \exp(\tau_t^2 \fA^- \cdot \fB^+)}{\exp(\tau_t^2 \fA^+ \cdot \fB^+)}
        \z^+ (\z^-)\trans d
      }. 
  \end{align*}
  Since $S_\A = (1 + K)\inv \pm O(\tau_t^2)$ and $S_\B = (1 + K)\inv \pm O(\tau_t^2)$, we have 
  \[
    2 - S_\A(\xA^+, \xB^+) - S_\B(\xA^+, \xB^+)
    = 2 - \frac{2}{1 + K} \pm O(\tau_t^2) 
    = \frac{2 K}{1 + K} \pm O(\tau_t^2) .
  \]
  As a result, 
  \[
    \Q_1
    = \frac{2K}{1 + K} \E \braces{ \z^+(\z^+)\trans d }
      - \frac{K}{1 + K} \E \braces{ \z^- (\z^+)\trans d } 
      - \frac{K}{1 + K} \E \braces{ \z^+ (\z^-)\trans d }
      \pm O\left( d \tau_t^2 \right) . 
  \]
  Note that $\E\braces{\z^- (\z^+)\trans} = 0$ and $\E \braces{ \z^+(\z^+)\trans } = \Id_d / d$. Therefore, 
  \[
    \Q_1
    = \frac{2K}{1 + K} \Id_d \pm O\left( d \tau_t^2 \right) . 
  \]
\end{proof}

\begin{proof}[Proof of Lemma~\ref{lemma: stage 1: estimations for Q1xi Q2}]
  Recall that 
  \begin{align*}
    \Q_{1, \xi_\A}
    &:= \E \braces{
        \left( 2 - S_\A(\xA^+, \xB^+) - S_\B(\xA^+, \xB^+) \right)
        \bxi_\A^+(\z^+)\trans d
      } \\
      &\qquad
      - K \E \braces{
        \frac{ S_\A(\xA^+, \xB^+) \exp(\tau_t^2 \fA^+ \cdot \fB^-)}{\exp(\tau_t^2 \fA^+ \cdot \fB^+)}
        \bxi_\A^+ (\z^-)\trans d
      } \\ 
      &\qquad
      - K \E \braces{
        \frac{S_\B(\xA^+, \xB^+) \exp(\tau_t^2 \fA^- \cdot \fB^+)}{\exp(\tau_t^2 \fA^+ \cdot \fB^+)}
        \bxi_\A^- (\z^+)\trans d
      }. 
  \end{align*}
  Note that if some quantity $X$ does not depend on $\bxi$, then $\E\braces{ X \bxi } = 0$. Hence, by 
  Lemma~\ref{lemma: stage 1: estimations for S}, we have 
  \[
    \Q_{1, \xi_\A}
    = \pm O\left( \tau_t^2 d^2 \rho_{N/S} \delta_{\xi, \perp} \right). 
  \] 
  The proof for $\Q_{1, \xi_\A}$ and $\Q_2$ is essentially the same.
\end{proof}

\begin{proof}[Proof of Corollary~\ref{cor: stage 1: d norm}]
  Recall that 
  \begin{align*}
    \frac{\rd}{\rd t} \norm{[\KA]_p}^2
    &= 2 \frac{\inprod{[\KA]_p}{[\KB\Q_1]_p}}{N_\A N_\B d} \sigma_p^2
      + 2 \frac{\inprod{[\KA]_p}{[\KBxi\Q_{1, \xi_\B}]_p}}{N_\A N_\B d}  \sigma_p^2
      + 2 \frac{\norm{[\KA]_p}^2}{N_\A^2 d} Q_0 \sigma_p^2  \\
    &= \sum_{i=1}^3 \Term_i\left( \frac{\rd}{\rd t} \norm{[\KA]_p}^2 \right). 
  \end{align*}
  We now estimate these terms one-by-one. For $\Term_1$, we have 
  \begin{align*}
    \inprod{[\KA]_p}{[\KB\Q_1]_p}
    &= \inprod{[\KA]_p}{[\KB]_p} [\Q_1]_{p, p}
      + \sum_{k \ne p} \inprod{[\KA]_p}{[\KB]_k} [\Q_1]_{k, p}  \\
    &= \frac{2 K}{1 + K} \inprod{[\KA]_p}{[\KB]_p} \pm O\left( \tau_t^2 d \kappa_p^2  \right)
      \pm O\left( \tau_t^2 d \kappa_p^2 \kappa_0 \delta_{\AB, \perp} \right) \\
    &= \frac{2 K}{1 + K} \inprod{[\KA]_p}{[\KB]_p} \pm O\left( \tau_t^2 d \kappa_p^2  \right). 
  \end{align*}
  Hence, 
  \[
    \Term_1\left( \frac{\rd}{\rd t} \norm{[\KA]_p}^2 \right)
    = \frac{4 K}{1 + K} 
      \frac{\sigma_p^2}{N_\A N_\B d} 
      \inprod{[\KA]_p}{[\KB]_p} 
      \pm O\left( 
        \frac{\sigma_p^2}{N_\A N_\B d} 
        \tau_t^2 d \kappa_p^2  
      \right). 
  \]
  For $\Term_2$, we compute 
  \begin{align*}
    \Term_2\left( \frac{\rd}{\rd t} \norm{[\KA]_p}^2 \right)
    &= \frac{2 \sigma_p^2}{N_\A N_\B d} 
      \sum_{k=1}^{d-r} \inprod{[\KA]_p}{[\KBxi]_k } [\Q_{1, \xi_\B}]_{k, p}  \\
    &= \pm O(1) \frac{\sigma_p^2}{N_\A N_\B d} 
      \sum_{k=1}^{d-r} \norm{[\KA]_p}^2 \frac{ \norm{[\KBxi]_k } }{ \norm{[\KA]_p} }
        \inprod{\ol{[\KA]_p}}{\ol{[\KBxi]_k}} [\Q_{1, \xi_\B}]_{k, p}  \\
    &= \pm O\left(
        \frac{\sigma_p^2}{N_\A N_\B d} 
        \tau_t^2 d^3
        \rho_{N/S}^2 \delta_{\xi, \perp}^2  
        \kappa_p^2
      \right). 
  \end{align*}
  Finally, for $\Term_3$, we compute 
  \begin{align*}
    \Term_3\left( \frac{\rd}{\rd t} \norm{[\KA]_p}^2 \right)
    &= 2 \frac{\norm{[\KA]_p}^2}{N_\A^2 d} 
      \left(  
        - \frac{2 K}{1 + K} \frac{\inprod{\KA}{\KB}}{N_\A N_\B d} 
        \pm O\left( d \tau_t^2 \right)
      \right) 
      \sigma_p^2   \\
    &= - \frac{4 K}{1 + K} 
      \frac{\sigma_p^2}{N_\A N_\B d} 
      \left( 1 \pm \sqrt{\delta_{\A/\B}} \right) 
      \frac{\inprod{\KA}{\KB}}{N_\A N_\B d} 
      \norm{[\KA]_p}^2
      \pm O\left( 
        \frac{ \kappa_p^2 }{N_\A N_\B d} 
        \tau_t^2 d \sigma_p^2  
      \right) \\
    &= - \frac{4 K}{1 + K} 
      \frac{\sigma_p^2}{N_\A N_\B d} 
      \frac{\inprod{\KA}{\KB}}{N_\A N_\B d} 
      \pm O\left( 
        \frac{ \sigma_p^2   }{N_\A N_\B d} 
        \left( \sqrt{\delta_{\A/\B}} + \tau_t^2 d \right)
        \kappa_p^2
      \right). 
  \end{align*}
  Combine these together and we obtain 
  \begin{align*}
    \frac{\rd}{\rd t} \norm{[\KA]_p}^2
    &= \frac{4 K}{1 + K} 
      \frac{\sigma_p^2}{N_\A N_\B d} 
      \inprod{[\KA]_p}{[\KB]_p} 
      - \frac{4 K}{1 + K} 
      \frac{\sigma_p^2}{N_\A N_\B d} 
      \frac{\inprod{\KA}{\KB}}{N_\A N_\B d} 
      \norm{[\KA]_p}^2 \\
      &\qquad
      \pm O\left(
        \frac{\sigma_p^2}{N_\A N_\B d} 
        \tau_t^2 d^3
        \rho_{N/S}^2 \delta_{\xi, \perp}^2  
        \kappa_p^2
      \right) 
      \pm O\left( 
        \frac{ \sigma_p^2   }{N_\A N_\B d} 
        \left( \sqrt{\delta_{\A/\B}} + \tau_t^2 d \right)
        \kappa_p^2
      \right) \\
    &= \frac{4 K}{1 + K} 
      \frac{\sigma_p^2}{N_\A N_\B d} 
      \left(
        \frac{ \inprod{[\KA]_p}{[\KB]_p} }{ \norm{[\KA]_p}^2 }
        - \frac{\inprod{\KA}{\KB}}{N_\A N_\B d} 
      \right) 
      \norm{[\KA]_p}^2
      \\
      &\qquad
      \pm O\left( 
        \frac{ \sigma_p^2   }{N_\A N_\B d} 
        \left( \sqrt{\delta_{\A/\B}} + \tau_t^2 d \right)
        \kappa_p^2
      \right). 
  \end{align*}
  Interchange the roles of $\A$ and $\B$ and we obtain the formula for $\frac{\rd}{\rd t} \norm{[\KB]_p}^2$. 
  Now we consider $\KAxi$. We write 
  \begin{align*}
    \frac{\rd}{\rd t} \norm{[\KAxi]_q}^2
    &= 2 \frac{ \inprod{[\KAxi]_q}{[\KB \Q_{1, \bxi_\A}\trans]_q} }{ N_\A N_\B d}  \sigma_\xi^2
      + 2 \frac{ \inprod{[\KAxi]_q}{[\KBxi \Q_2]_q} }{N_\A N_\B d} \sigma_\xi^2
      + 2 \frac{\norm{[\KAxi]_q}_F^2}{N_\A^2 d} Q_0 \sigma_\xi^2 \\
    &= \sum_{i=1}^3 \Term_i\left( \frac{\rd}{\rd t} \norm{[\KAxi]_q}^2 \right).
  \end{align*}
  For $\Term_1$, we compute 
  \begin{align*}
    \Term_1\left( \frac{\rd}{\rd t} \norm{[\KAxi]_q}^2 \right)
    &= 2 \sum_{k=1}^r \frac{ \inprod{[\KAxi]_q}{[\KB]_k} [\Q_{1, \bxi_\A}]_{q, k}}{ N_\A N_\B d}  \sigma_\xi^2 \\
    &= \frac{2 \sigma_\xi^2}{N_\A N_\B d}
      \sum_{k=1}^r \norm{[\KAxi]_q}^2 \frac{ \norm{[\KB]_k} }{\norm{[\KAxi]_q}}
        \inprod{\ol{[\KAxi]_q}}{\ol{[\KB]_k}} [\Q_{1, \bxi_\A}]_{q, k}   \\
    &= \pm O(1) \frac{2 \sigma_\xi^2}{N_\A N_\B d} \norm{[\KAxi]_q}^2
      \sum_{k=1}^r  \frac{\kappa_0}{\rho_{N/S}}
        \delta_{\xi, \perp} \tau_t^2 d^2 \rho_{N/S} \delta_{\xi, \perp}   \\
    &= \pm O\left(
        \frac{\sigma_\xi^2}{N_\A N_\B d} 
        \tau_t^2 d^3 \kappa_0 \delta_{\xi, \perp}^2
        \norm{[\KAxi]_q}^2
      \right). 
  \end{align*}
  Similarly, one can show that this bound also holds for $\Term_2$. Finally, for $\Term_3$, by 
  Lemma~\ref{lemma: stage 1: estimations for Q0}, we have 
  \begin{align*}
    \Term_3\left( \frac{\rd}{\rd t} \norm{[\KAxi]_q}^2 \right)
    &= 2 \frac{\norm{[\KAxi]_q}_F^2}{N_\A N_\B d} 
      \left( 1 \pm \sqrt{\delta_{\A/\B}} \right)
      \left(
        - \frac{2 K}{1 + K} \frac{\inprod{\KA}{\KB}}{N_\A N_\B d} 
        \pm O\left( d \tau_t^2 \right)
      \right)
      \sigma_\xi^2 \\
    &= - \frac{4 K}{1 + K}  
      \frac{\sigma_\xi^2}{N_\A N_\B d} 
      \frac{\inprod{\KA}{\KB}}{N_\A N_\B d} 
      \norm{[\KAxi]_q}_F^2 \\
      &\qquad
      \pm O\left( 
          \frac{\sigma_\xi^2}{N_\A N_\B d} 
          \left( d \tau_t^2 + \sqrt{\delta_{\A/\B}} \right) 
          \norm{[\KAxi]_q}_F^2
        \right). 
  \end{align*}
  Combine these together and we obtain 
  \[ 
    \frac{\rd}{\rd t} \norm{[\KAxi]_q}^2
    = - \frac{4 K}{1 + K}  
      \frac{\sigma_\xi^2}{N_\A N_\B d} 
      \frac{\inprod{\KA}{\KB}}{N_\A N_\B d} 
      \norm{[\KAxi]_q}_F^2 
      \pm O\left( 
          \frac{\sigma_\xi^2}{N_\A N_\B d} 
          \left( d \tau_t^2 + \sqrt{\delta_{\A/\B}} \right) 
          \norm{[\KAxi]_q}_F^2
        \right). 
  \] 
  Interchange the roles of $\A$ and $\B$ and we obtain the formula for $\KBxi$. 
\end{proof}

\begin{proof}[Proof of Corollary~\ref{cor: stage 1: d bar}]
  We write 
  \begin{align*}
    \frac{\rd}{\rd t} \ol{ [\KA]_p }
    &= \left( \Id - \ol{ [\KA]_p } \left( \ol{ [\KA]_p } \right)\trans  \right)
      \frac{[\KB \Q_1]_p }{ \norm{[\KA]_p} }
      \frac{\sigma_p^2 }{N_\A N_\B d } \\
      &\qquad
      + \left( \Id - \ol{ [\KA]_p } \left( \ol{ [\KA]_p } \right)\trans  \right)
      \frac{[\KBxi \Q_{1, \xi_\B}]_p}{\norm{[\KA]_p} }
      \frac{\sigma_p^2 }{N_\A N_\B d } \\
    &= \Term_1\left( \frac{\rd}{\rd t} \ol{ [\KA]_p } \right) 
      + \Term_2\left( \frac{\rd}{\rd t} \ol{ [\KA]_p } \right). 
  \end{align*}
  For $\Term_1$, we have 
  \begin{align*}
    \Term_1\left( \frac{\rd}{\rd t} \ol{ [\KA]_p } \right) 
    &= \sum_{k=1}^r 
      \left( \Id - \ol{ [\KA]_p } \left( \ol{ [\KA]_p } \right)\trans  \right)
      \ol{[\KB]_k}
      \frac{\norm{[\KB]_k} [\Q_1]_{k, p} }{ \norm{[\KA]_p} }
      \frac{\sigma_p^2 }{N_\A N_\B d }  \\
    &= \left( \Id - \ol{ [\KA]_p } \left( \ol{ [\KA]_p } \right)\trans  \right)
      \ol{[\KB]_p}
      \frac{\norm{[\KB]_p} [\Q_1]_{p, p} }{ \norm{[\KA]_p} }
      \frac{\sigma_p^2 }{N_\A N_\B d } \\
      &\qquad
      + \sum_{k \ne p} 
      \left( \Id - \ol{ [\KA]_p } \left( \ol{ [\KA]_p } \right)\trans  \right)
      \ol{[\KB]_k}
      \frac{\norm{[\KB]_k} [\Q_1]_{k, p} }{ \norm{[\KA]_p} }
      \frac{\sigma_p^2 }{N_\A N_\B d }. 
  \end{align*}
  For the first term, by Lemma~\ref{lemma: stage 1: estimations for Q1}, we have 
  \[
    \frac{\norm{[\KB]_p} [\Q_1]_{p, p} }{ \norm{[\KA]_p} }
    = \left( 1 \pm \sqrt{\delta_{\A/\B}} \right) 
      \left( \frac{2K}{1 + K} \pm O\left( d \tau_t^2 \right) \right)
    = \frac{2K}{1 + K}
      \pm O\left( d \tau_t^2 + \sqrt{\delta_{\A/\B}} \right). 
  \]
  Also by Lemma~\ref{lemma: stage 1: estimations for Q1}, for each summand in the second term, we have 
  \[
    \frac{\norm{[\KB]_k} [\Q_1]_{k, p} }{ \norm{[\KA]_p} }
    = \pm O\left( \tau_t^2 d \kappa_0 \right). 
  \] 
  Therefore, 
  \begin{align*}
    \Term_1\left( \frac{\rd}{\rd t} \ol{ [\KA]_p } \right) 
    &= \frac{2K}{1 + K} \frac{\sigma_p^2 }{N_\A N_\B d }
      \left( \Id - \ol{ [\KA]_p } \left( \ol{ [\KA]_p } \right)\trans  \right)
      \ol{[\KB]_p} \\
      &\qquad
      \pm \frac{\sigma_p^2 }{N_\A N_\B d }
      \sum_{k=1}^r  
      O\left( \tau_t^2 d \kappa_0 + \sqrt{\delta_{\A/\B}} \right)
      \ol{[\KB]_k}. 
  \end{align*}
  For $\Term_2$, by Lemma~\ref{lemma: stage 1: estimations for Q1xi Q2}, we have 
  \begin{align*}
    \Term_2\left( \frac{\rd}{\rd t} \ol{ [\KA]_p } \right) 
    &= \frac{\sigma_p^2 }{N_\A N_\B d }
      \sum_{k=1}^{d-r}
      \left( \Id - \ol{ [\KA]_p } \left( \ol{ [\KA]_p } \right)\trans  \right)
      \ol{[\KBxi]_k}
      \frac{\norm{[\KBxi]_k} [\Q_{1, \xi_\B}]_{k, p}}{\norm{[\KA]_p} } \\ 
    &= \frac{\sigma_p^2 }{N_\A N_\B d }
      \sum_{k=1}^{d-r}
        O\left( \tau_t^2 d^2 \rho_{N/S}^2 \delta_{\xi, \perp} \right)
        \ol{[\KBxi]_k}. 
  \end{align*}
  Combine these together, and we obtain 
  \begin{align*}
    \frac{\rd}{\rd t} \ol{ [\KA]_p }
    &= \frac{2K}{1 + K} \frac{\sigma_p^2 }{N_\A N_\B d }
      \left( \Id - \ol{ [\KA]_p } \left( \ol{ [\KA]_p } \right)\trans  \right)
      \ol{[\KB]_p} \\
      &\qquad
      \pm \frac{\sigma_p^2 }{N_\A N_\B d }
      \sum_{k=1}^r  
      O\left( \tau_t^2 d \kappa_0 + \sqrt{\delta_{\A/\B}} \right)
      \ol{[\KB]_k}
      \pm \frac{\sigma_p^2 }{N_\A N_\B d }
      \sum_{k=1}^{d-r}
        O\left( \tau_t^2 d^2 \rho_{N/S}^2 \delta_{\xi, \perp} \right)
        \ol{[\KBxi]_k} \\ 
    &= \frac{2K}{1 + K} \frac{\sigma_p^2 }{N_\A N_\B d }
      \left( \Id - \ol{ [\KA]_p } \left( \ol{ [\KA]_p } \right)\trans  \right)
      \ol{[\KB]_p}
      \pm O\left(
        \frac{\sigma_p^2 }{N_\A N_\B d }
        \left( \tau_t^2 d^2 \kappa_0 + d \sqrt{\delta_{\A/\B}} \right)
      \right). 
  \end{align*}
  Now, we consider $\KAxi$. Again, we write 
  \begin{align*}
    \frac{\rd}{\rd t} \ol{[\KAxi]_q}
    &= \left( \Id - \ol{ [\KAxi]_q } \left( \ol{ [\KAxi]_q } \right)\trans \right)
      \frac{ [\KB \Q_{1, \bxi_\A}\trans]_q }{ \norm{[\KAxi]_q} }
      \frac{\sigma_\xi^2 }{N_\A N_\B d } \\
      &\qquad
      + \left( \Id - \ol{ [\KAxi]_q } \left( \ol{ [\KAxi]_q } \right)\trans \right)
        \frac{ [\KBxi \Q_2]_q }{ \norm{[\KAxi]_q} }
        \frac{\sigma_\xi^2 }{N_\A N_\B d } \\
    &=: \Term_1\left( \frac{\rd}{\rd t} \ol{[\KAxi]_q} \right)
      + \Term_2\left( \frac{\rd}{\rd t} \ol{[\KAxi]_q} \right). 
  \end{align*}
  For the first term, by Lemma~\ref{lemma: stage 1: estimations for Q1xi Q2}, we have
  \begin{align*}
    \Term_1\left( \frac{\rd}{\rd t} \ol{[\KAxi]_q} \right)
    &= \sum_{k=1}^r
      \left( \Id - \ol{ [\KAxi]_q } \left( \ol{ [\KAxi]_q } \right)\trans \right)
      \ol{[\KB]_k}
      \frac{ \norm{[\KB]_k} [\Q_{1, \bxi_\A}]_{q, k} }{ \norm{[\KAxi]_q} }
      \frac{\sigma_\xi^2 }{N_\A N_\B d } \\
    &= \pm \sum_{k=1}^r
      O\left( 
        \frac{\sigma_\xi^2 }{N_\A N_\B d } 
        \frac{\kappa_0}{\rho_{N/S}}  
        \tau_t^2 d^2 \rho_{N/S} \delta_{\xi, \perp}
      \right) \\
    &= \pm O\left( 
        \frac{\sigma_\xi^2 }{N_\A N_\B d } 
        \tau_t^2 d^3 \kappa_0  \delta_{\xi, \perp}
      \right). 
  \end{align*}
  The same bound also hold for $\Term_2$. In fact, we can have a slightly sharper bound for it because 
  we no longer have $\norm{[\KB]_k} / \norm{[\KAxi]_q}$. Combine these and we obtain 
  \[
    \frac{\rd}{\rd t} \ol{[\KAxi]_q} 
    = \pm O\left( 
        \frac{\sigma_\xi^2 }{N_\A N_\B d } 
        \tau_t^2 d^3 \kappa_0  \delta_{\xi, \perp}
      \right).
  \]
\end{proof}

\begin{proof}[Proof of Lemma~\ref{lemma: non-contrastive equiv stage 1}]
  The proof of Corollary~\ref{cor: stage 1: d bar} and Corollary~\ref{cor: stage 1: d norm}, 
  \textit{mutatis mutandis}, yields
  \begin{align*}
    \frac{\rd}{\rd t} \KA 
    &= \frac{2K}{1 + K} \frac{\KB}{N_\A N_\B d} \diag(\bsigma^2)  
      - \frac{2 K}{1 + K} \frac{\KA}{N_\A^2 d} \frac{\inprod{\KA}{\KB}}{N_\A N_\B d} \diag(\bsigma^2)  \\
      &\qquad
      \pm O\left( 
          \frac{\sigma_{\max}^2}{N_\A N_\B d} 
          \left( d \tau_t^2 + \sqrt{\delta_{\A/\B}} \right) 
          \norm{\KA}_F
        \right), \\
    \frac{\rd}{\rd t} \KAxi
    &= - \frac{2 K}{1 + K} \frac{\KAxi}{N_\A^2 d}
      \frac{\inprod{\KA}{\KB}}{N_\A N_\B d} 
      \sigma_\xi^2 
      \pm O\left( 
        \frac{\sigma_\xi^2 }{N_\A^2 d} d \tau_t^2 \norm{\KAxi}_F
      \right). 
  \end{align*}
  Recall from Lemma~\ref{lemma: grad non-contrastive} that, in the non-contrastive case, we have 
  \begin{align*}
    \frac{\rd}{\rd t} \KA 
    &= \E\braces{
        \frac{ \fB^+(\z^+)\trans}{N_\A} 
        - \inprod{\fA^+}{\fB^+} \frac{\KA}{ N_\A^2 d } 
      } \diag(\bsigma^2) \\
    &= \frac{ \KB }{N_\A N_\B d} \diag(\bsigma^2)
      - \frac{ \inprod{\KA}{\KB} }{ N_\A N_\B d } \frac{\KA}{ N_\A^2 d } 
        \diag(\bsigma^2), \\
    \frac{\rd}{\rd t} \KAxi
    &=  - \frac{ \inprod{\KA}{\KB} }{ N_\A N_\B d } \frac{\KAxi}{ N_\A^2 d } \sigma_\xi^2.  
  \end{align*}
  Note that they are exactly the same, except for a $2 K / (1 + K)$ factor and some higher order error terms. 
\end{proof}

\subsection{Convergence rate and the condition number} 
\label{sec: stage 1: convergence, condition number}

In this subsection, we estimate the rate at which $1 - \rho_-$ and $\rho_{N/S}$ converge to $0$ and the 
growth rate of the condition number.

\begin{lemma}[Convergence rate of $\rho_-$]
  \label{lemma: stage 1: convergence rate of rho-}
  In Stage~1, we have, for any $p \in [r]$, 
  \begin{align*}
    \frac{\rd}{\rd t}  \inprod{\ol{[\KA]_p}}{ \ol{[\KB]_p} }
    &= \frac{4 K}{1 + K} \frac{\sigma_p^2 }{N_\A N_\B d }
      \left( 1 + \inprod{\ol{[\KA]_p}}{ \ol{[\KB]_p} } \right)
      \left( 1 - \inprod{\ol{[\KA]_p}}{ \ol{[\KB]_p} } \right) \\ 
      &\qquad
      \pm O\left(
        \frac{\sigma_p^2 }{N_\A N_\B d }
        \left( \tau_t^2 d^2 \kappa_0 + \sqrt{\delta_{\A/\B}} \right)
      \right). 
  \end{align*}
\end{lemma}

Now we consider the noise-signal ratio. For some technical reason, instead of characterizing the dynamics of 
$\rho_{N/S}$, we consider 
\[
  \hat{\rho}_{N/S}
  := \frac{\norm{\KAxi}_F}{\norm{\KA + \KB}_F}. 
\]
Note that we always have 
\[
  \hat{\rho}_{N/S}^2 
  \ge \Theta(1) \frac{\norm{\KAxi}_F^2}{\norm{\KA}_F^2}
  \ge \Theta(1) \frac{ (d - r) \norm{[\KAxi]_q}^2}{\kappa_0^2 r \norm{[\KA]_p}^2}
\]
In other words, $\rho_{N/S}$ can be bounded by $\rho_{N/S} \le O\left( \frac{d \kappa_0}{r} \hat{\rho}_{N/S} \right)$. 

\begin{lemma}[Convergence rate of $\rho_{N/S}$]
  \label{lemma: stage 1: convergence rate of rhoNS}
  In Stage~1, we have 
  \[ 
    \frac{\rd}{\rd t} \hat{\rho}_{N/S}^2
    \le 
      - \frac{4 K}{1 + K} \frac{\sigma_{\min}^2}{N_\A N_\B d}
      \hat{\rho}_{N/S}^2
      + O\left( 
        \frac{\sigma_{\max}^2}{N_\A N_\B d} 
        \left( d \tau_t^2 + \sqrt{\delta_{\A/\B}} \right) 
        \hat{\rho}_{N/S}^2 
      \right). 
  \] 
\end{lemma}

\begin{lemma}[Growth rate of the condition number]
  \label{lemma: stage 1: growth rate of the condition number}
  Define $\rho_{p/q} = \norm{[\KA]_p}^2 / \norm{[\KA]_q}^2$. In Stage~1, we have 
  \[
    \dot{\rho}_{p/q}
    \le \frac{16 K}{1 + K} 
      \frac{\sigma_{\max}^2}{N_\A N_\B d} 
      \left( \rho_- + \min\braces{ \hat{\rho}_{N/S}, 1} \right) 
      \rho_{p/q} 
      \pm O\left( 
        \frac{ \sigma_{\max}^2   }{N_\A N_\B d} 
        \left( \sqrt{\delta_{\A/\B}} + \tau_t^2 d \right)
        \rho_{p/q}
      \right). 
  \]
\end{lemma}

\subsubsection*{Omitted proofs of this subsection}

\begin{proof}[Proof of Lemma~\ref{lemma: stage 1: convergence rate of rho-}]
  By Corollary~\ref{cor: stage 1: d bar}, we have 
  \begin{align*}
    \inprod{\frac{\rd}{\rd t} \ol{[\KA]_p}}{ \ol{[\KB]_p} }
    &= \frac{2K}{1 + K} \frac{\sigma_p^2 }{N_\A N_\B d }
      \left( 1 - \inprod{\ol{[\KA]_p}}{ \ol{[\KB]_p} }^2 \right)
      \pm O\left(
        \frac{\sigma_p^2 }{N_\A N_\B d }
        \left( \tau_t^2 d^2 \kappa_0 + \sqrt{\delta_{\A/\B}} \right)
      \right) \\ 
    &= \frac{2K}{1 + K} \frac{\sigma_p^2 }{N_\A N_\B d }
    \left( 1 + \inprod{\ol{[\KA]_p}}{ \ol{[\KB]_p} } \right)
    \left( 1 - \inprod{\ol{[\KA]_p}}{ \ol{[\KB]_p} } \right) \\ 
    &\qquad
    \pm O\left(
      \frac{\sigma_p^2 }{N_\A N_\B d }
      \left( \tau_t^2 d^2 \kappa_0 + \sqrt{\delta_{\A/\B}} \right)
    \right). 
  \end{align*}
  Hence, by symmetry, we have 
  \begin{align*}
    \frac{\rd}{\rd t}  \inprod{\ol{[\KA]_p}}{ \ol{[\KB]_p} }
    &= \frac{4 K}{1 + K} \frac{\sigma_p^2 }{N_\A N_\B d }
      \left( 1 + \inprod{\ol{[\KA]_p}}{ \ol{[\KB]_p} } \right)
      \left( 1 - \inprod{\ol{[\KA]_p}}{ \ol{[\KB]_p} } \right) \\ 
      &\qquad
      \pm O\left(
        \frac{\sigma_p^2 }{N_\A N_\B d }
        \left( \tau_t^2 d^2 \kappa_0 + \sqrt{\delta_{\A/\B}} \right)
      \right). 
  \end{align*}
\end{proof}

\begin{proof}[Proof of Lemma~\ref{lemma: stage 1: convergence rate of rhoNS}]
  Similar to the proof of Corollary~\ref{cor: stage 1: d norm} and Corollary~\ref{cor: stage 1: d bar}, we have 
  \begin{align*}
    \frac{\rd}{\rd t} \KA 
    &= \frac{\KB}{N_\A N_\B d} \Q_1 \diag(\bsigma^2) 
      + \frac{\KBxi}{N_\A N_\B d} \Q_{1, \xi_\B} \diag(\bsigma^2) 
      + \frac{\KA}{N_\A^2 d} Q_0 \diag(\bsigma^2) \\
    &= \frac{\KB}{N_\A N_\B d} 
      \left( \frac{2K}{1 + K} \Id_d \pm O\left( d \tau_t^2 \right) \right) \diag(\bsigma^2)  \\
      &\qquad
      \pm \frac{\KBxi}{N_\A N_\B d} 
        O\left( \tau_t^2 d^2 \rho_{N/S} \delta_{\xi, \perp} \right)  \diag(\bsigma^2)  \\
      &\qquad
      + \frac{\KA}{N_\A^2 d} 
      \left( - \frac{2 K}{1 + K} \frac{\inprod{\KA}{\KB}}{N_\A N_\B d} \pm O\left( d \tau_t^2 \right) \right)
      \diag(\bsigma^2) \\
    &= \frac{2K}{1 + K} \frac{\KB}{N_\A N_\B d} \diag(\bsigma^2)  
      - \frac{2 K}{1 + K} \frac{\KA}{N_\A N_\B d} \frac{\inprod{\KA}{\KB}}{N_\A N_\B d} \diag(\bsigma^2)  \\
      &\qquad
      \pm O\left( 
          \frac{\sigma_{\max}^2}{N_\A N_\B d} 
          \left( d \tau_t^2 + \sqrt{\delta_{\A/\B}} \right) 
          \norm{\KA}_F
        \right). 
  \end{align*}
  Define $\K = \KA + \KB$. Then, by symmetry, we have 
  \[ 
    \frac{\rd}{\rd t} \K 
    = \frac{2K}{1 + K} \frac{1}{N_\A N_\B d} 
      \left(
        1 -  \frac{\inprod{\KA}{\KB}}{N_\A N_\B d} 
      \right)
      \K \diag(\bsigma^2)  
      \pm O\left( 
          \frac{\sigma_{\max}^2}{N_\A N_\B d} 
          \left( d \tau_t^2 + \sqrt{\delta_{\A/\B}} \right) 
          \norm{\K}_F
        \right). 
  \]
  Hence, 
  \begin{align*}
    \frac{\rd}{\rd t} \norm{\K}_F^2
    &= \frac{4 K}{1 + K} \frac{1}{N_\A N_\B d} 
      \left(
        1 -  \frac{\inprod{\KA}{\KB}}{N_\A N_\B d} 
      \right)
      \inprod{\K}{\K \diag(\bsigma^2)}
      \pm O\left( 
          \frac{\sigma_{\max}^2}{N_\A N_\B d} 
          \left( d \tau_t^2 + \sqrt{\delta_{\A/\B}} \right) 
          \norm{\K}_F^2
        \right) \\
    &\ge 
      \frac{4 K}{1 + K} \frac{\sigma_{\min}^2}{N_\A N_\B d} 
      \left(
        1 -  \frac{\inprod{\KA}{\KB}}{N_\A N_\B d} 
      \right)
      \norm{\K}_F^2 
      - O\left( 
        \frac{\sigma_{\max}^2}{N_\A N_\B d} 
        \left( d \tau_t^2 + \sqrt{\delta_{\A/\B}} \right) 
        \norm{\K}_F^2
      \right). 
  \end{align*}
  Similarly, we also have 
  \begin{align*}
    \frac{\rd}{\rd t} \KAxi
    &= \frac{ \KB }{ N_\A N_\B d} \Q_{1, \bxi_\A}\trans \sigma_\xi^2
      + \frac{\KBxi}{N_\A N_\B d} \Q_2 \sigma_\xi^2
      + \frac{\KAxi}{N_\A^2 d} Q_0 \sigma_\xi^2 \\
    &= \frac{ \KB }{ N_\A N_\B d} O\left( \tau_t^2 d^2 \rho_{N/S} \delta_{\xi, \perp} \right) \sigma_\xi^2
      + \frac{\KBxi}{N_\A N_\B d} O\left( \tau_t^2 d^2 \rho_{N/S} \delta_{\xi, \perp} \right) \sigma_\xi^2 \\
      &\qquad
      + \frac{\KAxi}{N_\A^2 d}
        \left( - \frac{2 K}{1 + K} \frac{\inprod{\KA}{\KB}}{N_\A N_\B d} \pm O\left( d \tau_t^2 \right) \right) 
        \sigma_\xi^2 \\
    &= - \frac{2 K}{1 + K} \frac{\KAxi}{N_\A^2 d}
      \frac{\inprod{\KA}{\KB}}{N_\A N_\B d} 
      \sigma_\xi^2 
      \pm O\left( 
        \frac{\sigma_\xi^2 }{N_\A^2 d} d \tau_t^2 \norm{\KAxi}_F
      \right)
      \pm O\left( 
        \frac{ \sigma_\xi^2 }{ N_\A N_\B d} 
        \tau_t^2 d^2 \delta_{\xi, \perp} 
        \frac{d \kappa_0}{r} \norm{\KAxi}_F
      \right) \\
    &= - \frac{2 K}{1 + K} \frac{\KAxi}{N_\A^2 d}
      \frac{\inprod{\KA}{\KB}}{N_\A N_\B d} 
      \sigma_\xi^2 
      \pm O\left( 
        \frac{\sigma_\xi^2 }{N_\A^2 d} d \tau_t^2 \norm{\KAxi}_F
      \right). 
  \end{align*}
  Therefore, 
  \[
    \frac{\rd}{\rd t} \norm{\KAxi}_F^2 
    = - \frac{4 K}{1 + K} \frac{\sigma_\xi^2}{N_\A^2 d}
      \frac{\inprod{\KA}{\KB}}{N_\A N_\B d}  
      \norm{\KAxi}_F^2
      \pm O\left( 
        \frac{\sigma_\xi^2 }{N_\A^2 d} d \tau_t^2 \norm{\KAxi}_F^2
      \right). 
  \]
  Then, we compute 
  \begin{align*}
    \frac{\rd}{\rd t} \hat{\rho}_{N/S}^2
    &= \frac{\frac{\rd}{\rd t} \norm{\KAxi}_F^2}{\norm{\K}_F^2}
      - \hat{\rho}_{N/S}^2 \frac{\frac{\rd}{\rd t} \norm{\KA}_F^2}{\norm{\K}_F^2} \\ 
    &\le 
      - \frac{4 K}{1 + K} \frac{\sigma_\xi^2}{N_\A^2 d}
      \frac{\inprod{\KA}{\KB}}{N_\A N_\B d}  
      \hat{\rho}_{N/S}^2
      \pm O\left( 
        \frac{\sigma_\xi^2 }{N_\A^2 d} d \tau_t^2 \hat{\rho}_{N/S}^2
      \right) \\ 
      &\qquad
      - \hat{\rho}_{N/S}^2 
        \left(
          \frac{4 K}{1 + K} \frac{\sigma_{\min}^2}{N_\A N_\B d} 
          \left(
            1 -  \frac{\inprod{\KA}{\KB}}{N_\A N_\B d} 
          \right)
          - O\left( 
            \frac{\sigma_{\max}^2}{N_\A N_\B d} 
            \left( d \tau_t^2 + \sqrt{\delta_{\A/\B}} \right) 
          \right)
        \right) \\
    &\le 
      - \frac{4 K}{1 + K} \frac{\sigma_{\min}^2}{N_\A N_\B d}
      \hat{\rho}_{N/S}^2
      + O\left( 
        \frac{\sigma_{\max}^2}{N_\A N_\B d} 
        \left( d \tau_t^2 + \sqrt{\delta_{\A/\B}} \right) 
        \hat{\rho}_{N/S}^2 
      \right). 
  \end{align*}
\end{proof}

\begin{proof}[Proof of Lemma~\ref{lemma: stage 1: growth rate of the condition number}]
  For notational simplicity, define $\rho_{p/q} := \norm{[\KA]_p}^2 / \norm{[\KA]_q}^2$. By 
  Corollary~\ref{cor: stage 1: d norm}, we have 
  \begin{align*}
    \dot{\rho}_{p/q}
    &= \frac{\frac{\rd}{\rd t} \norm{[\KA]_p}^2 }{ \norm{[\KA]_q}^2 } 
      - \rho_{p/q} \frac{ \frac{\rd}{\rd t} \norm{[\KA]_q}^2 }{ \norm{[\KA]_q}^2 }  \\
    &= 
      \frac{4 K}{1 + K} 
      \frac{\sigma_p^2}{N_\A N_\B d} 
      \left(
        \frac{ \inprod{[\KA]_p}{[\KB]_p} }{ \norm{[\KA]_p}^2 }
        - \frac{\inprod{\KA}{\KB}}{N_\A N_\B d} 
      \right) 
      \rho_{p/q}
      \pm O\left( 
        \frac{ \sigma_p^2   }{N_\A N_\B d} 
        \left( \sqrt{\delta_{\A/\B}} + \tau_t^2 d \right)
        \rho_{p/q} 
      \right)
      \\
      &\qquad
      - \rho_{p/q} \left(
        \frac{4 K}{1 + K} 
        \frac{\sigma_q^2}{N_\A N_\B d} 
        \left(
          \frac{ \inprod{[\KA]_q}{[\KB]_q} }{ \norm{[\KA]_q}^2 }
          - \frac{\inprod{\KA}{\KB}}{N_\A N_\B d} 
        \right) 
        \pm O\left( 
          \frac{ \sigma_q^2   }{N_\A N_\B d} 
          \left( \sqrt{\delta_{\A/\B}} + \tau_t^2 d \right)
        \right)
      \right) \\ 
    &= \frac{4 K}{1 + K} 
      \frac{\sigma_p^2}{N_\A N_\B d} 
      \left(
        \inprod{\ol{[\KA]_p}}{\ol{[\KB]_p}} 
        - \frac{\inprod{\KA}{\KB}}{N_\A N_\B d} 
      \right) 
      \rho_{p/q} \\ 
      &\qquad
      - \frac{4 K}{1 + K} 
        \frac{\sigma_q^2}{N_\A N_\B d} 
        \left(
          \inprod{\ol{[\KA]_q}}{\ol{[\KB]_q}} 
          - \frac{\inprod{\KA}{\KB}}{N_\A N_\B d} 
        \right)
        \rho_{p/q} \\
      &\qquad
      \pm O\left( 
        \frac{ \sigma_{\max}^2   }{N_\A N_\B d} 
        \left( \sqrt{\delta_{\A/\B}} + \tau_t^2 d \right)
        \rho_{p/q}
      \right). 
  \end{align*}
  Now we bound $\inprod{\ol{[\KA]_p}}{\ol{[\KB]_p}} - \inprod{\KA}{\KB} / (N_\A N_\B d)$. Clear that this 
  term is bounded by $2$. Meanwhile, by definition, we have $\inprod{\ol{[\KA]_p}}{\ol{[\KB]_p}} = 1 \pm \rho_-$. 
  For the second term, we have 
  \begin{align*}
    \frac{\inprod{\KA}{\KB}}{N_\A N_\B d} 
    &= \sum_{k=1}^r \inprod{\ol{[\KA]_k}}{\ol{[\KB]_k}}
      \frac{\norm{\KA}_k \norm{\KB}_k}{\norm{\KA}_F \norm{\KB}_F}
      \frac{\norm{\KA}_F \norm{\KB}_F}{N_\A N_\B d} \\ 
    &= \sum_{k=1}^r \left( 1 \pm \rho_- \right)
      \left( 1 \pm \min\braces{ \hat{\rho}_{N/S}, 1} \right)
      \frac{\kappa_k^2}{\norm{\bkappa}^2}
      \left( 1 \pm \sqrt{\delta_{A/\B}} \right) \\ 
    &= \left( 1 \pm \rho_- \pm \min\braces{ \hat{\rho}_{N/S}, 1} \right)
    \left( 1 \pm \sqrt{\delta_{A/\B}} \right). 
  \end{align*}
  Combine these together and we obtain 
  \[ 
    \left| \inprod{\ol{[\KA]_p}}{\ol{[\KB]_p}} - \frac{\inprod{\KA}{\KB}}{N_\A N_\B d}  \right|
    \le 2 \rho_-  + 2 \min\braces{ \hat{\rho}_{N/S}, 1} \pm \sqrt{\delta_{A/\B}}. 
  \] 
  The same is also true for $q$. Thus, 
  \[ 
    \dot{\rho}_{p/q}
    \le \frac{16 K}{1 + K} 
      \frac{\sigma_{\max}^2}{N_\A N_\B d} 
      \left( \rho_- + \min\braces{ \hat{\rho}_{N/S}, 1} \right) 
      \rho_{p/q} 
      \pm O\left( 
        \frac{ \sigma_{\max}^2   }{N_\A N_\B d} 
        \left( \sqrt{\delta_{\A/\B}} + \tau_t^2 d \right)
        \rho_{p/q}
      \right). 
  \] 
\end{proof}

\subsection{Controlling the discretization error}
\label{sec: stage 1: discretization}

In this subsection, we estimate the growth rate of the errors described in \eqref{eq: stage 1: finite-width 
approx infinite-width}. As in the previous subsections, the proofs are deferred to the end of this 
subsection. 

First, we consider the relative difference between $\norm{[\KA]_p}^2$ and $\norm{[\KB]_p}^2$. Instead of 
directly control the difference, we define 
\[
  \rho_{\A/\B, p} := \frac{\norm{[\KA]_p}^2}{\norm{[\KB]_p}^2} 
  \quad\text{and}\quad 
  \rho_{\B/\A, p} := \frac{\norm{[\KB]_p}^2}{\norm{[\KA]_p}^2}.
\]
Note that $\rho_{\A/\B, p} + \rho_{\B/\A, p} \ge 2$, with equality attained iff $\norm{[\KA]_p}^2 = 
\norm{[\KB]_p}^2$. Meanwhile, at initialization, this quantity can be made arbitrarily close to $2$. Hence,
it suffices to control the growth of this quantity. The reason we consider $\rho_{\A/\B, p} + \rho_{\B/\A, p}$
is to leverage the symmetry. Similarly, we also define 
\[
  \rho_{\A/\B, \xi, q} := \frac{\norm{[\KAxi]_q}^2}{\norm{[\KBxi]_q}^2} 
  \quad\text{and}\quad 
  \rho_{\B/\A, \xi, q} := \frac{\norm{[\KBxi]_q}^2}{\norm{[\KAxi]_q}^2},
\]
and analyze $\rho_{\A/\A, \xi} + \rho_{\B/\A, \xi}$. 

\begin{lemma}[Difference of diagonal terms]
  \label{lemma: stage 1: diff of diagonal}
  In Stage~1, we have 
  \begin{align*}
    \frac{\rd}{\rd t} \left( \rho_{\A/\B, p} + \rho_{\B/\A, p} \right)
    &\le O\left( 
        \frac{\sigma_p^2}{N_\A N_\B d} 
        \left( \delta_{\A/\B}^2 + \tau_t^2 d   \right)
      \right), \\
    \frac{\rd}{\rd t} \left( \rho_{\A/\B, \xi, q} + \rho_{\B/\A, \xi, q} \right)
    &\le O\left( 
        \frac{\sigma_{\max}^2}{N_\A N_\B d} 
        \left( \delta_{\A/\B}^2 + \tau_t^2 d   \right)
      \right). 
  \end{align*}
\end{lemma}

Now we consider the condition number of $\KAxi$. Unlike $\KA$, for the noises, the $\sigma$'s for different 
coordinates are the same. Hence, we have the following simple bound on the growth rate of $\delta_{\xi, \kappa_0}$. 

\begin{lemma}[Condition number of $\KAxi$]
  \label{lemma: stage 1: condition number of KAxi}
  Define $\rho_{\xi, p/q} := \norm{[\KAxi]_p}^2 / \norm{[\KAxi]_q}^2$. In Stage~1, we have 
  \[ 
    \dot{\rho}_{\xi, p/q}
    \le O\left( 
        \frac{\sigma_\xi^2}{N_\A N_\B d} 
        \left( d \tau_t^2 + \sqrt{\delta_{\A/\B}} \right) 
      \right). 
  \] 
\end{lemma}

Then, we consider the orthogonality conditions. 

\begin{lemma}[Orthogonality between signals]
  \label{lemma: stage 1: orthogonality between signals}
  For any $p \ne q$, define 
  \[
    \hat{\delta}_{\perp, p, q}
    := \inprod{\ol{[\KA]_p}}{\ol{[\KB]_q}}^2
      + \inprod{\ol{[\KB]_p}}{\ol{[\KA]_q}}^2
      + \inprod{\ol{[\KA]_p}}{\ol{[\KA]_q}}^2
      + \inprod{\ol{[\KB]_p}}{\ol{[\KB]_q}}^2. 
  \]
  In Stage~1, we have 
  \[
    \frac{\rd}{\rd t} \hat{\delta}_{\perp, p, q}
    \le O\left( \frac{\sigma_{\max}^2 }{N_\A N_\B d } \right) \rho_- \hat{\delta}_{\perp, p, q}
      + O\left(
        \frac{\sigma_{\max}^2 }{N_\A N_\B d }
        \left( \tau_t^2 d^2 \kappa_0 + d \sqrt{\delta_{\A/\B}} \right)
      \right).
  \]
\end{lemma}
Recall that $\rho_-$ converges to $0$ at a sufficiently rate so that, by Lemma~\ref{lemma: stage 1: gronwall},
$\hat{\delta}_{\perp, p, q}$ will not blow up. Meanwhile, since $\hat{\delta}_{\perp, p, q}$ can be made 
arbitrarily small at initialization, this implies that we can make sure it is still small at the end of 
Stage~1. Finally, we consider the orthogonality conditions between the signals and noises and between noises. 
The proof follows the same spirit.  

\begin{lemma}[Orthogonality between signals and noises]
  \label{lemma: stage 1: orthogonality between signals and noises}
  For any $p \in [r]$ and $q \in [d-r]$, define 
  \[
    \hat{\delta}_{\perp, \xi_\A, p, q} 
    = \inprod{\ol{[\KA]_p}}{\ol{[\KAxi]_q}}^2 
      + \inprod{\ol{[\KB]_p}}{\ol{[\KAxi]_q}}^2,
  \]
  and define $\hat{\delta}_{\perp, \xi_\B, p, q}$ similarly. In Stage~1, we have 
  \[ 
    \frac{\rd}{\rd t} \hat{\delta}_{\perp, \xi_\A, p, q} 
    \le O\left( \frac{\sigma_p^2 }{N_\A N_\B d } \right) \rho_-
        \hat{\delta}_{\perp, \xi_\A, p, q} 
      \pm O\left(
        \frac{\sigma_{\max}^2 }{N_\A N_\B d }
        \left( \tau_t^2 d^2 \kappa_0 + d \sqrt{\delta_{\A/\B}} \right)
      \right).
  \] 
  For any $q, s \in [d - r]$, define 
  \[
    \hat{\delta}_{\perp, \xi, p, q}
    = \inprod{\ol{[\KAxi]_q}}{\ol{[\KBxi]_s}}^2 
      + \inprod{\ol{[\KBxi]_q}}{\ol{[\KAxi]_s}}^2,
  \]
  In Stage~1, we have 
  \[
    \frac{\rd}{\rd t} \hat{\delta}_{\perp, \xi, p, q}
    \le O\left( 
        \frac{\sigma_\xi^2 }{N_\A N_\B d } 
        \tau_t^2 d^3 \kappa_0  \delta_{\xi, \perp} \sqrt{\hat{\delta}_{\perp, \xi, p, q}}
      \right). 
  \]
\end{lemma}

\subsubsection*{Omitted proofs of this subsection}
\begin{proof}[Proof of Lemma~\ref{lemma: stage 1: diff of diagonal} ]
  Note that we cannot directly use Corollary~\ref{cor: stage 1: d norm} as the error term contains 
  $\delta_{\A/\B}$, the quantity we wish to control. However, by the proof of it, we still have 
  \begin{align*}
    \frac{\rd}{\rd t} \norm{[\KA]_p}^2
    &= \frac{4 K}{1 + K} 
    \frac{\sigma_p^2}{N_\A N_\B d} 
      \inprod{[\KA]_p}{[\KB]_p} 
      - \frac{4 K}{1 + K} 
      \frac{\inprod{\KA}{\KB} \sigma_p^2}{N_\A N_\B d} 
      \frac{\norm{[\KA]_p}^2}{N_\A^2 d} \\
      &\qquad
      \pm O\left( 
        \frac{\sigma_p^2}{N_\A N_\B d} 
        \tau_t^2 d \kappa_p^2  
      \right). 
  \end{align*}
  Interchange the roles of $\A$ and $\B$ and we obtain 
  \begin{align*}
    \frac{\rd}{\rd t} \norm{[\KB]_p}^2
    &= \frac{4 K}{1 + K} 
      \frac{\sigma_p^2}{N_\A N_\B d} 
      \inprod{[\KA]_p}{[\KB]_p} 
      - \frac{4 K}{1 + K} 
      \frac{\inprod{\KA}{\KB} \sigma_p^2}{N_\A N_\B d} 
      \frac{\norm{[\KB]_p}^2}{N_\B^2 d} \\
      &\qquad
      \pm O\left( 
        \frac{\sigma_p^2}{N_\A N_\B d} 
        \tau_t^2 d \kappa_p^2  
      \right). 
  \end{align*}
  For notational simplicity, define $\rho_{\A/\B, p} = \norm{[\KA]_p}^2 / \norm{[\KB]_p}^2$. Then, we compute 
  \begin{align*}
    \dot{\rho}_{\A/\B, p}
    &= \frac{\frac{\rd}{\rd t} \norm{[\KA]_p}^2}{\norm{[\KB]_p}^2}
      - \rho_{\A/\B, p} \frac{\frac{\rd}{\rd t} \norm{[\KB]_p}^2}{\norm{[\KB]_p}^2} \\
    &= \frac{4 K}{1 + K} 
        \frac{\sigma_p^2}{N_\A N_\B d} 
        \frac{ \inprod{[\KA]_p}{[\KB]_p} }{\norm{[\KA]_p}^2} 
        \rho_{\A/\B, p}
      - \frac{4 K}{1 + K} 
        \frac{\inprod{\KA}{\KB} \sigma_p^2}{N_\A N_\B d} 
        \frac{\rho_{\A/\B, p}}{N_\A^2 d} \\
      &\qquad
      - \frac{4 K}{1 + K} 
        \frac{\sigma_p^2}{N_\A N_\B d} 
        \frac{ \inprod{[\KA]_p}{[\KB]_p} }{ \norm{[\KB]_p}^2 }
        \rho_{\A/\B, p}
      + \frac{4 K}{1 + K} 
        \frac{\inprod{\KA}{\KB} \sigma_p^2}{N_\A N_\B d} 
        \frac{\rho_{\A/\B, p}}{N_\B^2 d}
      \\
      &\qquad
      \pm O\left( 
        \frac{\sigma_p^2}{N_\A N_\B d} 
        \tau_t^2 d  
      \right) \\
    &= \frac{4 K}{1 + K} 
        \frac{\sigma_p^2}{N_\A N_\B d} 
        \inprod{[\KA]_p}{[\KB]_p} 
        \left( 
          \frac{1}{\norm{[\KA]_p}^2} 
          - \frac{1}{\norm{[\KB]_p}^2} 
        \right)
        \rho_{\A/\B, p} \\
      &\qquad
      - \frac{4 K}{1 + K} 
        \frac{\inprod{\KA}{\KB} \sigma_p^2}{N_\A N_\B d} 
        \left( \frac{1}{N_\A^2 d} - \frac{1}{N_\B^2 d}  \right)
        \rho_{\A/\B, p}
      \\
      &\qquad
      \pm O\left( 
        \frac{\sigma_p^2}{N_\A N_\B d} 
        \tau_t^2 d  
      \right). 
  \end{align*}
  Then, by symmetry, we have 
  \begin{align*}
    & \frac{\rd}{\rd t} \left( \rho_{\A/\B, p} + \rho_{\B/\A, p} \right) \\
    =\;& \frac{4 K}{1 + K} 
        \frac{\sigma_p^2}{N_\A N_\B d} 
        \inprod{[\KA]_p}{[\KB]_p} 
        \left( 
          \frac{1}{\norm{[\KA]_p}^2} 
          - \frac{1}{\norm{[\KB]_p}^2} 
        \right)
        \left( \rho_{\A/\B, p} - \rho_{\B/\A, p} \right) \\
      &\qquad
      - \frac{4 K}{1 + K} 
        \frac{\inprod{\KA}{\KB} \sigma_p^2}{N_\A N_\B d} 
        \left( \frac{1}{N_\A^2 d} - \frac{1}{N_\B^2 d}  \right)
        \left( \rho_{\A/\B, p} - \rho_{\B/\A, p} \right) \\
      &\qquad
      \pm O\left( 
        \frac{\sigma_p^2}{N_\A N_\B d} 
        \tau_t^2 d  
      \right) \\
    \le\;& O\left( 
        \frac{\sigma_p^2}{N_\A N_\B d} 
        \left( \delta_{\A/\B}^2 + \tau_t^2 d   \right)
      \right).  
  \end{align*}
  The above proof, \textit{mutatis mutandis}, yields the result for $\rho_{\A/\B, \xi, q} + \rho_{\B/\A, \xi, q}$.
\end{proof}

\begin{proof}[Proof of Lemma~\ref{lemma: stage 1: condition number of KAxi}]
  By Corollary~\ref{cor: stage 1: d norm}, we have 
  \begin{align*}
    \dot{\rho}_{\xi, p/q}
    &= \frac{\frac{\rd}{\rd t} \norm{[\KAxi]_p}^2}{\norm{[\KAxi]_q}^2}
      - \rho_{\xi, p/q} \frac{\frac{\rd}{\rd t} \norm{[\KAxi]_q}^2}{\norm{[\KAxi]_q}^2} \\
    &= - \frac{4 K}{1 + K}  
        \frac{\sigma_\xi^2}{N_\A N_\B d} 
        \frac{\inprod{\KA}{\KB}}{N_\A N_\B d} 
        \rho_{\xi, p/q}
      \pm O\left( 
          \frac{\sigma_\xi^2}{N_\A N_\B d} 
          \left( d \tau_t^2 + \sqrt{\delta_{\A/\B}} \right) 
        \right) \\
      &\qquad
      - \rho_{\xi, p/q} \left(
        - \frac{4 K}{1 + K}  
        \frac{\sigma_\xi^2}{N_\A N_\B d} 
        \frac{\inprod{\KA}{\KB}}{N_\A N_\B d} 
        \pm O\left( 
            \frac{\sigma_\xi^2}{N_\A N_\B d} 
            \left( d \tau_t^2 + \sqrt{\delta_{\A/\B}} \right) 
          \right)
      \right) \\
    &= \pm O\left( 
        \frac{\sigma_\xi^2}{N_\A N_\B d} 
        \left( d \tau_t^2 + \sqrt{\delta_{\A/\B}} \right) 
      \right). 
  \end{align*}
\end{proof}

\begin{proof}[Proof of Lemma~\ref{lemma: stage 1: orthogonality between signals}]
  By Corollary~\ref{cor: stage 1: d bar}, we have 
  \begin{align*}
    \inprod{ \frac{\rd}{\rd t} \ol{ [\KA]_p } }{ \ol{[\KB]_q} }
    &= \frac{2K}{1 + K} \frac{\sigma_p^2 }{N_\A N_\B d }
      \left(
        \inprod{\ol{[\KB]_p}}{\ol{[\KB]_q}}
        - \inprod{\ol{ [\KA]_p }}{\ol{[\KB]_q}}
          \inprod{\ol{ [\KA]_p }}{\ol{[\KB]_p}}
      \right) \\
      &\qquad
      \pm O\left(
        \frac{\sigma_p^2 }{N_\A N_\B d }
        \left( \tau_t^2 d^2 \kappa_0 + d \sqrt{\delta_{\A/\B}} \right)
      \right) \\
    &= \frac{2K}{1 + K} \frac{\sigma_p^2 }{N_\A N_\B d }
      \left(
        \inprod{\ol{[\KB]_p}}{\ol{[\KB]_q}}
        - \inprod{\ol{ [\KA]_p }}{\ol{[\KB]_q}}
      \right) \\
      &\qquad
      + \frac{2K}{1 + K} \frac{\sigma_p^2 }{N_\A N_\B d }
        \inprod{\ol{ [\KA]_p }}{\ol{[\KB]_q}}
        \left( 1 - \inprod{\ol{ [\KA]_p }}{\ol{[\KB]_p}} \right) \\
      &\qquad
      \pm O\left(
        \frac{\sigma_p^2 }{N_\A N_\B d }
        \left( \tau_t^2 d^2 \kappa_0 + d \sqrt{\delta_{\A/\B}} \right)
      \right). 
  \end{align*}
  Interchange the roles of $p, q$ and $\A, \B$ and we obtain 
  \begin{align*}
    \inprod{ \ol{[\KA]_p} }{ \frac{\rd}{\rd t} \ol{ [\KB]_q } }
    &= \frac{2K}{1 + K} \frac{\sigma_q^2 }{N_\A N_\B d }
      \left(
        \inprod{\ol{[\KA]_p}}{\ol{[\KA]_q}}
        - \inprod{\ol{ [\KA]_p }}{\ol{[\KB]_q}}
      \right) \\
      &\qquad
      + \frac{2K}{1 + K} \frac{\sigma_q^2 }{N_\A N_\B d }
        \inprod{\ol{ [\KA]_p }}{\ol{[\KB]_q}}
        \left( 1 - \inprod{\ol{ [\KA]_q }}{\ol{[\KB]_q}} \right) \\
      &\qquad
      \pm O\left(
        \frac{\sigma_q^2 }{N_\A N_\B d }
        \left( \tau_t^2 d^2 \kappa_0 + d \sqrt{\delta_{\A/\B}} \right)
      \right).
  \end{align*}
  Therefore, 
  \begin{align*}
    \frac{\rd}{\rd t} \inprod{ \ol{[\KA]_p} }{ \ol{ [\KB]_q } }
    &= \frac{2K}{1 + K} \frac{\sigma_p^2 }{N_\A N_\B d }
      \left(
        \inprod{\ol{[\KB]_p}}{\ol{[\KB]_q}}
        - \inprod{\ol{ [\KA]_p }}{\ol{[\KB]_q}}
      \right) \\
      &\qquad
      + \frac{2K}{1 + K} \frac{\sigma_q^2 }{N_\A N_\B d }
      \left(
        \inprod{\ol{[\KA]_p}}{\ol{[\KA]_q}}
        - \inprod{\ol{ [\KA]_p }}{\ol{[\KB]_q}}
      \right) \\
      &\qquad
      \pm O\left( \frac{\sigma_{\max}^2 }{N_\A N_\B d } \right)
      \rho_- \inprod{\ol{ [\KA]_p }}{\ol{[\KB]_q}} \\
      &\qquad
      \pm O\left(
        \frac{\sigma_{\max}^2 }{N_\A N_\B d }
        \left( \tau_t^2 d^2 \kappa_0 + d \sqrt{\delta_{\A/\B}} \right)
      \right).
  \end{align*}
  Interchange the roles of $p$ and $q$ and we obtain 
  \begin{align*}
    \frac{\rd}{\rd t} \inprod{ \ol{ [\KB]_p } }{ \ol{[\KA]_q} }
    &= \frac{2K}{1 + K} \frac{\sigma_q^2 }{N_\A N_\B d }
      \left(
        \inprod{\ol{[\KB]_p}}{\ol{[\KB]_q}}
        - \inprod{\ol{[\KB]_p}}{\ol{ [\KA]_q }}
      \right) \\
      &\qquad
      + \frac{2K}{1 + K} \frac{\sigma_p^2 }{N_\A N_\B d }
      \left(
        \inprod{\ol{[\KA]_p}}{\ol{[\KA]_q}}
        - \inprod{\ol{[\KB]_p}}{\ol{ [\KA]_q }}
      \right) \\
      &\qquad
      \pm O\left( \frac{\sigma_{\max}^2 }{N_\A N_\B d } \right)
      \rho_- \inprod{\ol{[\KB]_p}}{\ol{ [\KA]_q }} \\
      &\qquad
      \pm O\left(
        \frac{\sigma_{\max}^2 }{N_\A N_\B d }
        \left( \tau_t^2 d^2 \kappa_0 + d \sqrt{\delta_{\A/\B}} \right)
      \right).
  \end{align*}
  Similarly, we compute 
  \begin{align*}
    \inprod{ \frac{\rd}{\rd t} \ol{[\KA]_p} }{ \ol{[\KA]_q} }
    &= \frac{2K}{1 + K} \frac{\sigma_p^2 }{N_\A N_\B d }
      \left( 
        \inprod{\ol{[\KB]_p}}{\ol{[\KA]_q}}
        - \inprod{ \ol{ [\KA]_p } }{ \ol{[\KA]_q} }
          \inprod{ \ol{ [\KA]_p } }{ \ol{[\KB]_p} }
      \right) \\
      &\qquad
      \pm O\left(
        \frac{\sigma_p^2 }{N_\A N_\B d }
        \left( \tau_t^2 d^2 \kappa_0 + d \sqrt{\delta_{\A/\B}} \right)
      \right) \\
    &= \frac{2K}{1 + K} \frac{\sigma_p^2 }{N_\A N_\B d }
      \left( 
        \inprod{\ol{[\KB]_p}}{\ol{[\KA]_q}}
        - \inprod{ \ol{ [\KA]_p } }{ \ol{[\KA]_q} }
      \right) \\
      &\qquad
      + \frac{2K}{1 + K} \frac{\sigma_p^2 }{N_\A N_\B d }
        \inprod{ \ol{ [\KA]_p } }{ \ol{[\KA]_q} }
        \left( 1 - \inprod{ \ol{ [\KA]_p } }{ \ol{[\KB]_p} } \right)  \\
      &\qquad
      \pm O\left(
        \frac{\sigma_p^2 }{N_\A N_\B d }
        \left( \tau_t^2 d^2 \kappa_0 + d \sqrt{\delta_{\A/\B}} \right)
      \right) \\
    &= \frac{2K}{1 + K} \frac{\sigma_p^2 }{N_\A N_\B d }
      \left( 
        \inprod{\ol{[\KB]_p}}{\ol{[\KA]_q}}
        - \inprod{ \ol{ [\KA]_p } }{ \ol{[\KA]_q} }
      \right) \\
      &\qquad
      \pm  O\left( \frac{\sigma_{\max}^2 }{N_\A N_\B d } \right)
        \rho_-
        \inprod{ \ol{ [\KA]_p } }{ \ol{[\KA]_q} }
      \pm O\left(
        \frac{\sigma_{\max}^2 }{N_\A N_\B d }
        \left( \tau_t^2 d^2 \kappa_0 + d \sqrt{\delta_{\A/\B}} \right)
      \right). 
  \end{align*}
  Then, by interchanging the roles of $p$ and $q$, we obtain 
  \begin{align*}
    \inprod{ \ol{[\KA]_p} }{ \frac{\rd}{\rd t} \ol{[\KA]_q} }
    &= \frac{2K}{1 + K} \frac{\sigma_q^2 }{N_\A N_\B d }
      \left( 
        \inprod{\ol{[\KA]_p}}{\ol{[\KB]_q}}
        - \inprod{ \ol{[\KA]_p} }{ \ol{ [\KA]_q } }
      \right) \\
      &\qquad
      \pm  O\left( \frac{\sigma_{\max}^2 }{N_\A N_\B d } \right)
        \rho_-
        \inprod{ \ol{ [\KA]_p } }{ \ol{[\KA]_q} }
      \pm O\left(
        \frac{\sigma_{\max}^2 }{N_\A N_\B d }
        \left( \tau_t^2 d^2 \kappa_0 + d \sqrt{\delta_{\A/\B}} \right)
      \right). 
  \end{align*}
  Add them together and we get 
  \begin{align*}
    \frac{\rd}{\rd t} \inprod{ \ol{[\KA]_p} }{ \ol{[\KA]_q} }
    &= \frac{2K}{1 + K} \frac{\sigma_p^2 }{N_\A N_\B d }
      \left( 
        \inprod{\ol{[\KB]_p}}{\ol{[\KA]_q}}
        - \inprod{ \ol{ [\KA]_p } }{ \ol{[\KA]_q} }
      \right) \\
      &\qquad
      + \frac{2K}{1 + K} \frac{\sigma_q^2 }{N_\A N_\B d }
        \left( 
          \inprod{\ol{[\KA]_p}}{\ol{[\KB]_q}}
          - \inprod{ \ol{[\KA]_p} }{ \ol{ [\KA]_q } }
        \right) \\
      &\qquad
      \pm  O\left( \frac{\sigma_{\max}^2 }{N_\A N_\B d } \right)
        \rho_-
        \inprod{ \ol{ [\KA]_p } }{ \ol{[\KA]_q} }
      \pm O\left(
        \frac{\sigma_{\max}^2 }{N_\A N_\B d }
        \left( \tau_t^2 d^2 \kappa_0 + d \sqrt{\delta_{\A/\B}} \right)
      \right). 
  \end{align*}
  Interchange the roles of $p$ and $q$ and we obtain 
  \begin{align*}
    \frac{\rd}{\rd t} \inprod{ \ol{[\KB]_p} }{ \ol{[\KB]_q} }
    &= \frac{2K}{1 + K} \frac{\sigma_p^2 }{N_\A N_\B d }
      \left( 
        \inprod{\ol{[\KA]_p}}{\ol{[\KB]_q}}
        - \inprod{ \ol{ [\KB]_p } }{ \ol{[\KB]_q} }
      \right) \\
      &\qquad
      + \frac{2K}{1 + K} \frac{\sigma_q^2 }{N_\A N_\B d }
        \left( 
          \inprod{\ol{[\KB]_p}}{\ol{[\KA]_q}}
          - \inprod{ \ol{[\KB]_p} }{ \ol{ [\KB]_q } }
        \right) \\
      &\qquad
      \pm  O\left( \frac{\sigma_{\max}^2 }{N_\A N_\B d } \right) \rho_-
        \inprod{ \ol{ [\KB]_p } }{ \ol{[\KB]_q} }
      \pm O\left(
        \frac{\sigma_{\max}^2 }{N_\A N_\B d }
        \left( \tau_t^2 d^2 \kappa_0 + d \sqrt{\delta_{\A/\B}} \right)
      \right). 
  \end{align*}
  For notational simplicity, define $Z_1 = \inprod{\ol{[\KA]_p}}{\ol{[\KB]_q}}$, 
  $Z_2 = \inprod{\ol{[\KB]_p}}{\ol{[\KA]_q}}$, $Z_3 = \inprod{\ol{[\KA]_p}}{\ol{[\KA]_q}}$, and 
  $Z_4 = \inprod{\ol{[\KB]_p}}{\ol{[\KB]_q}}$. Also define $G_p = \frac{2 K}{1 + K} \frac{\sigma_p^2}{N_\A N_\B d}$.
  Then, we can summarize the above equations as 
  \begin{multline*}
    \frac{\rd}{\rd t} \begin{bmatrix} Z_1 \\ Z_2 \\ Z_3 \\ Z_4 \end{bmatrix}
    = 
      \begin{bmatrix}
        - G_p - G_q & 0 & G_q & G_p \\
        0 & - G_p - G_q & G_p & G_q \\
        G_q & G_p & - G_p - G_q & 0 \\
        G_p & G_q & 0 & - G_p - G_q \\
      \end{bmatrix} 
      \begin{bmatrix} Z_1 \\ Z_2 \\ Z_3 \\ Z_4 \end{bmatrix} \\
      \pm O\left( \frac{\sigma_{\max}^2 }{N_\A N_\B d } \right) \rho_-
        \begin{bmatrix} Z_1 \\ Z_2 \\ Z_3 \\ Z_4 \end{bmatrix}  
      \pm O\left(
        \frac{\sigma_{\max}^2 }{N_\A N_\B d }
        \left( \tau_t^2 d^2 \kappa_0 + d \sqrt{\delta_{\A/\B}} \right)
      \right).
  \end{multline*}
  Note that the eigenvalues of the first matrix is $-2G_p - 2G_q$, $-2G_p$, $-2G_q$ and $0$. Namely, 
  it is negative semi-definite. Thus, 
  \[
    \frac{\rd}{\rd t} \norm{\Z}^2
    \le O\left( \frac{\sigma_{\max}^2 }{N_\A N_\B d } \right) \rho_- \norm{\Z}^2
      + O\left(
        \frac{\sigma_{\max}^2 }{N_\A N_\B d }
        \left( \tau_t^2 d^2 \kappa_0 + d \sqrt{\delta_{\A/\B}} \right)
      \right).
  \]
\end{proof}

\begin{proof}[Proof of Lemma~\ref{lemma: stage 1: orthogonality between signals and noises}]
  By Corollary~\ref{cor: stage 1: d bar},
  \begin{align*}
    \inprod{\frac{\rd}{\rd t} \ol{[\KA]_p}}{\ol{[\KAxi]_q}}
    &= \frac{2K}{1 + K} \frac{\sigma_p^2 }{N_\A N_\B d }
      \left( 
        \inprod{\ol{[\KB]_p}}{\ol{[\KAxi]_q}}
        - \inprod{ \ol{ [\KA]_p } }{ \ol{[\KAxi]_q} }
          \inprod{ \ol{ [\KA]_p } }{ \ol{[\KB]_p} }
      \right) \\
      &\qquad
      \pm O\left(
        \frac{\sigma_p^2 }{N_\A N_\B d }
        \left( \tau_t^2 d^2 \kappa_0 + d \sqrt{\delta_{\A/\B}} \right)
      \right),
  \end{align*}
  and 
  \[ 
    \inprod{\ol{[\KA]_p}}{\frac{\rd}{\rd t} \ol{[\KAxi]_q}}
    = \pm O\left( 
        \frac{\sigma_\xi^2 }{N_\A N_\B d } 
        \tau_t^2 d^3 \kappa_0  \delta_{\xi, \perp}
      \right). 
  \]
  Therefore, 
  \begin{align*}
    \frac{\rd}{\rd t} \inprod{\ol{[\KA]_p}}{\ol{[\KAxi]_q}}
    &= \frac{2K}{1 + K} \frac{\sigma_p^2 }{N_\A N_\B d }
      \left( 
        \inprod{\ol{[\KB]_p}}{\ol{[\KAxi]_q}}
        - \inprod{ \ol{ [\KA]_p } }{ \ol{[\KAxi]_q} }
      \right) \\
      &\qquad
      \pm O\left( \frac{\sigma_p^2 }{N_\A N_\B d } \right) \rho_-
         \inprod{ \ol{ [\KA]_p } }{ \ol{[\KAxi]_q} }  
      \pm O\left(
        \frac{\sigma_{\max}^2 }{N_\A N_\B d }
        \left( \tau_t^2 d^2 \kappa_0 + d \sqrt{\delta_{\A/\B}} \right)
      \right). 
  \end{align*}
  Similarly, we also have 
  \begin{align*}
    \frac{\rd}{\rd t} \inprod{\ol{[\KB]_p}}{\ol{[\KAxi]_q}}
    &= \frac{2K}{1 + K} \frac{\sigma_p^2 }{N_\A N_\B d }
      \left( 
        \inprod{\ol{[\KA]_p}}{\ol{[\KAxi]_q}}
        - \inprod{ \ol{ [\KB]_p } }{ \ol{[\KAxi]_q} }
      \right) \\
      &\qquad
      \pm O\left( \frac{\sigma_p^2 }{N_\A N_\B d } \right) \rho_-
         \inprod{ \ol{ [\KB]_p } }{ \ol{[\KAxi]_q} }  
      \pm O\left(
        \frac{\sigma_{\max}^2 }{N_\A N_\B d }
        \left( \tau_t^2 d^2 \kappa_0 + d \sqrt{\delta_{\A/\B}} \right)
      \right). 
  \end{align*}
  Thus, 
  \begin{align*}
    & \frac{\rd}{\rd t} \left( 
        \inprod{\ol{[\KA]_p}}{\ol{[\KAxi]_q}}^2 
        + \inprod{\ol{[\KB]_p}}{\ol{[\KAxi]_q}}^2
      \right) \\
    =\;& 
      - \frac{4 K}{1 + K} \frac{\sigma_p^2 }{N_\A N_\B d }
      \left( 
        \inprod{\ol{[\KA]_p}}{\ol{[\KAxi]_q}}^2 
        - \inprod{\ol{[\KB]_p}}{\ol{[\KAxi]_q}}^2
      \right)^2 \\
      &\qquad
      \pm O\left( \frac{\sigma_p^2 }{N_\A N_\B d } \right) \rho_-
        \left( 
          \inprod{\ol{[\KA]_p}}{\ol{[\KAxi]_q}}^2 
          + \inprod{\ol{[\KB]_p}}{\ol{[\KAxi]_q}}^2
        \right) \\ 
      &\qquad
      \pm O\left(
        \frac{\sigma_{\max}^2 }{N_\A N_\B d }
        \left( \tau_t^2 d^2 \kappa_0 + d \sqrt{\delta_{\A/\B}} \right)
      \right) \\
    \le\;& 
        O\left( \frac{\sigma_p^2 }{N_\A N_\B d } \right) \rho_-
        \left( 
          \inprod{\ol{[\KA]_p}}{\ol{[\KAxi]_q}}^2 
          + \inprod{\ol{[\KB]_p}}{\ol{[\KAxi]_q}}^2
        \right) \\ 
      &\qquad
      \pm O\left(
        \frac{\sigma_{\max}^2 }{N_\A N_\B d }
        \left( \tau_t^2 d^2 \kappa_0 + d \sqrt{\delta_{\A/\B}} \right)
      \right). 
  \end{align*}
  For the orthogonality between noises, by Corollary~\ref{cor: stage 1: d bar}, we have 
  \[ 
    \inprod{\ol{[\KAxi]_q}}{\frac{\rd}{\rd t} \ol{[\KAxi]_s}}
    = \pm O\left( 
      \frac{\sigma_\xi^2 }{N_\A N_\B d } 
      \tau_t^2 d^3 \kappa_0  \delta_{\xi, \perp}
    \right)
  \]
  Clear that this bound holds for all other combinations. Thus, 
  \[
    \frac{\rd}{\rd t} \hat{\delta}_{\perp, \xi, p, q}
    \le O\left( 
        \frac{\sigma_\xi^2 }{N_\A N_\B d } 
        \tau_t^2 d^3 \kappa_0  \delta_{\xi, \perp} \sqrt{\hat{\delta}_{\perp, \xi, p, q}}
      \right). 
  \]
\end{proof}

\subsection{Proof of the main lemma of Stage 1}
\label{sec: stage 1: proof of the main lemma}

\begin{proof}[Proof of Lemma~\ref{lemma: stage 1: main}]
  \def\currentprefix{proof: stage 1: main}
  First, we recap the estimations we have derived in previous subsections and introduce some notations. 
  By Lemma~\ref{lemma: stage 1: convergence rate of rho-} and Lemma~\ref{lemma: stage 1: convergence rate of rhoNS},
  we have 
  \begin{align*}
    \frac{\rd}{\rd t} \rho_-
    &\le - \Omega(1) \frac{\sigma_{\min}^2 }{N_\A N_\B d } \rho_- 
        + O\left(
          \frac{\sigma_{\max}^2 }{N_\A N_\B d }
          \left( \tau_t^2 d^2 \kappa_0 + \sqrt{\delta_{\A/\B}} \right)
        \right), \\
    \frac{\rd}{\rd t} \hat{\rho}_{N/S}
    &\le 
      - \Omega(1) \frac{\sigma_{\min}^2}{N_\A N_\B d} \hat{\rho}_{N/S}
      + O\left( 
        \frac{\sigma_{\max}^2}{N_\A N_\B d} 
        \left( d \tau_t^2 + \sqrt{\delta_{\A/\B}} \right) 
        d
      \right). 
  \end{align*}
  Define $\rho := \max\braces{ \rho_-, \hat{\rho}_{N/S} }$ to be the indicator of the progress we have 
  made. We have 
  \begin{equation}
    \locallabel{eq: d rho}
    \frac{\rd}{\rd t} \rho 
    \le - \Omega(1) \frac{\sigma_{\min}^2 }{N_\A N_\B d } \rho 
      + O\left(
        \frac{\sigma_{\max}^2 }{N_\A N_\B d }
        \left( \tau_t^2 d^2 \kappa_0 + d \sqrt{\delta_{\A/\B}} \right). 
      \right). 
  \end{equation}
  Let $\hat{\kappa}_0 := \max_{p, q} \norm{[\KA]_p}^2 / \norm{[\KA]_q}^2$ be the condition number at time $t$. 
  By Lemma~\ref{lemma: stage 1: growth rate of the condition number}, we have  
  \begin{equation}
    \locallabel{eq: d hat kappa0}
    \frac{\rd}{\rd t} \hat{\kappa}_0
    \le O(1) \frac{\sigma_{\max}^2}{N_\A N_\B d} \rho \hat{\kappa}_0. 
  \end{equation}
  Now we consider the discretization errors, define 
  \[
    \hat{\delta}_{\A/\B}
    = \max_{p \in [r], q \in [d-r]}\braces{  
        \rho_{\A/\B, p} + \rho_{\B/\A, p} - 2, 
        \rho_{\A/\B, \xi, q} + \rho_{\B/\A, \xi, q} - 2
      }. 
  \]
  Note that at time $t$, the first condition of \eqref{eq: stage 1: finite-width approx infinite-width} holds 
  with $\delta_{\A/\B}$ replaced by $O(\hat\delta_{\A/\B})$. Meanwhile, by Lemma~\ref{lemma: stage 1: 
  diff of diagonal}, we have 
  \begin{equation}
    \locallabel{eq: d deltaAB}
    \frac{\rd}{\rd t} \hat{\delta}_{\A/\B}
    \le O\left( 
        \frac{\sigma_{\max}^2}{N_\A N_\B d} 
        \left( \delta_{\A/\B}^2 + \tau_t^2 d   \right)
      \right). 
  \end{equation}
  Let $\hat{\delta}_{\xi, \kappa_0}(t)$ be the smallest number such that the second condition of 
  \eqref{eq: stage 1: finite-width approx infinite-width} holds at time $t$. By Lemma~\ref{lemma: stage 1:
  condition number of KAxi}, we have 
  \begin{equation}
    \locallabel{eq: d delta xi kappa0}
    \frac{\rd}{\rd t} \hat{\delta}_{\xi, \kappa_0}
    \le O\left( 
        \frac{\sigma_\xi^2}{N_\A N_\B d} 
        \left( d \tau_t^2 + \sqrt{\delta_{\A/\B}} \right) 
      \right). 
  \end{equation}
  Then, define 
  \[
    \hat{\delta}_\perp 
    := \max\braces{ 
        \hat{\delta}_{\perp, p, q}, 
        \hat{\delta}_{\perp, \xi_\A, p, k}, 
        \hat{\delta}_{\perp, \xi_\B, p, k}, 
        \hat{\delta}_{\perp, \xi, k, l}, 
        \,:\, 
        p \ne q \in [d], k, l \in [d-r]
      }
  \]
  Clear that the last two conditions hold at time $t$ when $\delta_{\AB, \perp}$ and $\delta_{\xi, \perp}$ 
  are replaced by $\sqrt{\hat{\delta}_\perp(t)}$. By Lemma~\ref{lemma: stage 1: orthogonality between signals} 
  and Lemma~\ref{lemma: stage 1: orthogonality between signals and noises}, we have 
  \begin{equation}
    \locallabel{eq: d delta perp}
    \frac{\rd}{\rd t} \hat{\delta}_\perp 
    \le O\left( \frac{\sigma_{\max}^2 }{N_\A N_\B d } \right) \rho \hat{\delta}_\perp 
      + O\left(
        \frac{\sigma_{\max}^2 }{N_\A N_\B d }
        \left( \tau_t^2 d^2 \kappa_0 + d \sqrt{\delta_{\A/\B}} \right)
      \right).
  \end{equation}

  Now, we are ready to show that the errors do not blow up in Stage~1. Note that for all these $\delta$'s, 
  we can make them arbitrarily inverse-polynomially small by choosing a sufficiently large $m$.

  First, we consider $\hat\delta_{\A/\B}$. Note that the dependence of the RHS of (\localref{eq: d deltaAB}) on 
  $\hat\delta_{\A/\B}$ is quadratic. Hence, by making the initial value of $\hat\delta_{\A/\B}$, the RHS can 
  be made to be dominated the $\tau_t^2$-related terms. Hence, $\hat{\delta}_{\A/\B} \le 
  O\left( \frac{\sigma_{\max}^2}{N_\A N_\B d} \tau_t^2 d T_1 \right)$. As we will see later, $T_1 = \poly(d)$. 
  Therefore, by choosing a sufficiently small $\tau_t^2$, we can make $\hat\delta_{\A/\B}$ remain small 
  throughout Stage~1. 

  Then, we consider $\hat{\delta}_{\xi, \kappa_0}$. As we have argued earlier, the RHS of 
  (\localref{eq: d delta xi kappa0}) can be made arbitrarily small, so that $\hat\delta_{\xi, \kappa_0}$ remains 
  small in Stage~1. 

  Now, we consider the condition number $\hat\kappa_0$ and $\hat\delta_\perp$. For $\hat\delta_\perp$, by our 
  previous argument, the second term of the RHS of (\localref{eq: d delta perp}) can be merged into the first 
  term, by choosing a sufficiently large $m$ and a sufficiently small $\tau_t^2$. The same is also true 
  for (\localref{eq: d rho}). Hence, for these quantities, 
  we have 
  \[ 
    \frac{\rd}{\rd t} \rho 
    \le - \Omega(1) \frac{\sigma_{\min}^2 }{N_\A N_\B d } \rho,  \quad 
    \frac{\rd}{\rd t} \hat\kappa_0 
    \le O(1) \frac{\sigma_{\max}^2}{N_\A N_\B d} \rho \hat{\kappa}_0, \quad 
    \frac{\rd}{\rd t} \hat{\delta}_\perp 
    \le O(1) \frac{\sigma_{\max}^2 }{N_\A N_\B d } \rho \hat{\delta}_\perp . 
  \] 
  Hence, by Lemma~\ref{lemma: stage 1: gronwall}, we have 
  \[
    \hat\kappa_0 
    \le \hat\kappa_0(0) \exp\left( O\left( \frac{\sigma_{\max}^2}{\sigma_{\min}^2} \right) \rho(0) \right), 
    \quad 
    \hat{\delta}_\perp 
    \le \hat{\delta}_\perp (0) \exp\left( O\left( \frac{\sigma_{\max}^2}{\sigma_{\min}^2} \right) \rho(0) \right).
  \]
  Note that $\rho_-(0) = O(1)$ and $\hat\rho_{N/S} \le \frac{ (d - r) \sigma_\xi^2 }{ r \sigma_{\min}^2 } $. 
  Therefore, 
  \[
    \exp\left( O\left( \frac{\sigma_{\max}^2}{\sigma_{\min}^2} \right) \rho(0) \right)
    \le \exp\left( 
        O(1) 
        \frac{\sigma_{\max}^2}{\sigma_{\min}^2}
        \max\braces{ 1, \frac{ (d - r) \sigma_\xi^2 }{ r \sigma_{\min}^2 } }
      \right)
    \le \exp\left( \frac{1}{2} \log d  \right)
    = \sqrt{d}. 
  \]
  In other words, both $\hat\kappa_0$ and $\hat{\delta}_\perp$ can at most grow $\sqrt{d}$ times. 

  Finally, we derive an upper bound on $T_1$ to complete the proof. Similar to the proof for the condition
  number, one can show that $\norm{[\KA]_p}^2$ can at most grow $\sqrt{d}$ times in Stage~1. As a result, 
  $1 / (N_\A N_\B d)$ is lower bounded by some $1 / \poly(d)$. Thus, by (\localref{eq: d rho}), $T_1 
  \le \poly(d)$. 
\end{proof}

\subsection{Negative results}
\label{sec: stage 1: negative results}

\begin{lemma}
  There exists a $\bsigma \in \R^r$ satisfying the assumptions of Theorem~\ref{thm: main} such that, at the 
  end of Stage~1, the condition number of $\KA$ is $d^{\Omega(1)}$. 
\end{lemma}
\begin{proof}
  We choose $d = r$ and $\sigma_1^2 = c \log d $, $\sigma_2^2 = \cdots = \sigma_d^2 = 1$. Clear that this 
  satisfies the condition of Theorem~\ref{thm: main}. Note that it suffices to consider the infinite-width case, 
  since, as we have proved earlier, the finite-width trajectory tracks the infinite-width one. 
  By Lemma~\ref{lemma: stage 1: convergence rate of rho-}, we have 
  \begin{align*}
    \frac{\rd}{\rd t}  \inprod{\ol{[\KA]_p}}{ \ol{[\KB]_p} }
    \approx \frac{4 K}{1 + K} \frac{\sigma_p^2 }{N_\A N_\B d }
      \left( 1 + \inprod{\ol{[\KA]_p}}{ \ol{[\KB]_p} } \right)
      \left( 1 - \inprod{\ol{[\KA]_p}}{ \ol{[\KB]_p} } \right). 
  \end{align*}
  By the proof of Lemma~\ref{lemma: stage 1: growth rate of the condition number}, we have 
  \begin{align*}
    \dot{\rho}_{p/q}
    &\approx \frac{4 K}{1 + K} 
      \frac{\sigma_p^2}{N_\A N_\B d} 
      \left(
        \inprod{\ol{[\KA]_p}}{\ol{[\KB]_p}} 
        - \frac{\inprod{\KA}{\KB}}{N_\A N_\B d} 
      \right) 
      \rho_{p/q} \\ 
      &\qquad
      - \frac{4 K}{1 + K} 
        \frac{\sigma_q^2}{N_\A N_\B d} 
        \left(
          \inprod{\ol{[\KA]_q}}{\ol{[\KB]_q}} 
          - \frac{\inprod{\KA}{\KB}}{N_\A N_\B d} 
        \right)
        \rho_{p/q}. 
  \end{align*}
  Note that, in the infinite-width case, we have 
  \[
    \inprod{\ol{[\KA]_1}}{\ol{[\KB]_1}}
    \ge \inprod{\ol{[\KA]_2}}{\ol{[\KB]_2}}
    = \cdots
    = \inprod{\ol{[\KA]_d}}{\ol{[\KB]_d}}. 
  \]
  Therefore, $\inprod{\ol{[\KA]_p}}{\ol{[\KB]_p}} - \frac{\inprod{\KA}{\KB}}{N_\A N_\B d} \ge 0$a
  and $\inprod{\ol{[\KA]_q}}{\ol{[\KB]_q}} - \frac{\inprod{\KA}{\KB}}{N_\A N_\B d} \le 0$ for any $q \ge 2$. 
  Hence, 
  \begin{align*}
    \dot{\rho}_{1/2}
    &\ge \frac{4 K}{1 + K} 
      \frac{\sigma_1^2}{N_\A N_\B d} 
      \left(
        \inprod{\ol{[\KA]_1}}{\ol{[\KB]_1}} 
        - \frac{\inprod{\KA}{\KB}}{N_\A N_\B d} 
      \right) 
      \rho_{1/2} \\
    &\ge \frac{4 K}{1 + K} 
      \frac{\sigma_1^2}{N_\A N_\B d} 
      \left(
        \left( 1 - \frac{\kappa_1^2}{\norm{\bkappa}^2} \right) \inprod{\ol{[\KA]_1}}{\ol{[\KB]_1}} 
        - \frac{(d - 1) \kappa_1^2}{\norm{\bkappa}^2} \inprod{\ol{[\KA]_2}}{\ol{[\KB]_2}} 
      \right) 
      \rho_{1/2}. 
  \end{align*}
  For notational simplicity, define $X_1 = 1 - \inprod{\ol{[\KA]_1}}{\ol{[\KB]_1}}$, 
  $X_2 = 1 - \inprod{\ol{[\KA]_2}}{\ol{[\KB]_2}}$, $Y = \rho_{1/2}$, $A = \frac{4 K}{1 + K} 
  \frac{1}{N_\A N_\B d}$. Then we have 
  \begin{align*}
    \dot{X}_1 
    \le - \sigma_1^2 A X_1, \quad 
    \dot{X}_2
    \ge - 2 A X_2. 
  \end{align*}
  First, by Gronwall's lemma, we have $X_1(T) \le \exp\left( - \sigma_1^2 \int_0^T A \right)$ and 
  \[
    X_2(T) 
    \ge \exp\left( -2 \int_0^T A \right)
    \ge X_1^{2/\sigma_1^2}(T). 
  \]
  As a result, when $X_1$ reaches $1/d$, we have $X_2 = \Omega(1)$. Let $T_1$ be the time $X_1$ reaches $1/d$
  and $T_2$ the time $X_2(T_2) = X_2(T_1) / 2$. On $[T_1, T_2]$, we have 
  \[
    \dot{\rho}_{1/2} \ge \Omega(1) \sigma_1^2 A \rho_{1/2}. 
  \]
  By Gronwall's lemma, in order for $X_2$ to half, we need $\exp(- 2 \int_{T_1}^{T_2} A) = 1/2$. Hence, 
  \[
    \rho_{1/2}(T_2)
    \ge \rho_{1/2}(T_1) \exp\left( \Omega(1) \sigma_1^2 \int_{T_1}^{T_2} A  \right)
    \ge 2^{\Omega(\sigma_1^2)}
    = d^{\Omega(1)}. 
  \]
\end{proof}

\section{Stage 2}
\label{sec: stage 2}

In this section, we show that, throughout Stage~2, the discretization error and the noise-signal 
ratio still remain small, and, at the end of Stage~2, the condition number is close to $1$. Formally, we prove the 
following. 

\begin{lemma}[Stage 2]
  \label{lemma: stage 2: main}
  Suppose that at the beginning of Stage~2, we have $\kappa_0 \le \sqrt{d}$ and all errors mentioned in 
  \eqref{eq: stage 2: finite-width approx infinite-width} are sufficiently small\footnote{By Lemma~\ref{lemma: 
  stage 1: main}, this condition indeed holds.}. Let $c_{\texttt{Target}} > 1$ be a constant. Let $T_1$
  be the earliest time that 
  \[
    \frac{\norm{[\KA]_p}^2}{\norm{[\KA]_q}^2}
    \le c_{\texttt{Target}}, 
    \quad
    \forall p, q \in [r].
  \]
  We have $T_1 \le \poly(d)$. Moreover, throughout Stage~2, we have 
  \begin{equation}
    \label{eq: stage 2: finite-width approx infinite-width}
    \begin{aligned}
      & \max\braces{
          1 - \inprod{\ol{[\KA]_p}}{\ol{[\KB]_p}}, 
          \left| 1 - \frac{\norm{[\KA]_p}^2}{\norm{[\KB]_p}^2} \right|
        }
        \le \delta_-
      & \forall p \in [r],  \\
      & \max\braces{
          \left| 1 - \frac{\norm{[\KAxi]_p}^2}{\norm{[\KBxi]_p}^2} \right|,
          \left| 1 - \frac{\norm{[\KAxi]_p}^2}{\norm{[\KAxi]_q}^2} \right| 
        }
        \le \delta_-
      & \forall p, q \in [d - r],  \\
      & \max\braces{
          \frac{\norm{[ \K_{\C, \xi} ]_q}}{ \norm{[ \K_\D ]_p} }
          \,:\,
          \C, \D \in \{ \A, \B \}
        }
        \le \delta_{N/S}, 
      & \forall p \in [r], q \in [d - r], \\
      & \max\braces{
          \left| \inprod{ \ol{[\K_{\C}]_p} }{ \ol{[\K_{\D}]_q} } \right|
          \,:\,
          \C, \D \in \braces{ \A, \B }
        }
        \le \delta_{\AB, \perp}, 
      & \forall p \ne  q \in [r],  \\
      & \max\braces{
          \left| \inprod{ \ol{[\K_{\C}]_p} }{ \ol{[\K_{\D, \bxi}]_q} } \right|,
          \left| \inprod{ \ol{[\KAxi]_s} }{ \ol{[\KBxi]_q} } \right|
          \,:\,
          \C, \D \in \braces{ \A, \B }
        }
        \le \delta_{\xi, \perp} ,
      & \forall p \in [r], q, s \in [d-r],
    \end{aligned}
  \end{equation}
  where the $\delta$'s are some small $1 / \poly(d)$ values. 
\end{lemma}

The rest of this section is organized as follows. We derive estimations for the $\Q$-matrices in 
Section~\ref{sec: stage 2: estimations for Q}. In Section~\ref{sec: stage 2: orthogonality}, we maintain the last two 
conditions of \eqref{eq: stage 2: finite-width approx infinite-width}. In Section~\ref{sec: stage 2: KA approx KB}, 
we handle the first two conditions of \eqref{eq: stage 2: finite-width approx infinite-width}. In 
Section~\ref{sec: stage 2: noise-signal ratio}, we deal with the noise-signal ratio. We estimate the 
convergence rate in Section~\ref{sec: stage 2: convergence rate}. Finally, we prove Lemma~\ref{lemma: stage 2: main}
in Section~\ref{sec: stage 2: proof of the main lemma}.

\subsection{Estimations for $Q$}
\label{sec: stage 2: estimations for Q}

As in Stage~1, we estimate the $\Q$-matrices in this subsection. The analysis here will be more complicated than the 
one in Section~\ref{sec: stage 1: Q} since now $\tau_t^2$ is no longer close to $0$, and we cannot simply approximate 
$S_\A$ and $S_\B$ with $(1 + K)\inv$. First, we need the following lemma which gives closed-form formula for some 
expectations we will encounter later.

\begin{lemma}
  \label{lemma: stage 2: Zc}
  Define $\inprod{\z^+}{\z^-}_{\hat\bkappa^2} := \sum_{k=1}^r \hat\kappa_k^2 z^+_k z^-_k$ and 
  $T_p := \tanh\left( \frac{\hat\kappa_p^2 / d}{N_\A N_\B} \right)$, $p \in [r]$. For any $p \ne q \in [r]$, 
  we have 
  \begin{align*}
    \E_{\z^-} \braces{ \exp\left( \frac{\inprod{\z^+}{\z^-}_{\hat\bkappa^2}}{N_\A N_\B} \right) }
    &= \prod_{k=1}^r \cosh\left( \frac{\hat\kappa_k^2 / d}{N_\A N_\B} \right)
    =: Z_c, \\
    \E_{\z^-} \braces{ \exp\left( \frac{\inprod{\z^+}{\z^-}_{\hat\bkappa^2}}{N_\A N_\B} \right) z^-_p }
    &= Z_c T_p z^+_p, \\
    \E_{\z^-} \braces{ \exp\left( \frac{\inprod{\z^+}{\z^-}_{\hat\bkappa^2}}{N_\A N_\B} \right) z^-_p z^-_q}
    &= Z_c T_p T_q z^+_p z^+_q. 
  \end{align*}
\end{lemma}

Then, we derive estimations for $S_\A$ and $S_\B$. There are two types of errors we need to consider. The first 
one comes from the noises and the second one from the non-diagonalness of $\KA\trans\KB$. Similar to 
Lemma~\ref{lemma: stage 1: estimations for Q1}, we deal with them separately. The next lemma handles the 
first type of error. 

\begin{lemma}[Estimations for $S$]
  \label{lemma: stage 2: estimations for S}
  Define 
  \begin{gather*}
    E_{+, -} 
    := \exp\left( \frac{\inprod{\KA\z^+}{\KB\z^-}}{N_\A N_\B} \right), \quad 
    \tilde{E}_{+, -}
    := \exp\left( \frac{ \inprod{\z^+}{\z^-}_{\hat\bkappa^2} }{N_\A N_\B} \right), \\
    \delta_{+, \xi_-}
    := \frac{\inprod{\KA\z^+}{\KBxi\bxi_\B^-}}{N_\A N_\B}, \quad 
    \delta_{\xi+, T+}
    := \frac{
        \inprod{\KAxi\bxi_\A^+}{
          \KB \diag([T_k]_{k \in [r]}) \z^+
        }
      }{N_\A N_\B}, \\
    Z_{\A, c}
    := \E_{\z^-} E_{+, -}, \quad
    Z_{\B, c}
    := \E_{\z^-} E_{-, +}, \\ 
    \tilde{S}_\A 
    := \frac{E_{+, +}}{E_{+, +} + K Z_{\A, c}}, \quad
    \tilde{S}_\B 
    := \frac{E_{+, +}}{E_{+, +} + K Z_{\B, c}}, \quad 
    \tilde{S}
    := \frac{\tilde{E}_{+, +}}{\tilde{E}_{+, +} + K Z_c}. 
  \end{gather*}
  In Stage~2, we have 
  \begin{align*}
    S_\A 
    &= \tilde{S}_\A 
      + \tilde{S} (1 - \tilde{S})
      \left( \delta_{+, \xi+} + \delta_{\xi+, +} - \delta_{\xi+, T+} \right) 
      \pm O\left( 
        d^3 
        \left( \delta_{\AB, \perp} + \delta_{N/S} \right) 
        \delta_{\xi, \perp} 
        \delta_{N/S} 
      \right), \\
    S_\B 
    &= \tilde{S}_\B
      + \tilde{S} ( 1 - \tilde{S} )
      \left( 
        \delta_{+, \bxi+} + \delta_{\bxi+, +} 
        - \delta_{T+, \bxi+} 
      \right)
      \pm O\left( 
        d^3 
        \left( \delta_{\AB, \perp} + \delta_{N/S} \right) 
        \delta_{\xi, \perp} 
        \delta_{N/S} 
      \right), 
  \end{align*}
  and 
  \begin{align*}
    \frac{ S_\A(\xA^+, \xB^+) \exp(\fA^+ \cdot \fB^-) }{ \exp(\fA^+ \cdot \fB^+) } 
    &= \frac{ \tilde{S}_\A E_{+, -} }{ E_{+, +} }
      + \frac{ \tilde{S} \tilde{E}_{+, -} }{ \tilde{E}_{+, +} }
      \left(
        \delta_{+, \bxi-} + \delta_{\bxi+, -} 
        - \tilde{S} \left( \delta_{+, \bxi+} + \delta_{\bxi+, +} \right)
        - ( 1 - \tilde{S} ) \delta_{\bxi+, T+} 
      \right) \\ 
      &\qquad
      \pm O\left( 
        d^3 
        \left( \delta_{\AB, \perp} + \delta_{N/S} \right) 
        \delta_{\xi, \perp} 
        \delta_{N/S} 
      \right), \\ 
    \frac{ S_\B(\xA^+, \xB^+) \exp(\fA^- \cdot \fB^+) }{ \exp(\fA^+ \cdot \fB^+) } 
    &= \frac{ \tilde{S}_\B E_{-, +} }{ E_{+, +} }
      + \frac{ \tilde{S} \tilde{E}_{-, +} }{ \tilde{E}_{+, +} }
      \left(
        1
        + \delta_{\bxi-, +} + \delta_{-, \bxi+} 
        - \tilde{S} \left( \delta_{+, \bxi+} + \delta_{\bxi+, +} \right)
        - ( 1 - \tilde{S} ) \delta_{T+, \bxi+} 
      \right) \\
      &\qquad
      \pm O\left( 
        d^3 
        \left( \delta_{\AB, \perp} + \delta_{N/S} \right) 
        \delta_{\xi, \perp} 
        \delta_{N/S} 
      \right). 
  \end{align*}
\end{lemma}

Then, we consider the error comes from the non-diagonalness of $\KA\trans\KB$. 
\begin{lemma}[Further estimations for $S$]
  \label{lemma: stage 2: further estimations for S}
  Define 
  \begin{gather*}
    \tilde{\delta}_{+, -}
    = \sum_{i \ne j} \frac{ \inprod{[\KA]_i}{[\KB]_j} }{N_\A N_\B} z^+_i z^-_j, \quad 
    \tilde{\delta}_{+, T+}
    = \sum_{i \ne j} \frac{ \inprod{[\KA]_i}{[\KB]_j} }{N_\A N_\B} z^+_i T_j z^+_j, \\ 
    \tilde{E}_0
    := \tilde{E}_{+, +}
    = \tilde{E}_{-, -}
    = \exp\left( \frac{\norm{\hat\bkappa}^2}{N_\A N_\B} \right). 
  \end{gather*}
  In Stage~2, we have 
  \begin{align*}
    \tilde{S}_\A(\z^+)
    &= \tilde{S}
      \left(
        1 
        + ( 1 - \tilde{S} ) ( \tilde\delta_{+, +} - \tilde\delta_{+, T+} )
      \right)
      \pm O\left( d^2 \delta_{\AB, \perp}^2 \right), \\
    \tilde{S}_\B(\z^+)
    &= \tilde{S}
      \left(
        1 
        + ( 1 - \tilde{S} ) ( \tilde\delta_{+, +} - \tilde\delta_{T+, +} )
      \right)
      \pm O\left( d^2 \delta_{\AB, \perp}^2 \right), \\
    \frac{ \tilde{S}_\A E_{+, -} }{ E_{+, +} }
    &= \frac{ \tilde{S} \tilde{E}_{+, -}  }{ \tilde{E}_0 }
      \left(
        1 
        - \tilde{S} \tilde\delta_{+, +} 
        - ( 1 - \tilde{S} ) \tilde\delta_{+, T+}
        + \tilde\delta_{+, -} 
      \right)
      \pm O\left( d^2 \delta_{\AB, \perp}^2 \right), \\
    \frac{ \tilde{S}_\B E_{-, +} }{ E_{+, +} }
    &= \frac{ \tilde{S} \tilde{E}_{-, +} }{ \tilde{E}_0 }
      \left(
        1 
        - \tilde{S} \tilde\delta_{+, +} 
        - ( 1 - \tilde{S} ) \tilde\delta_{T+, +}
        + \tilde\delta_{-, +} 
      \right)
      \pm O\left( d^2 \delta_{\AB, \perp}^2 \right). 
  \end{align*}
\end{lemma}

With the above two lemmas in hand, we can now derive estimations for the $\Q$-matrices. 

\begin{lemma}[Estimations for $\Q_1$]
  \label{lemma: stage 2: estimations for Q1}
  Define $\KAB = \KA\trans\KB$ and $\KBA = \KB\trans\KA$. 
  In Stage~2, for any $p \ne q \in [r]$, we have  
  \begin{align*}
    [\Q_1]_{p, p}
    &= 2 (1 - \tilde{S}) (1 - T_p)
      \pm O\left( d^2 \delta_{\AB, \perp}^2 \right), \\
    [\Q_1]_{p, q}
    &= - \tilde{S} ( 1 - \tilde{S} )
      \frac{ [\KAB]_{p, q} + [\KBA]_{q, p} }{ N_\A N_\B d} (2 - T_p - T_q )  \\
      &\qquad
      - (1 - \tilde{S}) 
      \frac{ [\KAB]_{p, q} }{N_\A N_\B d}
      \tilde{S} \left( 2 T_p T_q - T_p - T_q \right) \\
      &\qquad
      - (1 - \tilde{S}) \frac{ [\KAB]_{q, p} }{N_\A N_\B d}
      \left(
        2 
        - \tilde{S} ( T_p + T_q)  
        - ( 1 - \tilde{S} ) ( T_p^2 + T_q^2 )
      \right) 
      \pm O\left( d^2 \delta_{\AB, \perp}^2 \right). 
  \end{align*}
  In particular, we have 
  \[
    |[\Q_1]_{p, q}|
    \le O\left( \frac{\kappa_0 \delta_{\AB, \perp} }{d} \right)
    \le O\left( \frac{ \delta_{\AB, \perp} }{\sqrt{d}} \right).
  \]
\end{lemma}

\begin{lemma}[Estimations for $\Q_{1, \xi}$]
  \label{lemma: stage 2: estimations for Q1xi}
  In Stage 2, for any $p \in [d - r]$ and $q \in [r]$, we have 
  \begin{align*}
    [ \Q_{1, \bxi_\A} ]_{p, q}
    &= -( 1 - \tilde{S} )
      \left(
        1 + \tilde{S} 
        + (1 - \tilde{S}) T_q^2  
      \right)
      \frac{ \inprod{ [\KB]_q }{ [\KAxi]_p } }{ N_\A N_\B d}
      \pm O\left( d^3 \left( \delta_{\AB, \perp} + \delta_{N/S} \right) \delta_{\xi, \perp} \delta_{N/S} \right), \\
    [ \Q_{1, \bxi_\B} ]_{p, q}
    &= -( 1 - \tilde{S} )
      \left(
        1 + \tilde{S} 
        + (1 - \tilde{S}) T_q^2  
      \right)
      \frac{ \inprod{ [\KA]_q }{ [\KBxi]_p } }{ N_\A N_\B d}
      \pm O\left( d^3 \left( \delta_{\AB, \perp} + \delta_{N/S} \right) \delta_{\xi, \perp} \delta_{N/S} \right). 
  \end{align*}
  In particular, we have 
  \[
    \max\braces{ [ \Q_{1, \bxi_\A} ]_{p, q}, [ \Q_{1, \bxi_\B} ]_{p, q} }
    \le O\left( \delta_{N/S} \delta_{\xi, \perp} \right).
  \]
\end{lemma}

\begin{lemma}[Estimations for $\Q_2$]
  \label{lemma: stage 2: estimations for Q2}
  In Stage~2, for any $p, q \in [d - r]$, we have 
  \[
    [\Q_2]_{p, q}
    = \pm O\left( d^3 \left( \delta_{\AB, \perp} + \delta_{N/S} \right) \delta_{\xi, \perp} \delta_{N/S} \right). 
  \]
\end{lemma}

\begin{lemma}[Estimations for $Q_0$]
  \label{lemma: stage 2: estimations for Q0}
  In Stage~2, we have 
  \[ 
    Q_0 
    = - \sum_{k=1}^r \frac{\kappa_k^2}{\norm{\bkappa}^2} [\Q_1]_{k, k}
      \pm O\left( 
        d^2 \delta_{\AB, \perp}^2 
        + \delta_- 
        + \kappa_0 d \delta_{N/S} \delta_{\xi, \perp}
      \right). 
  \] 
\end{lemma}

Finally, we use these estimations to simplify the formulas for the norms. We do not consider the 
tangent movement here since the situation is trickier there, and we will handle them in later subsections. 

\begin{corollary}[Dynamics of the norms]
  \label{cor: stage 2: d norm}
  In Stage~2, we have 
  \begin{align*}
    \frac{\rd}{\rd t} \norm{[\KA]_p}^2
    &= \frac{2 \sigma_p^2 \norm{[\KA]_p} \norm{[\KB]_p}}{N_\A N_\B d} 
      \inprod{\ol{[\KA]_p}}{\ol{[\KB]_p}} [\Q_1]_{p, p} 
      + 2 \frac{\norm{[\KA]_p}^2}{N_\A^2 d} Q_0 \sigma_p^2 \\ 
    &\qquad
    \pm O\left(  
      \frac{\sigma_p^2 \kappa_p^2}{N_\A N_\B d} 
      \kappa_0 d
      \delta_{\AB, \perp}^2 
    \right) \\ 
    \frac{\rd}{\rd t} \norm{[\KAxi]_q}^2
    &= \frac{2 \norm{[\KAxi]_q}^2}{N_\A^2 d} Q_0 \sigma_\xi^2
      \pm O\left( 
        \frac{\sigma_\xi^2  \norm{[\KAxi]_q}^2}{ N_\A N_\B d}  
        d \delta_{\xi, \perp}^2
      \right). 
  \end{align*}
  The formulas for $\norm{[\KB]_p}$ and $\norm{[\KBxi]_q}$ can be obtained by interchanging the roles of 
  $\A$ and $\B$. 
\end{corollary}

\subsubsection*{Omitted proofs of this subsection}

\begin{proof}[Proof of Lemma~\ref{lemma: stage 2: Zc}]
  First, we compute 
  \begin{align*}
    \E_{\z^-} \braces{ \exp\left( \frac{\inprod{\z^+}{\z^-}_{\hat\bkappa^2}}{N_\A N_\B} \right) }
    = \prod_{k=1}^r 
      \E_{z^-_k} \braces{ \exp\left( \frac{ \hat\kappa_k^2 z^+_k z^-_k }{N_\A N_\B} \right) }
    &= \prod_{k=1}^r 
      \frac{1}{2}
      \left( 
        \exp\left( \frac{ \hat\kappa_k^2 / d }{N_\A N_\B} \right)
        + \exp\left( - \frac{ \hat\kappa_k^2 / d }{N_\A N_\B} \right)
      \right)  \\
    &= \prod_{k=1}^r \cosh\left( \frac{ \hat\kappa_k^2 / d }{N_\A N_\B} \right). 
  \end{align*}
  Similarly, we also have 
  \[ 
    \E_{\z^-} \braces{ \exp\left( \frac{\inprod{\z^+}{\z^-}_{\hat\bkappa^2}}{N_\A N_\B} \right) z^-_p }
    = \E_{z^-_p} \braces{ \exp\left( \frac{ \hat\kappa_k^2 z^+_p z^-_p }{N_\A N_\B} \right) z^-_p }
      \prod_{k \ne p} 
      \E_{z^-_k} \braces{ \exp\left( \frac{ \hat\kappa_k^2 z^+_k z^-_k }{N_\A N_\B} \right) }. 
  \] 
  Each factor in $\prod_{k \ne p}$ is still $\cosh\left( \frac{ \hat\kappa_k^2 / d }{N_\A N_\B} \right)$. 
  For the first term, we have 
  \begin{align*}
    \E_{z^-_p} \braces{ \exp\left( \frac{ \hat\kappa_k^2 z^+_p z^-_p }{N_\A N_\B} \right) z^-_p }
    &= \frac{1}{2 \sqrt{d}} 
      \E_{z^-_p} \braces{ \exp\left( \frac{ \hat\kappa_k^2 z^+_p / \sqrt{d} }{N_\A N_\B} \right) }
      - \frac{1}{2 \sqrt{d}} 
        \E_{z^-_p} \braces{ \exp\left( \frac{ - \hat\kappa_k^2 z^+_p / \sqrt{d} }{N_\A N_\B} \right) z^-_p } \\
    &= \frac{1}{\sqrt{d}} 
      \sinh\left( \frac{ \hat\kappa_k^2 z^+_p / \sqrt{d} }{N_\A N_\B} \right)
    = \frac{\sgn z^+_o}{\sqrt{d}} 
      \sinh\left( \frac{ \hat\kappa_k^2 / d }{N_\A N_\B} \right)
    = \sinh\left( \frac{ \hat\kappa_k^2 / d }{N_\A N_\B} \right) z^+_p.
  \end{align*}
  Therefore, 
  \[
    \E_{\z^-} \braces{ \exp\left( \frac{\inprod{\z^+}{\z^-}_{\hat\bkappa^2}}{N_\A N_\B} \right) z^-_p }
    = z^+_p \sinh\left( \frac{ \hat\kappa_k^2 / d }{N_\A N_\B} \right) 
      \prod_{k \ne p} \cosh\left( \frac{ \hat\kappa_k^2 / d }{N_\A N_\B} \right)
    = Z_c T_p z^+_p. 
  \]
  The above calculation, \textit{mutatis mutandis}, also yields the last identity. 
\end{proof}

\begin{proof}[Proof of Lemma~\ref{lemma: stage 2: estimations for S}]
  First, we write 
  \begin{align*}
    \inprod{\fA^+}{\fB^-}
    &= \frac{\inprod{\KA\z^+ + \KAxi\bxi_\A^+}{\KB\z^- + \KBxi\bxi_\B^-}}{N_\A N_\B} \\
    &= \frac{\inprod{\KA\z^+}{\KB\z^-}}{N_\A N_\B} 
      + \frac{\inprod{\KA\z^+}{\KBxi\bxi_\B^-}}{N_\A N_\B} 
      + \frac{\inprod{\KAxi\bxi_\A^+}{\KB\z^-}}{N_\A N_\B} 
      \pm O\left( d^2 \delta_{\xi, \perp} \delta_{N/S}^2 \right) . 
  \end{align*}
  Also note that the middle terms are $O( d^2 \delta_{\xi, \perp} \delta_{N/S} )$. Then, we compute 
  \[ 
    \exp\left( \fA^+ \cdot \fB^- \right)
    = \exp\left(
        1
        + \frac{\inprod{\KA\z^+}{\KBxi\bxi_\B^-}}{N_\A N_\B} 
        + \frac{\inprod{\KAxi\bxi_\A^+}{\KB\z^-}}{N_\A N_\B} 
      \right)
      \pm O\left( d^2 \delta_{\xi, \perp} \delta_{N/S}^2 \right). 
  \] 
  Similar results also hold for other combinations of $\pm$. With the notations defined in this lemma, 
  we can write these results as 
  \begin{align*}
    \exp\left( \fA^+ \cdot \fB^- \right)
    &= E_{+, -} \left( 1 + \delta_{+, \xi-} + \delta_{\xi+, -} \right)
      \pm O\left( d^2 \delta_{\xi, \perp} \delta_{N/S}^2 \right), \\ 
    \exp\left( \fA^- \cdot \fB^+ \right)
    &= E_{-, +} \left( 1 + \delta_{-, \xi+} + \delta_{\xi-, +} \right)
      \pm O\left( d^2 \delta_{\xi, \perp} \delta_{N/S}^2 \right), \\ 
    \exp\left( \fA^+ \cdot \fB^+ \right)
    &= E_{+, +} \left( 1 + \delta_{+, \xi+} + \delta_{\xi+, +} \right)
      \pm O\left( d^2 \delta_{\xi, \perp} \delta_{N/S}^2 \right). 
  \end{align*}
  To compute $S_\A$ and $S_\B$, we then need to take expectations over the negative examples. Note that by 
  taking expectation over $\bxi_\B^-$, the term $E_{+, -} \delta_{+, \xi-}$ becomes $0$. Therefore, we have 
  \[
    \E_{\xB^-}\exp\left( \fA^+ \cdot \fB^- \right)
    = \E_{\xB^-} \braces{ E_{+, -} \left( 1 + \delta_{\xi+, -} \right) }
      \pm O\left( d^2 \delta_{\xi, \perp} \delta_{N/S}^2 \right). 
  \]
  Unfortunately, the same argument does not apply to $\delta_{\xi+, -}$ since both $E_{+, -}$ and 
  $\delta_{\xi+, -}$ depend on $\z^-$. However, it is still possible to further simplify the expression. 
  First, we write 
  \[
    \E_{\z^-} \braces{ E_{+, -} \delta_{\xi+, -} }
    = \E_{\z^-} \braces{ \tilde{E}_{+, -} \left( 1 \pm O(d \delta_{\AB, \perp})  \right) \delta_{\xi+, -} }
    = \E_{\z^-} \braces{ \tilde{E}_{+, -} \delta_{\xi+, -} }
      \pm O(d^3 \delta_{\AB, \perp} \delta_{\xi, \perp} \delta_{N/S} ). 
  \]
  Recall Lemma~\ref{lemma: stage 2: Zc}. Then, we compute 
  \begin{align*}
    \E_{\z^-} \braces{ \tilde{E}_{+, -} \delta_{\xi+, -} }
    = \E_{\z^-} \braces{ 
        \tilde{E}_{+, -} 
        \frac{\inprod{\KAxi\bxi_\A^+}{\KB\z^-}}{N_\A N_\B}
      }
    &= \frac{
        \inprod{\KAxi\bxi_\A^+}{
          \KB \E_{\z^-} \braces{ \tilde{E}_{+, -} \z^- }
        }
      }{N_\A N_\B} \\
    &= Z_c
      \frac{
        \inprod{\KAxi\bxi_\A^+}{
          \KB \diag([T_k]_{k \in [r]}) \z^+
        }
      }{N_\A N_\B} 
    = Z_c \delta_{\xi+, T+}. 
  \end{align*}
  Hence, 
  \begin{align*}
    \E_{\xB^-}\exp\left( \fA^+ \cdot \fB^- \right)
    &= \E_{\xB^-} \braces{ E_{+, -} }
      + \E_{\xB^-} \braces{ E_{+, -}\delta_{\xi+, -}  }
      \pm O\left( d^2 \delta_{\xi, \perp} \delta_{N/S}^2 \right) \\
    &= Z_{\A, c}
      + Z_c \delta_{\xi+, T+}
      \pm O(d^3 \delta_{\AB, \perp} \delta_{\xi, \perp} \delta_{N/S} ) 
      \pm O\left( d^2 \delta_{\xi, \perp} \delta_{N/S}^2 \right) \\
    &= Z_{\A, c}
      + Z_c \delta_{\xi+, T+}
      \pm O\left( 
        d^3 
        \left( \delta_{\AB, \perp} + \delta_{N/S} \right) 
        \delta_{\xi, \perp} 
        \delta_{N/S} 
      \right). 
  \end{align*}
  Similarly, we also have 
  \[
    \E_{\xA^-} \exp\left( \fA^+ \cdot \fB^- \right)
    = Z_{\B, c}
      + Z_c \delta_{T+, \xi+}
      \pm O\left( 
        d^3 
        \left( \delta_{\AB, \perp} + \delta_{N/S} \right) 
        \delta_{\xi, \perp} 
        \delta_{N/S} 
      \right). 
  \]
  Recall that $\exp\left( \fA^+ \cdot \fB^+ \right) = E_{+, +} \left( 1 + \delta_{+, \xi+} + \delta_{\xi+, +} \right)
  \pm O\left( d^2 \delta_{\xi, \perp} \delta_{N/S}^2 \right)$. Hence, we have 
  \begin{align*}
    S_\A(\xA^+, \xB^+)
    &= \frac{ E_{+, +} ( 1 + \delta_{+, \xi+} + \delta_{\xi+, +} ) }{
        E_{+, +} ( 1 + \delta_{+, \xi+} + \delta_{\xi+, +} )
        + K Z_{\A, c}
        + K Z_c \delta_{\xi+, T+}
      } 
      \pm O\left( 
        d^3 
        \left( \delta_{\AB, \perp} + \delta_{N/S} \right) 
        \delta_{\xi, \perp} 
        \delta_{N/S} 
      \right) \\
    &= \tilde{S}_\A 
      \left(
        1
        - \tilde{S}_\A ( \delta_{+, \xi+} + \delta_{\xi+, +} )
        - (1 - \tilde{S}_\A) \delta_{\xi+, T+}
      \right) 
      \\
      &\qquad
      + \tilde{S}_\A ( \delta_{+, \xi+} + \delta_{\xi+, +} )
      \pm O\left( 
        d^3 
        \left( \delta_{\AB, \perp} + \delta_{N/S} \right) 
        \delta_{\xi, \perp} 
        \delta_{N/S} 
      \right) \\
    &= \tilde{S}_\A 
      + \tilde{S} (1 - \tilde{S})
      \left( \delta_{+, \xi+} + \delta_{\xi+, +} - \delta_{\xi+, T+} \right) 
      \pm O\left( 
        d^3 
        \left( \delta_{\AB, \perp} + \delta_{N/S} \right) 
        \delta_{\xi, \perp} 
        \delta_{N/S} 
      \right). 
  \end{align*}
  Similarly, we also have 
  \[
    S_\B(\xA^+, \xB^+)
    = \tilde{S}_\B
      + \tilde{S} ( 1 - \tilde{S} )
      \left( 
        \delta_{+, \bxi+} + \delta_{\bxi+, +} 
        - \delta_{T+, \bxi+} 
      \right)
      \pm O\left( 
        d^3 
        \left( \delta_{\AB, \perp} + \delta_{N/S} \right) 
        \delta_{\xi, \perp} 
        \delta_{N/S} 
      \right). 
  \]
  Then, we compute 
  \begin{align*}
    \frac{ S_\A(\xA^+, \xB^+) \exp(\fA^+ \cdot \fB^-) }{ \exp(\fA^+ \cdot \fB^+) } 
    &= \left(
        \tilde{S}_\A
        + \tilde{S} ( 1 - \tilde{S} )
        \left( 
          \delta_{+, \bxi+} + \delta_{\bxi+, +} 
          - \delta_{\bxi+, T+} 
        \right)
      \right)
      \frac{ 
        E_{+, -} \left( 1 + \delta_{+, \bxi-} + \delta_{\bxi+, -} \right)
      }{ 
        E_{+, +} \left( 1 + \delta_{+, \bxi+} + \delta_{\bxi+, +} \right)
      } \\ 
      &\qquad
      \pm O\left( 
        d^3 
        \left( \delta_{\AB, \perp} + \delta_{N/S} \right) 
        \delta_{\xi, \perp} 
        \delta_{N/S} 
      \right) \\
    &= \frac{ \tilde{S}_\A E_{+, -} }{ E_{+, +} }
      \left(
        1
        + \delta_{+, \bxi-} + \delta_{\bxi+, -} 
        - \tilde{S} \left( \delta_{+, \bxi+} + \delta_{\bxi+, +} \right)
        - ( 1 - \tilde{S} ) \delta_{\bxi+, T+} 
      \right)  \\
      &\qquad
      \pm O\left( 
        d^3 
        \left( \delta_{\AB, \perp} + \delta_{N/S} \right) 
        \delta_{\xi, \perp} 
        \delta_{N/S} 
      \right) \\
    &=
      \frac{ \tilde{S}_\A E_{+, -} }{ E_{+, +} }
      + \frac{ \tilde{S} \tilde{E}_{+, -} }{ \tilde{E} }
      \left(
        \delta_{+, \bxi-} + \delta_{\bxi+, -} 
        - \tilde{S} \left( \delta_{+, \bxi+} + \delta_{\bxi+, +} \right)
        - ( 1 - \tilde{S} ) \delta_{\bxi+, T+} 
      \right) \\ 
      &\qquad
      \pm O\left( 
        d^3 
        \left( \delta_{\AB, \perp} + \delta_{N/S} \right) 
        \delta_{\xi, \perp} 
        \delta_{N/S} 
      \right), 
  \end{align*}
  and 
  \begin{align*}
    \frac{ S_\B(\xA^+, \xB^+) \exp(\fA^- \cdot \fB^+) }{ \exp(\fA^+ \cdot \fB^+) } 
    &=
      \frac{ \tilde{S}_\B E_{-, +} }{ E_{+, +} }
      + \frac{ \tilde{S} \tilde{E}_{-, +} }{ \tilde{E} }
      \left(
        1
        + \delta_{\bxi-, +} + \delta_{-, \bxi+} 
        - \tilde{S} \left( \delta_{+, \bxi+} + \delta_{\bxi+, +} \right)
        - ( 1 - \tilde{S} ) \delta_{T+, \bxi+} 
      \right) \\
      &\qquad
      \pm O\left( 
        d^3 
        \left( \delta_{\AB, \perp} + \delta_{N/S} \right) 
        \delta_{\xi, \perp} 
        \delta_{N/S} 
      \right). 
  \end{align*}
\end{proof}

\begin{proof}[Proof of Lemma~\ref{lemma: stage 2: further estimations for S}]
  We write 
  \[
    \frac{\inprod{\KA\z^+}{\KB\z^-}}{N_\A N_\B} 
    = \sum_{k=1}^r \frac{ \inprod{[\KA]_k}{[\KB]_k} }{N_\A N_\B} z^+_k z^-_k
      + \sum_{i \ne j} \frac{ \inprod{[\KA]_i}{[\KB]_j} }{N_\A N_\B} z^+_i z^-_j
    =: I_{+, -} + \tilde{\delta}_{+, -}. 
  \]
  Note that, as a special case, we have $I_{+, +} = I_{-, -} = I_0$. In other words, $I_{+, +}$ and $I_{-, -}$ 
  do not depend on the actual value of $\z^\pm$. Also note that $I_{+, -}$ is bounded by $O( d \delta_{\AB, \perp} )$.
  Then, we compute 
  \[
    E_{+, -} 
    = \exp(I_{+, -}) \left(
        1 + \tilde\delta_{+, -} \pm O\left( d^2 \delta_{\AB, \perp}^2 \right) 
      \right)
    \quad\text{and}\quad
    E_{+, +} 
    = \exp(I_0) \left( 
        1 + \tilde\delta_{+, +} \pm O\left( d^2 \delta_{\AB, \perp}^2 \right)  
      \right). 
  \]
  Take expectation over $\z^-$ and we obtain 
  \[
    \E_{\z^-} E_{+, -} 
    =  \E_{\z^-} \exp(I_{+, -}) 
      + \E_{\z^-} \braces{ \exp(I_{+, -})  \tilde\delta_{+, -} }
      \pm O\left( d^2 \delta_{\AB, \perp}^2 \right) .
  \]
  By Lemma~\ref{lemma: stage 2: Zc}, the first term is $Z_c$. For the second term, we compute 
  \begin{align*}
    \E_{\z^-} \braces{ \exp(I_{+, -})  \tilde\delta_{+, -} }
    &= \sum_{i \ne j} \frac{[\KAB]_{i, j}}{N_\A N_\B}
      \E_{\z^-} \braces{ \exp(I_{+, -}) z^-_j } z^+_i \\
    &= Z_c \sum_{i \ne j} \frac{[\KAB]_{i, j}}{N_\A N_\B} T_j z^+_j z^+_i \\
    &= Z_c \tilde\delta_{+, T+}, 
  \end{align*}
  where the second line again comes from Lemma~\ref{lemma: stage 2: Zc}. Hence, we have 
  \[
    \E_{\z^-} E_{+, -} 
    = Z_c
      + Z_c \tilde\delta_{+, T+}
      \pm O\left( d^2 \delta_{\AB, \perp}^2 \right) .
  \]
  Then, for $\tilde{S}_\A$, we have 
  \begin{align*}
    \tilde{S}_\A(\z^+)
    = \frac{ E_{+, +} }{ E_{+, +} + K \E_{\z^-} E_{+, -}  } 
    &= \frac{ 
        \exp(I_0) \left( 1 + \tilde\delta_{+, +}  \right)
      }{ 
        \exp(I_0) \left( 1 + \tilde\delta_{+, +}  \right) 
        + K Z_c
        + K Z_c \tilde\delta_{+, T+}
      } 
      \pm O\left( d^2 \delta_{\AB, \perp}^2 \right) \\
    &= \tilde{S}
      \left(
        1 
        - \tilde{S} \tilde\delta_{+, +} 
        - (1 - \tilde{S}) \tilde\delta_{+, T+}
      \right)
      + \tilde{S} \tilde\delta_{+, +}  
      \pm O\left( d^2 \delta_{\AB, \perp}^2 \right) \\
    &= \tilde{S}
      \left(
        1 
        + ( 1 - \tilde{S} ) ( \tilde\delta_{+, +} - \tilde\delta_{+, T+} )
      \right)
      \pm O\left( d^2 \delta_{\AB, \perp}^2 \right). 
  \end{align*}
  Similarly, we also have 
  \[
    \tilde{S}_\B(\z^+)
    = \tilde{S}
      \left(
        1 
        + ( 1 - \tilde{S} ) ( \tilde\delta_{+, +} - \tilde\delta_{T+, +} )
      \right)
      \pm O\left( d^2 \delta_{\AB, \perp}^2 \right). 
  \]
  Then, we compute 
  \begin{align*}
    \frac{ \tilde{S}_\A E_{+, -} }{ E_{+, +} }
    &= \tilde{S}
      \left(
        1 
        + ( 1 - \tilde{S} ) ( \tilde\delta_{+, +} - \tilde\delta_{+, T+} )
      \right)
      \frac{ 
        \exp(I_{+, -}) \left( 1 + \tilde\delta_{+, -} \right)
      }{ 
        \exp(I_{+, -}) \left( 1 + \tilde\delta_{+, +}  \right)
      }
      \pm O\left( d^2 \delta_{\AB, \perp}^2 \right) \\
    &= \frac{ \tilde{S} \exp(I_{+, -}) }{ \exp(I_0) }
      \left(
        1 
        - \tilde{S} \tilde\delta_{+, +} 
        - ( 1 - \tilde{S} ) \tilde\delta_{+, T+}
        + \tilde\delta_{+, -} 
      \right)
      \pm O\left( d^2 \delta_{\AB, \perp}^2 \right). 
  \end{align*}
  Similarly, we also have 
  \[
    \frac{ \tilde{S}_\B E_{-, +} }{ E_{+, +} }
    = \frac{ \tilde{S} \exp(I_{-, +}) }{ \exp(I_0) }
      \left(
        1 
        - \tilde{S} \tilde\delta_{+, +} 
        - ( 1 - \tilde{S} ) \tilde\delta_{T+, +}
        + \tilde\delta_{-, +} 
      \right)
      \pm O\left( d^2 \delta_{\AB, \perp}^2 \right). 
  \]
\end{proof}

\begin{proof}[Proof of Lemma~\ref{lemma: stage 2: estimations for Q1}]
  Recall that 
  \begin{align*}
    \Q_1
    &:= \E \braces{
        \left( 2 - S_\A(\xA^+, \xB^+) - S_\B(\xA^+, \xB^+) \right)
        \z^+(\z^+)\trans d
      } \\
      &\qquad
      - K \E \braces{
        \frac{ S_\A(\xA^+, \xB^+) \exp( \fA^+ \cdot \fB^-)}{\exp( \fA^+ \cdot \fB^+)}
        \z^- (\z^+)\trans d
      } 
      - K \E \braces{
        \frac{S_\B(\xA^+, \xB^+) \exp( \fA^- \cdot \fB^+)}{\exp( \fA^+ \cdot \fB^+)}
        \z^+ (\z^-)\trans d
      }. 
  \end{align*}
  Note that there is no $\bxi$ here other than the ones in the coefficient. As a result, all terms 
  contain $\delta_{+, \bxi+}$ and alike are $0$. Hence, we have 
  \begin{align*}
    \Q_1
    &:= 
      \E \braces{
        \left( 2 - \tilde{S}_\A(\xA^+, \xB^+) - \tilde{S}_\B(\xA^+, \xB^+) \right)
        \z^+(\z^+)\trans d
      } \\
      &\qquad
      - K \E \braces{
        \frac{\tilde{S}_\A E_{+, -}}{E_{+, +}}
        \z^- (\z^+)\trans d
      } 
      - K \E \braces{
        \frac{\tilde{S}_\B E_{-, +}}{E_{+, +}}
        \z^+ (\z^-)\trans d
      } \\
    &=: \Term_1(\Q_1) + \Term_2(\Q_1) + \Term_3(\Q_1). 
  \end{align*}
  Now we estimate each of these three terms separately. We also deal with the diagonal and 
  off-diagonal terms separately. By Lemma~\ref{lemma: stage 2: further estimations for S}, we have
  \begin{align*}
    \Term_1([\Q_1]_{p, p})
    &= \E \braces{
        2 - \tilde{S}_\A(\xA^+, \xB^+) - \tilde{S}_\B(\xA^+, \xB^+) 
      } \\
    &= \E \braces{
        2 
        - \tilde{S}
        \left(
          1 
          + ( 1 - \tilde{S} ) ( \tilde\delta_{+, +} - \tilde\delta_{+, T+} )
        \right)
        - \tilde{S}
        \left(
          1 
          + ( 1 - \tilde{S} ) ( \tilde\delta_{+, +} - \tilde\delta_{T+, +} )
        \right)
      } 
      \pm O\left( d^2 \delta_{\AB, \perp}^2 \right) \\
    &= 2 (1 - \tilde{S}) \pm O\left( d^2 \delta_{\AB, \perp}^2 \right). 
  \end{align*}
  Note that we use the fact that all these $\delta$'s have mean $0$. Also by Lemma~\ref{lemma: stage 2: 
  further estimations for S}, we have 
  \begin{align*}
    \Term_2( [\Q_1]_{p, p} )
    &= - K \E \braces{
        \left(
          \frac{ \tilde{S} \tilde{E}_{+, -} }{ \tilde{E}_0 }
          \left(
            1 
            - \tilde{S} \tilde\delta_{+, +} 
            - ( 1 - \tilde{S} ) \tilde\delta_{+, T+}
            + \tilde\delta_{+, -} 
          \right)
        \right)
        z^-_p z^+_p d
      } 
      \pm O\left( d^2 \delta_{\AB, \perp}^2 \right) \\
    &= - \frac{ \tilde{S} K }{ \tilde{E}_0 }
      \E \braces{
        \tilde{E}_{+, -}
        z^-_p z^+_p d
      }  \\
      &\qquad
      - \frac{ \tilde{S} K }{ \tilde{E}_0 }
      \E \braces{
        \left(
          \tilde{E}_{+, -}
          \left(
            - \tilde{S} \tilde\delta_{+, +} 
            - ( 1 - \tilde{S} ) \tilde\delta_{+, T+}
            + \tilde\delta_{+, -} 
          \right)
        \right)
        z^-_p z^+_p d
      } 
      \pm O\left( d^2 \delta_{\AB, \perp}^2 \right). 
  \end{align*}
  Note that $\tilde\delta_{+, +}$, $\tilde\delta_{+, -}$ and $\tilde\delta_{+, T+}$ only contain 
  cross terms of form $z^+_i z^\pm_j$ with $i \ne j$. Hence, the second term is $0$. Meanwhile, by Lemma~\ref{lemma: 
  stage 2: Zc}, the first term is 
  \[
    - \frac{ \tilde{S} K }{ \tilde{E}_0 }
    \E \braces{
      \tilde{E}_{+, -}
      z^-_p z^+_p d
    }
    = - \frac{ \tilde{S} K Z_c T_p  }{ \tilde{E}_0 }
    = - (1 - \tilde{S}) T_p .
  \]
  As a result, 
  \[
    \Term_2( [\Q_1]_{p, p} ) 
    = - (1 - \tilde{S}) T_p \pm O\left( d^2 \delta_{\AB, \perp}^2 \right). 
  \]
  Similarly, one can show that $\Term_3( [\Q_1]_{p, p} ) = - (1 - \tilde{S}) T_p 
  \pm O\left( d^2 \delta_{\AB, \perp}^2 \right)$ also holds. Thus, 
  \[
    [\Q_1]_{p, p}
    = 2 (1 - \tilde{S}) (1 - T_p)
      \pm O\left( d^2 \delta_{\AB, \perp}^2 \right). 
  \]

  Now, we consider the off-diagonal terms. For notational simplicity, we define $\KAB = \KA\trans\KB$. 
  For any $p \ne q$, we compute 
  \begin{align*}
    \Term_1([\Q_1]_{p, q})
    &= -
      \E \braces{
        \left( \tilde{S}_\A(\xA^+, \xB^+) + \tilde{S}_\B(\xA^+, \xB^+) \right)
        z^+_p z^+_q d
      } \\
    &= - \tilde{S} ( 1 - \tilde{S} )
      \E \braces{
        \left( 
          2 \tilde\delta_{+, +} - \tilde\delta_{+, T+} - \tilde\delta_{T+, T} 
        \right)
        z^+_p z^+_q d
      } 
      \pm O\left( d^2 \delta_{\AB, \perp}^2 \right) \\
    &= - \tilde{S} ( 1 - \tilde{S} )
      \sum_{i \ne j} \frac{ [\KAB]_{i, j} }{ N_\A N_\B } (2 - T_i - T_j )
      \E \braces{ z^+_i z^+_j z^+_p z^+_q d } 
      \pm O\left( d^2 \delta_{\AB, \perp}^2 \right). 
  \end{align*}
  Clear that the summand is nonzero only if $(i, j) = (p, q)$ or $(i, j) = (q, p)$. Hence, 
  \[
    \Term_1([\Q_1]_{p, q})
    = - \tilde{S} ( 1 - \tilde{S} )
      \frac{ [\KAB]_{p, q} + [\KAB]_{q, p} }{ N_\A N_\B d} (2 - T_p - T_q ) 
      \pm O\left( d^2 \delta_{\AB, \perp}^2 \right). 
  \]
  Then, for $\Term_2$, we compute 
  \begin{align*}
    \Term_2([\Q_1]_{p, q})
    &= 
      - K \E \braces{
        \frac{ S_\A(\xA^+, \xB^+) \exp( \fA^+ \cdot \fB^-)}{\exp( \fA^+ \cdot \fB^+)}
        z^-_p z^+_q d
      }  \\
    &= 
      - K \E \braces{
        \left(
          \frac{ \tilde{S} \tilde{E}_{+, -} }{ \tilde{E}_0 }
          \left(
            1 
            - \tilde{S} \tilde\delta_{+, +} 
            - ( 1 - \tilde{S} ) \tilde\delta_{+, T+}
            + \tilde\delta_{+, -} 
          \right)
        \right)
        z^-_p z^+_q d
      }  
      \pm O\left( d^2 \delta_{\AB, \perp}^2 \right) \\
    &= 
      - K \E \braces{
        \left(
          \frac{ \tilde{S} \tilde{E}_{+, -} }{ \tilde{E}_0 }
          \left(
            - \tilde{S} \tilde\delta_{+, +} 
            - ( 1 - \tilde{S} ) \tilde\delta_{+, T+}
            + \tilde\delta_{+, -} 
          \right)
        \right)
        z^-_p z^+_q d
      }  
      \pm O\left( d^2 \delta_{\AB, \perp}^2 \right) \\
    &= - K \sum_{i \ne j} \frac{ [\KAB]_{i, j} }{N_\A N_\B}
      \E \braces{
        \left(
          \frac{ \tilde{S} \tilde{E}_{+, -} }{ \tilde{E}_0 }
          \left(
            - \tilde{S} z^+_i z^+_j 
            - ( 1 - \tilde{S} ) z^+_i z^+_j T_j
            + z^+_i z^-_j 
          \right)
        \right)
        z^-_p z^+_q d
      }  \\ 
      &\qquad
      \pm O\left( d^2 \delta_{\AB, \perp}^2 \right). 
  \end{align*}
  Again, the summand is nonzero only if $(i, j) = (p, q)$ or $(i, j) = (q, p)$. By Lemma~\ref{lemma: stage 2: Zc},
  we have 
  \begin{align*}
    & \Term_2([\Q_1]_{p, q}) \\
    &= - \frac{\tilde{S} K}{\tilde{E}_0}
      \frac{ [\KAB]_{p, q} }{N_\A N_\B}
      \left(
        \left( - \tilde{S} - ( 1 - \tilde{S} ) T_q \right)
        \E \braces{ \exp(I_{+, -}) z^+_p z^-_p }
        + \E \braces{ \exp(I_{+, -}) z^+_p z^+_q z^-_p z^-_q d }  
      \right) \\
      &\qquad
      - \frac{\tilde{S} K}{\tilde{E}_0} \frac{ [\KAB]_{q, p} }{N_\A N_\B}
      \left(
        \left( - \tilde{S} - ( 1 - \tilde{S} ) T_p \right)
        \E \braces{ \exp(I_{+, -}) z^+_p z^-_p } 
        + d\inv \E \braces{ \exp(I_{+, -}) }  
      \right)
      \pm O\left( d^2 \delta_{\AB, \perp}^2 \right) \\
    &= - \frac{\tilde{S} K Z_c}{\tilde{E}_0}
      \frac{ [\KAB]_{p, q} }{N_\A N_\B d}
      \left(
        \left( - \tilde{S} - ( 1 - \tilde{S} ) T_q \right)
        T_p 
        + T_p T_q 
      \right) \\
      &\qquad
      - \frac{\tilde{S}  K Z_c}{\tilde{E}_0} \frac{ [\KAB]_{q, p} }{N_\A N_\B d}
      \left(
        \left( - \tilde{S} - ( 1 - \tilde{S} ) T_p \right)
        T_p 
        + 1
      \right)
      \pm O\left( d^2 \delta_{\AB, \perp}^2 \right) \\
    &= - (1 - \tilde{S}) 
      \frac{ [\KAB]_{p, q} }{N_\A N_\B d}
      \tilde{S} T_p ( T_q - 1 ) 
      - (1 - \tilde{S}) \frac{ [\KAB]_{q, p} }{N_\A N_\B d}
      \left(
        - \tilde{S} T_p - ( 1 - \tilde{S} ) T_p^2
        + 1
      \right)
      \pm O\left( d^2 \delta_{\AB, \perp}^2 \right). 
  \end{align*}
  Similarly, for $\Term_3$, we have 
  \begin{align*}
    & \Term_3([\Q_1]_{p, q}) \\
    &= - K \E \braces{
        \frac{S_\B(\xA^+, \xB^+) \exp( \fA^- \cdot \fB^+)}{\exp( \fA^+ \cdot \fB^+)}
        z^+_p z^-_q d
      } \\
    &= - K \E \braces{
        \frac{ \tilde{S} \tilde{E}_{-, +} }{ \tilde{E}_0 }
        \left(
          1 
          - \tilde{S} \tilde\delta_{+, +} 
          - ( 1 - \tilde{S} ) \tilde\delta_{T+, +}
          + \tilde\delta_{-, +} 
        \right)
        z^+_p z^-_q d
      } 
      \pm O\left( d^2 \delta_{\AB, \perp}^2 \right) \\
    &= - \frac{ \tilde{S} K }{ \tilde{E}_0 }
      \sum_{i \ne j} \frac{ [\KAB]_{i, j} }{ N_\A N_\B }
      \E \braces{
        \tilde{E}_{-, +}
        \left(
          \left( - \tilde{S} - ( 1 - \tilde{S} ) T_i \right) z^+_i z^+_j
          + z^-_i z^+_j 
        \right)
        z^+_p z^-_q d
      } 
      \pm O\left( d^2 \delta_{\AB, \perp}^2 \right) \\
    &= - \frac{ \tilde{S} K }{ \tilde{E}_0 }
      \frac{ [\KAB]_{p, q} }{ N_\A N_\B }
      \left(
        \left( - \tilde{S} - ( 1 - \tilde{S} ) T_p \right)
        Z_c T_q / d
        + Z_c T_p T_q / d
      \right) \\
      &\qquad
      - \frac{ \tilde{S} K }{ \tilde{E}_0 }
      \frac{ [\KAB]_{q, p} }{ N_\A N_\B }
      \left(
        \left( - \tilde{S} - ( 1 - \tilde{S} ) T_q \right)
        Z_c T_q / d
        + Z_c / d
      \right)
      \pm O\left( d^2 \delta_{\AB, \perp}^2 \right) \\
    &= - (1 - \tilde{S})
      \frac{ [\KAB]_{p, q} }{ N_\A N_\B d }
      \tilde{S} T_q ( T_p - 1 )
      - (1 - \tilde{S})
      \frac{ [\KAB]_{q, p} }{ N_\A N_\B d}
      \left(
        1 - \tilde{S} T_q - ( 1 - \tilde{S} ) T_q^2
      \right)
      \pm O\left( d^2 \delta_{\AB, \perp}^2 \right). 
  \end{align*}
  Combine these together and we obtain 
  \begin{align*}
    [\Q_1]_{p, q}
    &= - \tilde{S} ( 1 - \tilde{S} )
      \frac{ [\KAB]_{p, q} + [\KAB]_{q, p} }{ N_\A N_\B d} (2 - T_p - T_q )  \\
      &\qquad
      - (1 - \tilde{S}) 
      \frac{ [\KAB]_{p, q} }{N_\A N_\B d}
      \tilde{S} \left( 2 T_p T_q - T_p - T_q \right) \\
      &\qquad
      - (1 - \tilde{S}) \frac{ [\KAB]_{q, p} }{N_\A N_\B d}
      \left(
        2 
        - \tilde{S} ( T_p + T_q)  
        - ( 1 - \tilde{S} ) ( T_p^2 + T_q^2 )
      \right) 
      \pm O\left( d^2 \delta_{\AB, \perp}^2 \right). 
  \end{align*}
\end{proof}

\begin{proof}[Proof of Lemma~\ref{lemma: stage 2: estimations for Q1xi}]
  We write
  \begin{align*}
    \Q_{1, \xi_\B}
    &:= \E \braces{
        \left( 2 - S_\A(\xA^+, \xB^+) - S_\B(\xA^+, \xB^+) \right)
        \bxi_\B^+(\z^+)\trans d
      } \\
      &\qquad
      - K \E \braces{
        \frac{ S_\A(\xA^+, \xB^+) \exp( \fA^+ \cdot \fB^-)}{\exp( \fA^+ \cdot \fB^+)}
        \bxi_\B^- (\z^+)\trans d
      } 
      - K \E \braces{
        \frac{S_\B(\xA^+, \xB^+) \exp( \fA^- \cdot \fB^+)}{\exp( \fA^+ \cdot \fB^+)}
        \bxi_\B^+ (\z^-)\trans d
      } \\
    &=: \Term_1\left( \Q_{1, \xi_\B} \right)
      + \Term_2\left( \Q_{1, \xi_\B} \right)
      + \Term_3\left( \Q_{1, \xi_\B} \right). 
  \end{align*}
  We will use the fact that if some quantity $X$ does not depend on $\bxi$, then $\E\braces{ X \bxi } = 0$ to 
  simplify these terms. 
  For $\Term_1$, by Lemma~\ref{lemma: stage 2: estimations for S}, we have 
  \begin{align*}
    \Term_1\left( [\Q_{1, \xi_\B}]_{p, q} \right)
    &= - \E \braces{
        \left( S_\A(\xA^+, \xB^+) + S_\B(\xA^+, \xB^+) \right)
        [\bxi_\B^+]_p z^+_q d
      } \\
    &= - \tilde{S} ( 1 - \tilde{S} )
      \E \braces{
        \left( 
          2 \delta_{+, \bxi+} 
          + 2 \delta_{\bxi+, +} 
          - \delta_{\bxi+, T+} 
          - \delta_{T+, \bxi+} 
        \right)
        [\bxi_\B^+]_p z^+_q d
      } \\
      &\qquad
      \pm O\left( d^3 \left( \delta_{\AB, \perp} + \delta_{N/S} \right) \delta_{\xi, \perp} \delta_{N/S} \right) \\
    &= - \tilde{S} ( 1 - \tilde{S} )
      \E \braces{
        \left( 
          2 \delta_{+, \bxi+} 
          - \delta_{T+, \bxi+} 
        \right)
        [\bxi_\B^+]_p z^+_q d
      } 
      \pm O\left( d^3 \left( \delta_{\AB, \perp} + \delta_{N/S} \right) \delta_{\xi, \perp} \delta_{N/S} \right). 
  \end{align*}
  Recall that 
  \[
    \delta_{+, \xi+}
    = \sum_{i, j} \frac{[\KA\trans\KBxi]_{i, j}}{N_\A N_\B} z^+_i [\bxi_\B^+]_j, \quad
    \delta_{T+, \bxi+} 
    = \sum_{i, j} \frac{[\KA\trans\KBxi]_{i, j}}{N_\A N_\B} T_i z^+_i [\bxi_\B^+]_j.
  \]
  Hence, we can further rewrite $\Term_1\left( [\Q_{1, \xi_\B}]_{p, q} \right)$ as 
  \begin{multline*}
    \Term_1\left( [\Q_{1, \xi_\B}]_{p, q} \right)
    = - \tilde{S} ( 1 - \tilde{S} )
      \sum_{i, j} \frac{[\KA\trans\KBxi]_{i, j}}{N_\A N_\B} \left( 2 - T_i \right)
        \E \braces{
          z^+_i [\bxi_\B^+]_j
          [\bxi_\B^+]_p z^+_q d
        } \\
      \pm O\left( d^3 \left( \delta_{\AB, \perp} + \delta_{N/S} \right) \delta_{\xi, \perp} \delta_{N/S} \right). 
  \end{multline*}
  Note that the summand is nonzero only if $i=q$ and $j=p$. Thus, 
  \[
    \Term_1\left( [\Q_{1, \xi_\B}]_{p, q} \right)
    = - \tilde{S} ( 1 - \tilde{S} )
      \left( 2 - T_q \right)
      \frac{ \inprod{ [\KA]_q }{ [\KBxi]_p } }{ N_\A N_\B d}
      \pm O\left( d^3 \left( \delta_{\AB, \perp} + \delta_{N/S} \right) \delta_{\xi, \perp} \delta_{N/S} \right). 
  \]
  Similarly, we also have 
  \begin{align*}
    \Term_2\left( [\Q_{1, \xi_\B}]_{p, q} \right)
    &= - \frac{K \tilde{S}}{\tilde{E}}
      \E \braces{
        \tilde{E}_{+, -} 
        \delta_{+, \bxi-} 
        [\bxi_\B^-]_p z^+_q d
      }
      \pm O\left( d^3 \left( \delta_{\AB, \perp} + \delta_{N/S} \right) \delta_{\xi, \perp} \delta_{N/S} \right), \\
    \Term_3\left( [\Q_{1, \xi_\B}]_{p, q} \right)
    &= - \frac{ K \tilde{S} }{ \tilde{E} }
      \E \braces{
        \tilde{E}_{-, +}
        \left(
          \delta_{-, \bxi+} 
          - \tilde{S} \delta_{+, \bxi+} 
          - ( 1 - \tilde{S} ) \delta_{T+, \bxi+}
        \right)
        [\bxi_\B^-]_p z^+_q d
      } \\
      &\qquad
      \pm O\left( d^3 \left( \delta_{\AB, \perp} + \delta_{N/S} \right) \delta_{\xi, \perp} \delta_{N/S} \right). 
  \end{align*}
  Then, we write 
  \[
    \E \braces{
      \tilde{E}_{+, -} 
      \delta_{+, \bxi-} 
      [\bxi_\B^-]_p z^+_q d
    }
    = \sum_{i, j} \frac{ [\KA\trans\KBxi]_{i, j} }{ N_\A N_\B }
      \E \braces{
        \tilde{E}_{+, -} 
        z^+_i [\bxi_{\B}^-]_j
        [\bxi_\B^-]_p z^+_q d
      } .
  \]
  If $j \ne p$, clear that that summand is $0$. If $i \ne q$, then by flipping the sign of 
  $z^\pm_q$ simultaneously, we flip the sign of $\tilde{E}_{+, -} z^+_i z^+_q$. Therefore, 
  when $i \ne q$, the summand is also $0$. Thus, 
  \begin{align*}
    \Term_2\left( [\Q_{1, \xi_\B}]_{p, q} \right)
    &= - \frac{\tilde{S} K Z_c}{\tilde{E}}
      \frac{ \inprod{ [\KA]_q }{ [\KBxi]_p } }{ N_\A N_\B d}
      \pm O\left( d^3 \left( \delta_{\AB, \perp} + \delta_{N/S} \right) \delta_{\xi, \perp} \delta_{N/S} \right) \\
    &= -  ( 1 - \tilde{S} )
      \frac{ \inprod{ [\KA]_q }{ [\KBxi]_p } }{ N_\A N_\B d}
      \pm O\left( d^3 \left( \delta_{\AB, \perp} + \delta_{N/S} \right) \delta_{\xi, \perp} \delta_{N/S} \right). 
  \end{align*}
  Similarly, for $\Term_3$, we compute  
  \begin{align*}
    & \E \braces{
        \tilde{E}_{-, +}
        \left(
          \delta_{-, \bxi+} 
          - \tilde{S} \delta_{+, \bxi+} 
          - ( 1 - \tilde{S} ) \delta_{T+, \bxi+}
        \right)
        [\bxi_\B^-]_p z^+_q d
      } \\
    =\;& 
      \sum_{i, j} \frac{ [\KA\trans\KBxi]_{i, j} }{ N_\A N_\B }
      \E \braces{
        \tilde{E}_{-, +}
        \left(
          z^-_i [\bxi_\B^-]_j 
          - \tilde{S} z^+_i [\bxi_\B^-]_j 
          - ( 1 - \tilde{S} ) (1 - T_i )z^-_i [\bxi_\B^-]_j 
        \right)
        [\bxi_\B^-]_p z^+_q d
      } \\
    =\;& \frac{ [\KA\trans\KBxi]_{q, p} }{ N_\A N_\B }
      \E \braces{
        \tilde{E}_{-, +}
        \left(
          z^-_q 
          - \tilde{S} z^+_q  
          - ( 1 - \tilde{S} ) (1 - T_i ) z^-_q 
        \right)
        z^+_q 
      } \\
    =\;& Z_c \frac{ [\KA\trans\KBxi]_{q, p} }{ N_\A N_\B d}
      \left(
        \tilde{S} ( T_q - 1 - T_q^2 ) + T_q^2 
      \right) .
  \end{align*}
  Thus, 
  \begin{multline*}
    \Term_3\left( [\Q_{1, \xi_\B}]_{p, q} \right)
    = - (1 - \tilde{S})
       \frac{ \inprod{ [\KA]_q }{ [\KBxi]_p } }{ N_\A N_\B d}
      \left(
        \tilde{S} ( T_q - 1 - T_q^2 ) + T_q^2 
      \right) \\
      \pm O\left( d^3 \left( \delta_{\AB, \perp} + \delta_{N/S} \right) \delta_{\xi, \perp} \delta_{N/S} \right). 
  \end{multline*}
  Combine these together, rearrange terms and we obtain 
  \begin{align*}
    [ \Q_{1, \bxi_\B} ]_{p, q}
    &= -( 1 - \tilde{S} )
      \left(
        1 + \tilde{S} 
        + (1 - \tilde{S}) T_q^2  
      \right)
      \frac{ \inprod{ [\KA]_q }{ [\KBxi]_p } }{ N_\A N_\B d}
      \pm O\left( d^3 \left( \delta_{\AB, \perp} + \delta_{N/S} \right) \delta_{\xi, \perp} \delta_{N/S} \right). 
  \end{align*}
  To obtain the formula for $\Q_{1, \bxi_\A}$, it suffices to interchange the roles of $\A$ and $\B$. 
\end{proof}

\begin{proof}[Proof of Lemma~\ref{lemma: stage 2: estimations for Q2}]
  Recall that
  \begin{align*}
    \Q_2
    &:= \E \braces{
        \left( 2 - S_\A(\xA^+, \xB^+) - S_\B(\xA^+, \xB^+) \right)
        \bxi_\B^+ (\bxi_\A^+)\trans d
      } \\
      &\qquad
      - K \E \braces{
        \frac{ S_\A(\xA^+, \xB^+) \exp( \fA^+ \cdot \fB^-)}{\exp( \fA^+ \cdot \fB^+)}
        \bxi_\B^- (\bxi_\A^+)\trans d 
      }
      - K \E \braces{ 
        \frac{S_\B(\xA^+, \xB^+) \exp( \fA^- \cdot \fB^+)}{\exp( \fA^+ \cdot \fB^+)}
        \bxi_\B^+(\bxi_\A^-)\trans d
      } \\
    &=: \Term_1(\Q_2) + \Term_2(\Q_2) + \Term_3(\Q_2). 
  \end{align*} 
  We now estimate each of these three terms. Again, the strategy is to leverage the symmetry of the
  distribution of $\bxi$ to argue that some part of the expectation is $0$. For $\Term_1$, we have 
  \begin{multline*}
    \Term_1([\Q_2]_{p, q})
    = - \tilde{S} ( 1 - \tilde{S} )
      \E \braces{
        \left( 
          2 \delta_{+, \bxi+} + 2 \delta_{\bxi+, +} 
          - \delta_{\bxi+, T+} - \delta_{T+, \bxi+} 
        \right)
        [\bxi_\B^+]_p [\bxi_\A^+]_q d
      } \\
      \pm O\left( d^3 \left( \delta_{\AB, \perp} + \delta_{N/S} \right) \delta_{\xi, \perp} \delta_{N/S} \right). 
  \end{multline*}
  Note that none of these $\delta$'s depends on both $\bxi_\A^+$ and $\bxi_\B^+$. Therefore, 
  the first term is $0$. Similarly, for $\Term_2$, we have 
  \begin{align*}
    \Term_2([\Q_2]_{p, q})
    &= - \frac{K \tilde{S} }{ \tilde{E} }
      \E \braces{
        \left(
          \tilde{E}_{+, -}
          \left(
            \delta_{+, \bxi-} + \delta_{\bxi+, -} 
            - \tilde{S} \left( \delta_{+, \bxi+} + \delta_{\bxi+, +} \right)
            - ( 1 - \tilde{S} ) \delta_{\bxi+, T+} 
          \right)
        \right)
        [\bxi_\B^-]_p [\bxi_\A^+]_q d 
      } \\
      &\qquad
      \pm O\left( d^3 \left( \delta_{\AB, \perp} + \delta_{N/S} \right) \delta_{\xi, \perp} \delta_{N/S} \right) \\
    &= \pm O\left( d^3 \left( \delta_{\AB, \perp} + \delta_{N/S} \right) \delta_{\xi, \perp} \delta_{N/S} \right). 
  \end{align*}
  The same is also true for $\Term_3$. Thus, 
  \[
    [\Q_2]_{p, q}
    = \pm O\left( d^3 \left( \delta_{\AB, \perp} + \delta_{N/S} \right) \delta_{\xi, \perp} \delta_{N/S} \right). 
  \]
\end{proof}

\begin{proof}[Proof of Lemma~\ref{lemma: stage 2: estimations for Q0}]
  Recall from Corollary~\ref{cor: dynamics, Q} that $Q_0$ is defined as 
  \begin{align*}
    Q_0
    &:= 
      - \E_{\xA^+, \xB^+} \braces{ 
        \left( 2 - S_\A(\xA^+, \xB^+) - S_\B(\xA^+, \xB^+) \right)
        \inprod{\fA^+}{\fB^+} 
      } \\
      &\qquad
      + K 
      \E_{\xA^+, \xB^\pm} \braces{ 
        \frac{ S_\A(\xA^+, \xB^+) \exp(\tau_t^2 \fA^+ \cdot \fB^-)}{\exp(\tau_t^2 \fA^+ \cdot \fB^+)}
        \inprod{\fA^+}{\fB^-} 
      }  \\ 
      &\qquad
      + K 
      \E_{\xA^\pm, \xB^+} \braces{ 
        \frac{S_\B(\xA^+, \xB^+) \exp(\tau_t^2 \fA^- \cdot \fB^+)}{\exp(\tau_t^2 \fA^+ \cdot \fB^+)}
        \inprod{\fA^-}{\fB^+} 
      }. 
  \end{align*}
  We write 
  \begin{align*}
    \inprod{\fA^+}{\fB^-}
    &= \frac{ \inprod{\KA\z^+}{\KB\z^-} }{N_\A N_\B}
      \pm O\left( \kappa_0 d \delta_{N/S} \delta_{\xi, \perp} \right) \\
    &= \sum_{i, j \in [r]} \frac{[\KAB]_{i, j}}{N_\A N_\B} z^+_i z^-_j
      \pm O\left( \kappa_0 d \delta_{N/S} \delta_{\xi, \perp} \right). 
  \end{align*}
  Therefore, we have 
  \begin{align*}
    Q_0 
    &= - \sum_{i, j \in [r]} \frac{[\KAB]_{i, j}}{N_\A N_\B} [\Q_1]_{i, j}
      \pm O\left( \kappa_0 d \delta_{N/S} \delta_{\xi, \perp} \right) \\ 
    &= - \sum_{k=1}^r \frac{\kappa_k^2}{\norm{\bkappa}^2} [\Q_1]_{k, k}
      \pm O\left( 
        d^2 \delta_{\AB, \perp}^2 
        + \delta_- 
        + \kappa_0 d \delta_{N/S} \delta_{\xi, \perp}
      \right). 
  \end{align*}
\end{proof}

\begin{proof}[Proof of Corollary~\ref{cor: stage 2: d norm}]
  Recall from Lemma~\ref{lemma: d norm and d bar} that 
  \begin{align*}
    \frac{\rd}{\rd t} \norm{[\KA]_p}^2
    &= 2 \frac{\inprod{[\KA]_p}{[\KB\Q_1]_p}}{N_\A N_\B d} \sigma_p^2
      + 2 \frac{\inprod{[\KA]_p}{[\KBxi\Q_{1, \xi_\B}]_p}}{N_\A N_\B d}  \sigma_p^2
      + 2 \frac{\norm{[\KA]_p}^2}{N_\A^2 d} Q_0 \sigma_p^2 \\
    &=: \sum_{i=1}^3 \Term_i\left( \frac{\rd}{\rd t} \norm{[\KA]_p}^2 \right). 
  \end{align*}
  By Lemma~\ref{lemma: stage 2: estimations for Q1}, we have 
  \begin{align*}
    \Term_1\left( \frac{\rd}{\rd t} \norm{[\KA]_p}^2 \right)
    &= 2 \sum_{k=1}^r
      \frac{\norm{[\KA]_p} \norm{[\KB]_k}}{N_\A N_\B d} 
      \inprod{\ol{[\KA]_p}}{\ol{[\KB]_k}} [\Q_1]_{k, p} \sigma_p^2 \\ 
    &= \frac{2 \sigma_p^2 \norm{[\KA]_p} \norm{[\KB]_p}}{N_\A N_\B d} 
        \inprod{\ol{[\KA]_p}}{\ol{[\KB]_p}} [\Q_1]_{p, p} \\ 
      &\qquad
      \pm O\left(  
        \frac{\sigma_p^2 \kappa_p^2}{N_\A N_\B d} 
        \kappa_0 d
        \delta_{\AB, \perp}^2 
      \right). 
  \end{align*}
  By Lemma~\ref{lemma: stage 2: estimations for Q1xi}, we have 
  \begin{align*}
    \Term_2\left( \frac{\rd}{\rd t} \norm{[\KA]_p}^2 \right)
    &= 2 \sum_{k=1}^{d-r}
      \frac{\norm{[\KA]_p} \norm{[\KBxi]_k}}{N_\A N_\B d}  
      \inprod{\ol{[\KA]_p}}{\ol{[\KBxi]_k}}
      [\Q_{1, \xi_\B}]_{k, p} \sigma_p^2 \\ 
    &= O\left(
        \frac{\sigma_p^2 \kappa_p^2}{N_\A N_\B d}  
        d \delta_{N/S}^2 \delta_{\xi, \perp}^2
      \right). 
  \end{align*}
  Combine these together and we obtain 
  \begin{align*}
    \frac{\rd}{\rd t} \norm{[\KA]_p}^2
    &= \frac{2 \sigma_p^2 \norm{[\KA]_p} \norm{[\KB]_p}}{N_\A N_\B d} 
    \inprod{\ol{[\KA]_p}}{\ol{[\KB]_p}} [\Q_1]_{p, p} 
    + 2 \frac{\norm{[\KA]_p}^2}{N_\A^2 d} Q_0 \sigma_p^2 \\ 
    &\qquad
    \pm O\left(  
      \frac{\sigma_p^2 \kappa_p^2}{N_\A N_\B d} 
      \kappa_0 d
      \delta_{\AB, \perp}^2 
    \right). 
  \end{align*}
  To get the formula for $\frac{\rd}{\rd t} \norm{[\KB]_p}^2$, it suffices to interchange the roles of 
  $\A$ and $\B$. Similarly, for the noises, we write 
  \begin{align*}
    \frac{\rd}{\rd t} \norm{[\KAxi]_q}^2
    &= \frac{2 \inprod{[\KAxi]_q}{[\KB \Q_{1, \bxi_\A}\trans]_q} }{ N_\A N_\B d}  \sigma_\xi^2
      + \frac{2 \inprod{[\KAxi]_q}{[\KBxi \Q_2]_q} }{N_\A N_\B d} \sigma_\xi^2
      + \frac{2 \norm{[\KAxi]_q}_F^2}{N_\A^2 d} Q_0 \sigma_\xi^2 \\ 
    &= \sum_{i=1}^3 \Term_i\left( \frac{\rd}{\rd t} \norm{[\KAxi]_q}^2 \right). 
  \end{align*}
  By Lemma~\ref{lemma: stage 2: estimations for Q1xi}, we have 
  \begin{align*}
    \Term_1\left( \frac{\rd}{\rd t} \norm{[\KAxi]_q}^2 \right)
    &= 2 \sum_{k=1}^r
      \frac{\norm{[\KAxi]_q} \norm{[\KB]_k} }{ N_\A N_\B d}  
      \inprod{\ol{[\KAxi]_q}}{\ol{[\KB]_k}}
      [\Q_{1, \bxi_\A}]_{q, k} \sigma_\xi^2 \\ 
    &= O\left( 
        \frac{\sigma_\xi^2  \norm{[\KAxi]_q}^2}{ N_\A N_\B d}  
        d \delta_{\xi, \perp}^2
      \right). 
  \end{align*}
  By Lemma~\ref{lemma: stage 2: estimations for Q2}, we have 
  \begin{align*}
    \Term_1\left( \frac{\rd}{\rd t} \norm{[\KAxi]_q}^2 \right)
    &= 2 \sum_{k=1}^{d-r}
      \frac{ \norm{[\KAxi]_q} \norm{[\KBxi]_k } }{N_\A N_\B d} 
      \inprod{\ol{[\KAxi]_q}}{\ol{[\KBxi]_k}}
      [\Q_2]_{k, q} \sigma_\xi^2 \\
    &= O\left(
        \frac{ \sigma_\xi^2  \norm{[\KAxi]_q}^2 }{N_\A N_\B d} 
        d^4 \left( \delta_{\AB, \perp} + \delta_{N/S} \right) \delta_{\xi, \perp}^2 \delta_{N/S} 
      \right). 
  \end{align*}
  Combine these together and we get 
  \[ 
    \frac{\rd}{\rd t} \norm{[\KAxi]_q}^2
    = \frac{2 \norm{[\KAxi]_q}^2}{N_\A^2 d} Q_0 \sigma_\xi^2
      \pm O\left( 
        \frac{\sigma_\xi^2  \norm{[\KAxi]_q}^2}{ N_\A N_\B d}  
        d \delta_{\xi, \perp}^2
      \right). 
  \] 
\end{proof}

\subsection{Maintaining the orthogonality}
\label{sec: stage 2: orthogonality}

In this subsection, we control the growth of $\delta_{\AB, \perp}$ and $\delta_{\xi, \perp}$. First, we 
derive the equations that govern the evolution of the off-diagonal terms. 

\begin{lemma}
  \label{lemma: stage 2: d KABpq}
  In Stage~2, for any $p \ne q \in [r]$, we have 
  \begin{align*}
    \frac{\rd}{\rd t} \inprod{ \ol{[\KA]_p} }{ \ol{[\KB]_q} }
    &= 
      [\Q_1]_{q, p} 
      \frac{
        \kappa_q / \kappa_p \sigma_p^2
        + \kappa_p / \kappa_q \sigma_q^2
      }{N_\A N_\B d} \\
      &\qquad
      - \inprod{ \ol{ [\KA]_p } }{ \ol{[\KB]_q} } 
        \left(
          \frac{[\Q_1]_{p, p} \sigma_p^2}{N_\A N_\B d}
          + \frac{[\Q_1]_{q, q} \sigma_q^2}{N_\A N_\B d}
        \right)  \\
      &\qquad
      + \inprod{\ol{[\KA]_p}}{\ol{[\KA]_q}} \frac{[\Q_1]_{q, q} \sigma_q^2}{N_\A N_\B d}
      + \inprod{\ol{[\KB]_p}}{\ol{[\KB]_q}} \frac{[\Q_1]_{p, p} \sigma_p^2}{N_\A N_\B d}
      \\
      &\qquad
      \pm O\left( 
        \frac{ \sigma_{\max}^2}{N_\A N_\B d}
        d \kappa_0 \delta_{\AB, \perp} \left( \sqrt{\delta_-} + \delta_{\AB, \perp} \right)
      \right). 
  \end{align*}
\end{lemma}

\begin{lemma}
  \label{lemma: stage 2: d KAApq, d KBB pq}
  In Stage~2, for any $p \ne q \in [r]$, we have 
  \begin{align*}
    \frac{\rd}{\rd t} \inprod{\ol{[\KA]_p}}{\ol{[\KA]_q}}
    &=
      [\Q_1]_{q, p} \frac{ \kappa_q / \kappa_p \sigma_p^2 }{N_\A N_\B d }  
      + [\Q_1]_{p, q} \frac{ \kappa_p / \kappa_q \sigma_q^2 }{N_\A N_\B d }  
      \\
      &\qquad
      - \inprod{ \ol{ [\KA]_p } }{ \ol{[\KA]_q} } 
        \left(
          \frac{[\Q_1]_{p, p} \sigma_p^2 }{N_\A N_\B d }  
          + \frac{[\Q_1]_{q, q} \sigma_q^2 }{N_\A N_\B d }
        \right)
      \\
      &\qquad
      + \inprod{\ol{[\KA]_p}}{\ol{[\KB]_q}} \frac{[\Q_1]_{q, q} \sigma_q^2 }{N_\A N_\B d }
      + \inprod{\ol{[\KB]_p}}{\ol{[\KA]_q}} \frac{[\Q_1]_{p, p} \sigma_p^2 }{N_\A N_\B d }  
      \\
      &\qquad
      \pm \left( 
        \frac{\sigma_{\max}^2 }{N_\A N_\B d } 
        d \kappa_0 \delta_{\AB, \perp} \left( \sqrt{ \delta_- } + \delta_{\AB, \perp} \right) 
      \right),
  \end{align*}
  and 
  \begin{align*}
    \frac{\rd}{\rd t} \inprod{\ol{[\KB]_p}}{\ol{[\KB]_q}}
    &=
      [\Q_1]_{p, q} \frac{ \kappa_q / \kappa_p \sigma_p^2 }{N_\A N_\B d }  
      + [\Q_1]_{q, p} \frac{ \kappa_p / \kappa_q \sigma_q^2 }{N_\A N_\B d }  
      \\
      &\qquad
      - \inprod{ \ol{ [\KB]_p } }{ \ol{[\KB]_q} } 
        \left(
          \frac{[\Q_1]_{p, p} \sigma_p^2 }{N_\A N_\B d }  
          + \frac{[\Q_1]_{q, q} \sigma_q^2 }{N_\A N_\B d }
        \right)
      \\
      &\qquad
      + \inprod{\ol{[\KB]_p}}{\ol{[\KA]_q}} \frac{[\Q_1]_{q, q} \sigma_q^2 }{N_\A N_\B d }
      + \inprod{\ol{[\KA]_p}}{\ol{[\KB]_q}} \frac{[\Q_1]_{p, p} \sigma_p^2 }{N_\A N_\B d }  
      \\
      &\qquad
      \pm \left( 
        \frac{\sigma_{\max}^2 }{N_\A N_\B d } 
        d \kappa_0 \delta_{\AB, \perp} \left( \sqrt{ \delta_- } + \delta_{\AB, \perp} \right) 
      \right). 
  \end{align*}
\end{lemma}

Now, we are ready to control the off-diagonalness. The proof is similar to the one of Lemma~\ref{lemma: stage 1: 
orthogonality between signals}. 

\begin{lemma}[Orthogonality between signals]
  \label{lemma: stage 2: orthogonality between signals}
  For any $p \ne q$, define 
  \[
    \hat{\delta}_{\perp, p, q}
    := \inprod{\ol{[\KA]_p}}{\ol{[\KB]_q}}^2
      + \inprod{\ol{[\KB]_p}}{\ol{[\KA]_q}}^2
      + \inprod{\ol{[\KA]_p}}{\ol{[\KA]_q}}^2
      + \inprod{\ol{[\KB]_p}}{\ol{[\KB]_q}}^2. 
  \]
  In Stage~2, we have 
  \[ 
    \frac{\rd}{\rd t} \hat{\delta}_{\perp, p, q}
    \le  
      O\left(
        \frac{
          \sigma_{\max}^2 
        }{N_\A N_\B d}  
        \kappa_0
        \left( d^2 \delta_{\AB, \perp}^2 + \delta_{\AB, \perp} \sqrt{\delta_-} + \delta_- \right)
        \sqrt{\hat{\delta}_{\perp, p, q}}
      \right). 
  \] 
\end{lemma}
Note that the LHS is of order $\delta_{\AB, \perp}^2$ and for the RHS, the only term that can potentially have 
the same order is the $\left(\delta_- \sqrt{\hat{\delta}_{\perp, p, q}}\right)$-relatd term. We will show later 
that $\delta_- = o(\delta_{\AB, \perp})$, whence this is also a higher order term. 

Then, we consider the orthogonality between signals and noises. 

\begin{lemma}[Orthogonality between signals and noises]
  \label{lemma: stage 2: orthogonality between signals and noises}
  For any $p \in [r]$ and $q \in [d - r]$, define 
  \begin{align*}
    \hat{\delta}_{\xi, \perp, p, q}
    &:= \inprod{\ol{[\KA]_p} }{\ol{ [\KAxi]_q }}^2 + \inprod{\ol{[\KB]_p} }{\ol{ [\KAxi]_q }}^2 \\ 
      &\qquad
      + \inprod{\ol{[\KA]_p} }{\ol{ [\KBxi]_q }}^2 + \inprod{\ol{[\KB]_p} }{\ol{ [\KBxi]_q }}^2. 
  \end{align*}
  In Stage~2, we have 
  \[
    \frac{\rd}{\rd t} \hat{\delta}_{\xi, \perp, p, q}
    \le O\left(
      \frac{\sigma_{\max}^2 }{N_\A N_\B d } 
      d \kappa_0 
      \delta_{\xi, \perp}^2 \left( \delta_{\xi, \perp} +  \sqrt{\delta_-} \right)
    \right).
  \]
\end{lemma}

Finally, we deal with the orthogonality between the noises. 

\begin{lemma}[Orthogonality between noises]
  \label{lemma: stage 2: orthogonality between noises}
  In Stage~2, for any $p, q \in [d - r]$, we have 
  \[
    \frac{\rd}{\rd t} \inprod{ \ol{[\KAxi]_p} }{ \ol{[\KBxi]_q} }
    = \pm O\left( \frac{\sigma_\xi^2 }{N_\A N_\B d } d \delta_{\xi, \perp}^2 \right). 
  \]
\end{lemma}

\subsubsection*{Omitted proof of this subsection}

\begin{proof}[Proof of Lemma~\ref{lemma: stage 2: d KABpq}]
  Recall that 
  \begin{align*}
    \frac{\rd}{\rd t} \ol{ [\KA]_p }
    &= \left( \Id - \ol{ [\KA]_p } \left( \ol{ [\KA]_p } \right)\trans  \right)
      \left(
        \frac{[\KB \Q_1]_p }{ \norm{[\KA]_p} }
        + \frac{[\KBxi \Q_{1, \xi_\B}]_p}{\norm{[\KA]_p} }
      \right)
      \frac{\sigma_p^2 }{N_\A N_\B d }, \\
    \frac{\rd}{\rd t} \ol{ [\KB]_q }
    &= \left( \Id - \ol{ [\KB]_q } \left( \ol{ [\KB]_q } \right)\trans  \right)
      \left(
        \frac{[\KA \Q_1\trans]_q }{ \norm{[\KB]_q} }
        + \frac{[\KAxi \Q_{1, \xi_\A}]_q }{ \norm{[\KB]_q} }
      \right)
      \frac{\sigma_q^2 }{N_\A N_\B d }. 
  \end{align*}
  Then, we write 
  \begin{align*}
    \inprod{ \ol{[\KA]_p} }{ \frac{\rd}{\rd t} \ol{[\KB]_q} }
    &= 
      \ol{[\KA]_p}\trans
      \left( \Id - \ol{ [\KB]_q } \left( \ol{ [\KB]_q } \right)\trans  \right)
      \frac{[\KA \Q_1\trans]_q }{ \norm{[\KB]_q} }
      \frac{\sigma_q^2 }{N_\A N_\B d } \\
      &\qquad
      + \ol{[\KA]_p}\trans
      \left( \Id - \ol{ [\KB]_q } \left( \ol{ [\KB]_q } \right)\trans  \right)
      \frac{[\KAxi \Q_{1, \xi_\A}]_q }{ \norm{[\KB]_q} }
      \frac{\sigma_q^2 }{N_\A N_\B d }.  
  \end{align*}
  For the second term, by Lemma~\ref{lemma: stage 2: estimations for Q1xi}, we have 
  \begin{align*}
    & \ol{[\KA]_p}\trans
      \left( \Id - \ol{ [\KB]_q } \left( \ol{ [\KB]_q } \right)\trans  \right)
      \frac{[\KAxi \Q_{1, \xi_\A}]_q }{ \norm{[\KB]_q} } \\
    =\;& 
      \sum_{k=1}^{d-r}
      \left( 
        \inprod{ \ol{[\KA]_p} }{ \ol{[\KAxi]_k} }
        - \inprod{ \ol{[\KA]_p} }{ \ol{ [\KB]_q } }
          \inprod{ \ol{ [\KB]_q } }{ \ol{[\KAxi]_k} }
      \right)
      \frac{\norm{[\KAxi]_k} [\Q_{1, \xi_\A}]_{k, q} }{ \norm{[\KB]_q} } \\
    =\;& \pm O\left( d \delta_{\xi, \perp}^2 \delta_{N/S}^2 \right). 
  \end{align*}
  Hence, 
  \begin{multline*}
    \inprod{ \ol{[\KA]_p} }{ \frac{\rd}{\rd t} \ol{[\KB]_q} }
    = \ol{[\KA]_p}\trans
      \left( \Id - \ol{ [\KB]_q } \left( \ol{ [\KB]_q } \right)\trans  \right)
      \frac{[\KA \Q_1\trans]_q }{ \norm{[\KB]_q} }
      \frac{\sigma_q^2 }{N_\A N_\B d }  \\
      \pm O\left( \frac{\sigma_{\max}^2}{N_\A N_\B d} d \delta_{\xi, \perp}^2 \delta_{N/S}^2 \right). 
  \end{multline*}
  For the first term, we have 
  \begin{multline*}
    \ol{[\KA]_p}\trans
      \left( \Id - \ol{ [\KB]_q } \left( \ol{ [\KB]_q } \right)\trans  \right)
      \frac{[\KA \Q_1\trans]_q }{ \norm{[\KB]_q} } \\
    = \sum_{k=1}^r 
      \left( 
        \inprod{\ol{[\KA]_p}}{\ol{[\KA]_k}}
        - \inprod{ \ol{[\KA]_p} }{ \ol{ [\KB]_q } }
          \inprod{ \ol{ [\KB]_q } }{ \ol{[\KA]_k} }
      \right)
      \frac{ \norm{[\KA]_k} [\Q_1]_{q, k} }{ \norm{[\KB]_q} } . 
  \end{multline*}
  When $k \notin \{p, q\}$, we have $\left| \inprod{\ol{[\KA]_p}}{\ol{[\KA]_k}} \right| \le O(\delta_{\AB, \perp})$, 
  $\left| \inprod{ \ol{ [\KB]_q } }{ \ol{[\KA]_k} } \right| \le O( \delta_{\AB, \perp})$, and
  $\norm{[\KA]_k}/\norm{[\KB]_q} \le \kappa_0$. Meanwhile, by Lemma~\ref{lemma: stage 2: estimations for Q1}, 
  we also have $|[\Q_1]_{q, k}| \le O(\delta_{\AB, \perp} )$. Hence, 
  \begin{multline*}
    \left| 
      \sum_{k \notin \{p, q\}}
      \left( 
        \inprod{\ol{[\KA]_p}}{\ol{[\KA]_k}}
        - \inprod{ \ol{[\KA]_p} }{ \ol{ [\KB]_q } }
          \inprod{ \ol{ [\KB]_q } }{ \ol{[\KA]_k} }
      \right)
      \frac{ \norm{[\KA]_k} [\Q_1]_{q, k} }{ \norm{[\KB]_q} }
    \right| \\
    \le O\left( d \kappa_0 \delta_{\AB, \perp}^2 \right).
  \end{multline*}
  Therefore, 
  \begin{align*}
    & \inprod{ \ol{[\KA]_p} }{ \frac{\rd}{\rd t} \ol{[\KB]_q} } 
      \left( \frac{\sigma_q^2}{N_\A N_\B d}  \right)\inv \\
    =\;& 
      \left( 
        1
        - \inprod{ \ol{[\KA]_p} }{ \ol{ [\KB]_q } }
          \inprod{ \ol{ [\KB]_q } }{ \ol{[\KA]_p} }
      \right)
      \frac{ \norm{[\KA]_p} [\Q_1]_{q, p} }{ \norm{[\KB]_q} }  \\
      &\qquad
      + \left( 
        \inprod{\ol{[\KA]_p}}{\ol{[\KA]_q}}
        - \inprod{ \ol{[\KA]_p} }{ \ol{ [\KB]_q } }
          \inprod{ \ol{ [\KB]_q } }{ \ol{[\KA]_q} }
      \right)
      \frac{ \norm{[\KA]_q} [\Q_1]_{q, q} }{ \norm{[\KB]_q} }  \\
      &\qquad
      \pm O\left( d \kappa_0 \delta_{\AB, \perp}^2 \right)
      \pm O\left( d \delta_{\xi, \perp}^2 \delta_{N/S}^2 \right). 
  \end{align*}
  For the first term, we have 
  \begin{align*}
    & \left( 
        1
        - \inprod{ \ol{[\KA]_p} }{ \ol{ [\KB]_q } }
          \inprod{ \ol{ [\KB]_q } }{ \ol{[\KA]_p} }
      \right)
      \frac{ \norm{[\KA]_p} [\Q_1]_{q, p} }{ \norm{[\KB]_q} } \\
    =\;& 
      \left( 
        1 \pm \delta_{\AB, \perp}^2 
      \right)
      \frac{ \kappa_p [\Q_1]_{q, p} }{ \kappa_q } \left( 1 \pm \sqrt{\delta_-} \right) \\ 
    =\;& 
      \frac{ \kappa_p [\Q_1]_{q, p} }{ \kappa_q } 
      \pm O\left( \kappa_0 \delta_{\AB, \perp} \sqrt{\delta_-}  \right)
      \pm O\left( \kappa_0 \delta_{\AB, \perp}^3 \right). 
  \end{align*}
  For the second term, we have 
  \begin{align*}
    & \left( 
        \inprod{\ol{[\KA]_p}}{\ol{[\KA]_q}}
        - \inprod{ \ol{[\KA]_p} }{ \ol{ [\KB]_q } }
          \inprod{ \ol{ [\KB]_q } }{ \ol{[\KA]_q} }
      \right)
      \frac{ \norm{[\KA]_q} [\Q_1]_{q, q} }{ \norm{[\KB]_q} }  \\
    =\;& 
      \left( 
        \inprod{\ol{[\KA]_p}}{\ol{[\KA]_q}}
        - \inprod{ \ol{[\KA]_p} }{ \ol{ [\KB]_q } }
          \left(1 \pm \delta_- \right)
      \right)
      [\Q_1]_{q, q}
      \left( 1 \pm \sqrt{\delta_-} \right) \\
    =\;& 
      \left( 
        \inprod{\ol{[\KA]_p}}{\ol{[\KA]_q}}
        - \inprod{ \ol{[\KA]_p} }{ \ol{ [\KB]_q } }
      \right)
      [\Q_1]_{q, q}
      \pm O\left( \delta_\perp \sqrt{\delta_-} \right) . 
  \end{align*}
  Thus, 
  \begin{align*}
    & \inprod{ \ol{[\KA]_p} }{ \frac{\rd}{\rd t} \ol{[\KB]_q} }
      \left( \frac{\sigma_q^2}{N_\A N_\B d} \right)\inv  \\
    =\;&  
      \frac{ \kappa_p [\Q_1]_{q, p} }{ \kappa_q } 
      \pm O\left( \kappa_0 \delta_{\AB, \perp} \sqrt{\delta_-}  \right)
      \pm O\left( \kappa_0 \delta_{\AB, \perp}^3 \right) \\
      &\qquad
      + \left( 
        \inprod{\ol{[\KA]_p}}{\ol{[\KA]_q}}
        - \inprod{ \ol{[\KA]_p} }{ \ol{ [\KB]_q } }
      \right)
      [\Q_1]_{q, q}
      \pm O\left( \delta_\perp \sqrt{\delta_-} \right)  \\
      &\qquad
      \pm O\left( d \kappa_0 \delta_{\AB, \perp}^2 \right)
      \pm O\left( d \delta_{\xi, \perp}^2 \delta_{N/S}^2 \right) \\ 
    =\;&
      \frac{ \kappa_p [\Q_1]_{q, p} }{ \kappa_q } 
      + \left( 
        \inprod{\ol{[\KA]_p}}{\ol{[\KA]_q}}
        - \inprod{ \ol{[\KA]_p} }{ \ol{ [\KB]_q } }
      \right)
      [\Q_1]_{q, q}
      \\
      &\qquad
      \pm O\left( 
        d \kappa_0 \delta_{\AB, \perp} \left( \sqrt{\delta_-} + \delta_{\AB, \perp} \right)
      \right). 
  \end{align*}
  Hence, 
  \begin{align*}
    \inprod{ \ol{[\KA]_p} }{ \frac{\rd}{\rd t} \ol{[\KB]_q} } 
    &= 
      \frac{ \kappa_p [\Q_1]_{q, p} }{ \kappa_q } 
      \frac{\sigma_q^2}{N_\A N_\B d}
      + \left( 
        \inprod{\ol{[\KA]_p}}{\ol{[\KA]_q}}
        - \inprod{ \ol{[\KA]_p} }{ \ol{ [\KB]_q } }
      \right)
      \frac{[\Q_1]_{q, q} \sigma_q^2}{N_\A N_\B d}
      \\
      &\qquad
      \pm O\left( 
        \frac{ \sigma_{\max}^2}{N_\A N_\B d}
        d \kappa_0 \delta_{\AB, \perp} \left( \sqrt{\delta_-} + \delta_{\AB, \perp} \right)
      \right). 
  \end{align*}
  Interchange the roles of $\A, \B$, $p, q$, replace $\Q_1$ with $\Q_1\trans$, and we obtain 
  \begin{align*}
    \inprod{ \frac{\rd}{\rd t} \ol{[\KA]_p} }{ \ol{[\KB]_q} }
    &= 
      \frac{ \kappa_q [\Q_1]_{q, p} }{ \kappa_p } 
      \frac{\sigma_p^2}{N_\A N_\B d}
      + \left( 
        \inprod{\ol{[\KB]_p}}{\ol{[\KB]_q}}
        - \inprod{ \ol{ [\KA]_p } }{ \ol{[\KB]_q} }
      \right)
      \frac{[\Q_1]_{p, p} \sigma_p^2}{N_\A N_\B d}
      \\
      &\qquad
      \pm O\left( 
        \frac{ \sigma_{\max}^2}{N_\A N_\B d}
        d \kappa_0 \delta_{\AB, \perp} \left( \sqrt{\delta_-} + \delta_{\AB, \perp} \right)
      \right). 
  \end{align*}
  Combine these together and we complete the proof. 
\end{proof}

\begin{proof}[Proof of Lemma~\ref{lemma: stage 2: d KAApq, d KBB pq}]
  Similar to the proof of the previous lemma, we compute 
  \begin{align*}
    \inprod{\frac{\rd}{\rd t} \ol{[\KA]_p}}{\ol{[\KA]_q}}
    &= \ol{[\KA]_q}\trans 
      \left( \Id - \ol{ [\KA]_p } \left( \ol{ [\KA]_p } \right)\trans  \right)
      \left(
        \frac{[\KB \Q_1]_p }{ \norm{[\KA]_p} }
        + \frac{[\KBxi \Q_{1, \xi_\B}]_p}{\norm{[\KA]_p} }
      \right)
      \frac{\sigma_p^2 }{N_\A N_\B d } \\
    &= \ol{[\KA]_q}\trans 
      \left( \Id - \ol{ [\KA]_p } \left( \ol{ [\KA]_p } \right)\trans  \right)
      \frac{[\KB \Q_1]_p }{ \norm{[\KA]_p} }
      \frac{\sigma_p^2 }{N_\A N_\B d } 
      \pm O\left( \frac{\sigma_{\max}^2}{N_\A N_\B d} d \delta_{\xi, \perp}^2 \delta_{N/S}^2 \right).
  \end{align*}
  Again, we have 
  \begin{align*}
    & \ol{[\KA]_q}\trans 
      \left( \Id - \ol{ [\KA]_p } \left( \ol{ [\KA]_p } \right)\trans  \right)
      \frac{[\KB \Q_1]_p }{ \norm{[\KA]_p} } \\ 
    =\;& 
      \ol{[\KA]_q}\trans  
      \left( \Id - \ol{ [\KA]_p } \left( \ol{ [\KA]_p } \right)\trans  \right)
      \ol{[\KB]_p}
      \frac{\norm{[\KB]_p} [\Q_1]_{p, p} }{ \norm{[\KA]_p} } \\
      &\qquad
      + \ol{[\KA]_q}\trans 
      \left( \Id - \ol{ [\KA]_p } \left( \ol{ [\KA]_p } \right)\trans  \right)
      \ol{[\KB]_q}
      \frac{\norm{[\KB]_q} [\Q_1]_{q, p} }{ \norm{[\KA]_p} } \\
      &\qquad
      + \sum_{k \notin \{p, q\}}
      \ol{[\KA]_q}\trans 
      \left( \Id - \ol{ [\KA]_p } \left( \ol{ [\KA]_p } \right)\trans  \right)
      \ol{[\KB]_k}
      \frac{\norm{[\KB]_k} [\Q_1]_{k, p} }{ \norm{[\KA]_p} } \\
    =\;&
      \frac{ \kappa_q [\Q_1]_{q, p} }{ \kappa_p }    
      + \left( 
        \inprod{\ol{[\KB]_p}}{\ol{[\KA]_q}}
        - \inprod{ \ol{ [\KA]_p } }{ \ol{[\KA]_q} }
      \right)
      [\Q_1]_{p, p} 
      \pm \left( d \kappa_0 \delta_{\AB, \perp} \left( \sqrt{ \delta_- } + \delta_{\AB, \perp} \right) \right). 
  \end{align*}
  Therefore, 
  \begin{align*}
    \inprod{\frac{\rd}{\rd t} \ol{[\KA]_p}}{\ol{[\KA]_q}}
    &= 
      [\Q_1]_{q, p} \frac{ \kappa_q / \kappa_p \sigma_p^2 }{N_\A N_\B d }  
      + \left( 
        \inprod{\ol{[\KB]_p}}{\ol{[\KA]_q}}
        - \inprod{ \ol{ [\KA]_p } }{ \ol{[\KA]_q} }
      \right)
      \frac{[\Q_1]_{p, p} \sigma_p^2 }{N_\A N_\B d } 
      \\
      &\qquad
      \pm \left( 
        \frac{\sigma_{\max}^2 }{N_\A N_\B d } 
        d \kappa_0 \delta_{\AB, \perp} \left( \sqrt{ \delta_- } + \delta_{\AB, \perp} \right) 
      \right). 
  \end{align*}
  Interchange the roles of $p, q$ and we obtain 
  \begin{align*}
    \inprod{\ol{[\KA]_p}}{\frac{\rd}{\rd t} \ol{[\KA]_q}}
    &= 
      [\Q_1]_{p, q} \frac{ \kappa_p / \kappa_q \sigma_q^2 }{N_\A N_\B d }  
      + \left( 
        \inprod{\ol{[\KB]_q}}{\ol{[\KA]_p}}
        - \inprod{ \ol{ [\KA]_p } }{ \ol{[\KA]_q} }
      \right)
      \frac{[\Q_1]_{q, q} \sigma_q^2 }{N_\A N_\B d } 
      \\
      &\qquad
      \pm \left( 
        \frac{\sigma_{\max}^2 }{N_\A N_\B d } 
        d \kappa_0 \delta_{\AB, \perp} \left( \sqrt{ \delta_- } + \delta_{\AB, \perp} \right) 
      \right). 
  \end{align*}
  Combine these together and we get 
  \begin{align*}
    \frac{\rd}{\rd t} \inprod{\ol{[\KA]_p}}{\ol{[\KA]_q}}
    &=
      [\Q_1]_{q, p} \frac{ \kappa_q / \kappa_p \sigma_p^2 }{N_\A N_\B d }  
      + [\Q_1]_{p, q} \frac{ \kappa_p / \kappa_q \sigma_q^2 }{N_\A N_\B d }  
      \\
      &\qquad
      - \inprod{ \ol{ [\KA]_p } }{ \ol{[\KA]_q} } 
        \left(
          \frac{[\Q_1]_{p, p} \sigma_p^2 }{N_\A N_\B d }  
          + \frac{[\Q_1]_{q, q} \sigma_q^2 }{N_\A N_\B d }
        \right)
      \\
      &\qquad
      + \inprod{\ol{[\KA]_p}}{\ol{[\KB]_q}} \frac{[\Q_1]_{q, q} \sigma_q^2 }{N_\A N_\B d }
      + \inprod{\ol{[\KB]_p}}{\ol{[\KA]_q}} \frac{[\Q_1]_{p, p} \sigma_p^2 }{N_\A N_\B d }  
      \\
      &\qquad
      \pm \left( 
        \frac{\sigma_{\max}^2 }{N_\A N_\B d } 
        d \kappa_0 \delta_{\AB, \perp} \left( \sqrt{ \delta_- } + \delta_{\AB, \perp} \right) 
      \right). 
  \end{align*}
  Interchange the roles of $\A, \B$, replace $\Q_1$ with $\Q_1\trans$, and we obtain the formula for 
  $\frac{\rd}{\rd t} \inprod{\ol{[\KB]_p}}{\ol{[\KB]_q}}$. 
\end{proof}

\begin{proof}[Proof of Lemma~\ref{lemma: stage 2: orthogonality between signals}]

  First, we consider the $[\Q_1]_{p, q}$-related terms, by Lemma~\ref{lemma: stage 2: estimations for Q1}, we 
  have 
  \begin{align*}
    [\Q_1]_{p, q}
    &= - \tilde{S} ( 1 - \tilde{S} )
      \frac{ [\KAB]_{p, q} + [\KBA]_{q, p} }{ N_\A N_\B d} (2 - T_p - T_q )  \\
      &\qquad
      - (1 - \tilde{S}) 
      \frac{ [\KAB]_{p, q} }{N_\A N_\B d}
      \tilde{S} \left( 2 T_p T_q - T_p - T_q \right) \\
      &\qquad
      - (1 - \tilde{S}) \frac{ [\KAB]_{q, p} }{N_\A N_\B d}
      \left(
        2 
        - \tilde{S} ( T_p + T_q)  
        - ( 1 - \tilde{S} ) ( T_p^2 + T_q^2 )
      \right) 
      \pm O\left( d^2 \delta_{\AB, \perp}^2 \right) \\ 
    &= O\left( \frac{\kappa_p \kappa_q \delta_{\AB, \perp}}{N_A N_\B d} \right). 
  \end{align*}
  Hence, 
  \[
    [\Q_1]_{p, q} \frac{ \kappa_q / \kappa_p \sigma_p^2 }{ N_\A N_\B d }
    = O\left( 
        \frac{\kappa_q^2 }{N_A N_\B d} 
        \frac{ \sigma_p^2 }{ N_\A N_\B d }
        \delta_{\AB, \perp}
      \right)
    = O\left( 
      \frac{1}{\sqrt{d}}
      \frac{ \sigma_{\max}^2 }{ N_\A N_\B d }
      \delta_{\AB, \perp}
    \right). 
  \]
  The same bound also hold for other $[\Q_1]_{p, q}$-related terms. The important thing here is that 
  we have and additional $1 / \sqrt{d}$ factor. 

  Now, we are ready to prove the result. 
  For notational simplicity, define $Z_1 = \inprod{\ol{[\KA]_p}}{\ol{[\KB]_q}}$, 
  $Z_2 = \inprod{\ol{[\KB]_p}}{\ol{[\KA]_q}}$, $Z_3 = \inprod{\ol{[\KA]_p}}{\ol{[\KA]_q}}$, and 
  $Z_4 = \inprod{\ol{[\KB]_p}}{\ol{[\KB]_q}}$. Also define $G_p := [\Q_1]_{p, p} \sigma_p^2 / (N_\A N_\B d)$
  and similarly for $G_q$. Then, we can write the results of Lemma~\ref{lemma: stage 2: d KABpq} and 
  Lemma~\ref{lemma: stage 2: d KAApq, d KBB pq} as 
  \begin{align*}
    \frac{\rd}{\rd t} \begin{bmatrix} Z_1 \\ Z_2 \\ Z_3 \\ Z_4 \end{bmatrix}
    &= 
      \begin{bmatrix}
        - G_p - G_q & 0 & G_q & G_p \\
        0 & - G_p - G_q & G_p & G_q \\ 
        G_q & G_p & - G_p - G_q & 0 \\
        G_p & G_q & 0 & - G_p - G_q \\
      \end{bmatrix}
      \begin{bmatrix} Z_1 \\ Z_2 \\ Z_3 \\ Z_4 \end{bmatrix} \\
      &\qquad
      \pm O\left( 
        \frac{1}{\sqrt{d}}
        \frac{ \sigma_{\max}^2 }{ N_\A N_\B d }
        \delta_{\AB, \perp}
      \right)
      \pm O\left( 
        \frac{ \sigma_{\max}^2}{N_\A N_\B d}
        d \kappa_0 \delta_{\AB, \perp} \left( \sqrt{\delta_-} + \delta_{\AB, \perp} \right)
      \right). 
  \end{align*}
  The eigenvalues of the first matrix are $-2G_p, -2G_q, -2G_p - 2G_q$ and $0$. For the first three eigenvalues, 
  note that 
  \[
    G_p 
    = \frac{ [\Q_1]_{p, p} \sigma_p^2 }{ N_\A N_\B d } 
    \ge \frac{1}{\sqrt{d}} \frac{\sigma_{\max}^2}{N_\A N_\B d}.
  \]
  Hence, the signals will dominate the noises, in particular, the first term on the second line, and push 
  $\norm{\Z}$ toward $0$. Now we consider the eigen-pair $(0, (1, 1, 1, 1))$, for which we will use the actual 
  form of $[\Q_1]_{p, q}$ we obtained in Lemma~\ref{lemma: stage 2: estimations for Q1}. We have 
  \begin{multline*}
    \frac{\rd}{\rd t} \sum_{i=1}^4 Z_i
    = 
      2 \left( [\Q_1]_{p, q} + [\Q_1]_{q, p}  \right)
      \frac{
        \kappa_q / \kappa_p \sigma_p^2
        + \kappa_p / \kappa_q \sigma_q^2
      }{N_\A N_\B d}  \\
      \pm O\left( 
        \frac{ \sigma_{\max}^2}{N_\A N_\B d}
        d \kappa_0 \delta_{\AB, \perp} \left( \sqrt{\delta_-} + \delta_{\AB, \perp} \right)
      \right). 
  \end{multline*}
  By Lemma~\ref{lemma: stage 2: estimations for Q1}, we have 
  \begin{multline*}
    [\Q_1]_{p, q} + [\Q_1]_{q, p}  \\
    = - ( 1 - \tilde{S} )
      \frac{ [\KAB]_{p, q} + [\KBA]_{p, q} }{ N_\A N_\B d}
      \left( 2 - T_p^2 - T_q^2 + \tilde{S} (2 - T_p - T_q)^2 \right)
      \pm O\left( d^2 \delta_{\AB, \perp}^2 \right).
  \end{multline*}
  Then, we write 
  \begin{align*}
    [\KAB]_{p, q} + [\KBA]_{p, q} 
    &= \norm{[\KA]_p} \norm{[\KB]_q} Z_1 + \norm{[\KB]_p} \norm{[\KA]_q} Z_2 \\
    &= (Z_1 + Z_2) \kappa_p \kappa_q (1 \pm \sqrt{\delta_-}). 
  \end{align*}
  Hence, 
  \begin{align*}
    [\Q_1]_{p, q} + [\Q_1]_{q, p}  
    &= - ( 1 - \tilde{S} )
      \frac{\kappa_p \kappa_q}{ N_\A N_\B d} ( Z_1 + Z_2 )
      \left( 2 - T_p^2 - T_q^2 + \tilde{S} (2 - T_p - T_q)^2 \right) \\ 
      &\qquad
      \pm O\left( d^2 \delta_{\AB, \perp}^2 + \delta_{\AB, \perp} \sqrt{\delta_-} \right).
  \end{align*}
  To convert $Z_1 + Z_2$ to $\sum_{i=1}^4 Z_i$. Note that we have 
  \begin{align*}
    Z_1 + Z_2 - Z_3 - Z_4 
    = \inprod{\ol{[\KA]_p} - \ol{[\KB]_p} }{\ol{[\KA]_q} - \ol{[\KB]_q}}
    \le \delta_-. 
  \end{align*}
  Therefore, 
  \begin{align*}
    [\Q_1]_{p, q} + [\Q_1]_{q, p}  
    &= - \frac{ 1 - \tilde{S} }{2}
      \frac{\kappa_p \kappa_q}{ N_\A N_\B d} 
      \left( 2 - T_p^2 - T_q^2 + \tilde{S} (2 - T_p - T_q)^2 \right) 
      \sum_{i=1}^4 Z_i \\ 
      &\qquad
      \pm O\left( d^2 \delta_{\AB, \perp}^2 + \delta_{\AB, \perp} \sqrt{\delta_-} + \delta_- \right).
  \end{align*}
  Thus, 
  \begin{align*}
    \frac{\rd}{\rd t} \sum_{i=1}^4 Z_i
    &= 
      - ( 1 - \tilde{S}) 
      \frac{
        \kappa_q / \kappa_p \sigma_p^2
        + \kappa_p / \kappa_q \sigma_q^2
      }{N_\A N_\B d}  
      \frac{\kappa_p \kappa_q}{ N_\A N_\B d} 
      \left( 2 - T_p^2 - T_q^2 + \tilde{S} (2 - T_p - T_q)^2 \right) 
      \sum_{i=1}^4 Z_i
      \\
      &\qquad
      \pm O\left(
        \frac{
          \sigma_{\max}^2 
        }{N_\A N_\B d}  
        \kappa_0
        \left( d^2 \delta_{\AB, \perp}^2 + \delta_{\AB, \perp} \sqrt{\delta_-} + \delta_- \right)
      \right). 
  \end{align*}
  Note that the coefficient of the first term is negative. Combine this with the previous bound for $\norm{\Z}$,
  and we complete the proof. 
\end{proof}

\begin{proof}[Proof of Lemma~\ref{lemma: stage 2: orthogonality between signals and noises}]
  Recall that 
  \begin{align*}
    \frac{\rd}{\rd t} \ol{ [\KA]_p }
    &= \left( \Id - \ol{ [\KA]_p } \left( \ol{ [\KA]_p } \right)\trans  \right)
      \left(
        \frac{[\KB \Q_1]_p }{ \norm{[\KA]_p} }
        + \frac{[\KBxi \Q_{1, \xi_\B}]_p}{\norm{[\KA]_p} }
      \right)
      \frac{\sigma_p^2 }{N_\A N_\B d }, \\
    \frac{\rd}{\rd t} \ol{ [\KAxi]_q }
    &= \left( \Id - \ol{ [\KAxi]_q } \left( \ol{ [\KAxi]_q } \right)\trans \right)
      \left(
        \frac{ [\KB \Q_{1, \bxi_\A}\trans]_q }{ \norm{[\KAxi]_q} }
        + \frac{ [\KBxi \Q_2]_q }{ \norm{[\KAxi]_q} }
      \right) 
      \frac{\sigma_\xi^2 }{N_\A N_\B d }. 
  \end{align*}
  We now compute $\inprod{\frac{\rd}{\rd t} \ol{ [\KA]_p }}{\ol{ [\KAxi]_q }}$ and
  $\inprod{\ol{ [\KA]_p }}{\frac{\rd}{\rd t} \ol{ [\KAxi]_q }}$ separately. First, we write 
  \begin{align*}
    \inprod{\frac{\rd}{\rd t} \ol{[\KA]_p} }{\ol{ [\KAxi]_q }}
    &= \ol{ [\KAxi]_q }\trans
      \left( \Id - \ol{ [\KA]_p } \left( \ol{ [\KA]_p } \right)\trans  \right)
      \frac{[\KB \Q_1]_p }{ \norm{[\KA]_p} }
      \frac{\sigma_p^2 }{N_\A N_\B d } \\
      &\qquad
      + \ol{ [\KAxi]_q }\trans
      \left( \Id - \ol{ [\KA]_p } \left( \ol{ [\KA]_p } \right)\trans  \right)
      \frac{[\KBxi \Q_{1, \xi_\B}]_p}{\norm{[\KA]_p} }
      \frac{\sigma_p^2 }{N_\A N_\B d }  \\
    &=: \Term_1\left( \inprod{\frac{\rd}{\rd t} \ol{[\KA]_p} }{\ol{ [\KAxi]_q }} \right)
      + \Term_2\left( \inprod{\frac{\rd}{\rd t} \ol{[\KA]_p} }{\ol{ [\KAxi]_q }} \right). 
  \end{align*} 
  Then, we compute
  \begin{align*}
    & \Term_1\left( \inprod{\frac{\rd}{\rd t} \ol{[\KA]_p} }{\ol{ [\KAxi]_q }} \right) \\
    =\;&
      \left( 
        \inprod{ \ol{ [\KAxi]_q } }{\ol{[\KB]_p}}
        - \inprod{ \ol{ [\KAxi]_q } }{ \ol{ [\KA]_p } }
          \inprod{ \ol{ [\KA]_p } }{ \ol{[\KB]_p} }
      \right)
      \frac{ \norm{[\KB]_p} [\Q_1]_{p, p} }{ \norm{[\KA]_p} }
      \frac{\sigma_p^2 }{N_\A N_\B d } \\
      &\qquad
      + \sum_{k \ne p} 
      \left( 
        \inprod{\ol{ [\KAxi]_q }}{\ol{[\KB]_k}}
        - \inprod{ \ol{ [\KAxi]_q } }{ \ol{ [\KA]_p } }
          \inprod{ \ol{ [\KA]_p } }{ \ol{[\KB]_k} }
      \right)
      \frac{ \norm{[\KB]_k} [\Q_1]_{k, p} }{ \norm{[\KA]_p} }
      \frac{\sigma_p^2 }{N_\A N_\B d } \\
    =\;& \left( 
        \inprod{ \ol{ [\KAxi]_q } }{\ol{[\KB]_p}}
        - \inprod{ \ol{ [\KAxi]_q } }{ \ol{ [\KA]_p } }
      \right)
      \frac{[\Q_1]_{p, p} \sigma_p^2 }{N_\A N_\B d } 
      \pm O\left(
        \frac{\sigma_{\max}^2 }{N_\A N_\B d } 
        d \kappa_0 
        \delta_{\xi, \perp} \left( \delta_{\xi, \perp} +  \sqrt{\delta_-} \right)
      \right), 
  \end{align*}
  and, by Lemma~\ref{lemma: stage 2: estimations for Q1xi}, we have  
  \begin{align*}
    \Term_2\left( \inprod{\frac{\rd}{\rd t} \ol{[\KA]_p} }{\ol{ [\KAxi]_q }} \right)
    &= \sum_{k=1}^{d-r}
      \ol{ [\KAxi]_q }\trans
      \left( \Id - \ol{ [\KA]_p } \left( \ol{ [\KA]_p } \right)\trans  \right)
      \frac{[\KBxi]_k [\Q_{1, \xi_\B}]_{k, p}}{\norm{[\KA]_p} }
      \frac{\sigma_p^2 }{N_\A N_\B d } \\
    &= \pm O\left(
        \frac{\sigma_{\max}^2 }{N_\A N_\B d } 
        d \delta_{N/S}^2 \delta_{\xi, \perp}^2
      \right). 
  \end{align*}
  Combine these together and we get 
  \begin{align*}
    \inprod{\frac{\rd}{\rd t} \ol{[\KA]_p} }{\ol{ [\KAxi]_q }}
    &= 
      \left( 
        \inprod{ \ol{ [\KAxi]_q } }{\ol{[\KB]_p}}
        - \inprod{ \ol{ [\KAxi]_q } }{ \ol{ [\KA]_p } }
      \right)
      \frac{[\Q_1]_{p, p} \sigma_p^2 }{N_\A N_\B d } \\
      &\qquad
      \pm O\left(
        \frac{\sigma_{\max}^2 }{N_\A N_\B d } 
        d \kappa_0 
        \delta_{\xi, \perp} \left( \delta_{\xi, \perp} +  \sqrt{\delta_-} \right)
      \right).
  \end{align*}
  Then, we compute $\inprod{\ol{[\KA]_p} }{\frac{\rd}{\rd t} \ol{ [\KAxi]_q }}$. We write
  \begin{align*}
    \inprod{\ol{[\KA]_p} }{\frac{\rd}{\rd t} \ol{ [\KAxi]_q }}
    &= \sum_{k=1}^r
      \ol{[\KA]_p}\trans
      \left( \Id - \ol{ [\KAxi]_q } \left( \ol{ [\KAxi]_q } \right)\trans \right)
      \frac{ [\KB]_k [\Q_{1, \bxi_\A}]_{q, k} }{ \norm{[\KAxi]_q} }
      \frac{\sigma_\xi^2 }{N_\A N_\B d } \\
      &\qquad
      + \sum_{k=1}^{d-r}
      \ol{[\KA]_p}\trans
      \left( \Id - \ol{ [\KAxi]_q } \left( \ol{ [\KAxi]_q } \right)\trans \right)
      \frac{ [\KBxi]_k [\Q_2]_{k, q} }{ \norm{[\KAxi]_q} }
      \frac{\sigma_\xi^2 }{N_\A N_\B d } \\
    &=: \Term_1\left( \inprod{\ol{[\KA]_p} }{\frac{\rd}{\rd t} \ol{ [\KAxi]_q }} \right)
      + \Term_2\left( \inprod{\ol{[\KA]_p} }{\frac{\rd}{\rd t} \ol{ [\KAxi]_q }} \right).
  \end{align*}
  For $\Term_1$, when $k = p$, by Lemma~\ref{lemma: stage 2: estimations for Q1xi}, we have 
  \begin{align*}
    & \ol{[\KA]_p}\trans
      \left( \Id - \ol{ [\KAxi]_q } \left( \ol{ [\KAxi]_q } \right)\trans \right)
      \frac{ [\KB]_k [\Q_{1, \bxi_\A}]_{q, k} }{ \norm{[\KAxi]_q} } \\
    =\;& 
      \left( 
        \inprod{ \ol{[\KA]_p} }{ \ol{[\KB]_p} }
        - \inprod{ \ol{[\KA]_p} }{ \ol{ [\KAxi]_q } }
          \inprod{ \ol{ [\KAxi]_q } }{ \ol{[\KB]_p} }
      \right)
      \frac{ \norm{[\KB]_p} }{ \norm{[\KAxi]_q} } 
      [\Q_{1, \bxi_\A}]_{q, p} \\
    =\;&
      -( 1 - \tilde{S} )
      \left(
        1 + \tilde{S} 
        + (1 - \tilde{S}) T_p^2  
      \right)
      \left( 
        1 \pm O\left( \delta_- + \delta_{\xi, \perp}^2 \right)
      \right)
      \frac{ \norm{[\KB]_p}^2  }{ N_\A N_\B d}
      \inprod{ \ol{[\KB]_p} }{ \ol{[\KAxi]_q} }
      \\
      &\qquad
      \pm O\left( d^3 \left( \delta_{\AB, \perp} + \delta_{N/S} \right) \delta_{\xi, \perp} \delta_{N/S} \right) \\
      =\;&
        -\Theta\left( \frac{ \norm{[\KB]_p}^2  }{ N_\A N_\B d} \right)
          \inprod{ \ol{[\KB]_p} }{ \ol{[\KAxi]_q} }
        \pm O\left( d^3 \left( \delta_{\AB, \perp} + \delta_{N/S} \right) \delta_{\xi, \perp} \delta_{N/S} \right). 
  \end{align*}
  When $k \ne p$, we have 
  \[
    \ol{[\KA]_p}\trans
      \left( \Id - \ol{ [\KAxi]_q } \left( \ol{ [\KAxi]_q } \right)\trans \right)
      \frac{ [\KB]_k [\Q_{1, \bxi_\A}]_{q, k} }{ \norm{[\KAxi]_q} } 
    = O\left( \kappa_0 \delta_{\AB, \perp} \delta_{\xi, \perp} \right). 
  \]
  Hence, 
  \begin{align*}
    \Term_1\left( \inprod{\ol{[\KA]_p} }{\frac{\rd}{\rd t} \ol{ [\KAxi]_q }} \right)
    &= -\Theta\left( \frac{ \norm{[\KB]_p}^2  }{ N_\A N_\B d} \right)
      \frac{\sigma_\xi^2 }{N_\A N_\B d } 
      \inprod{ \ol{[\KB]_p} }{ \ol{[\KAxi]_q} }  \\
      &\qquad
      \pm O\left( 
        \frac{\sigma_{\max}^2 }{N_\A N_\B d } 
        d^3 \kappa_0 \left( \delta_{\AB, \perp} + \delta_{N/S} \right) \delta_{\xi, \perp} \delta_{N/S} 
      \right).
  \end{align*}
  Now we consider $\Term_2$. By Lemma~\ref{lemma: stage 2: estimations for Q2}, we have 
  \[
    \Term_2\left( \inprod{\ol{[\KA]_p} }{\frac{\rd}{\rd t} \ol{ [\KAxi]_q }} \right)
    = 
      \pm O\left( 
        \frac{\sigma_\xi^2 }{N_\A N_\B d }
        d^4 \delta_{\xi, \perp}^2 \delta_{N/S}
        \left( \delta_{\AB, \perp} + \delta_{N/S} \right) 
      \right). 
  \]
  Therefore, 
  \begin{align*}
    \inprod{\ol{[\KA]_p} }{\frac{\rd}{\rd t} \ol{ [\KAxi]_q }} 
    &= -\Theta\left( \frac{ \norm{[\KB]_p}^2  }{ N_\A N_\B d} \right)
      \frac{\sigma_\xi^2 }{N_\A N_\B d } 
      \inprod{ \ol{[\KB]_p} }{ \ol{[\KAxi]_q} }  \\
      &\qquad
      \pm O\left( 
        \frac{\sigma_{\max}^2 }{N_\A N_\B d } 
        d^3 \kappa_0 \left( \delta_{\AB, \perp} + \delta_{N/S} \right) \delta_{\xi, \perp} \delta_{N/S} 
      \right). 
  \end{align*}
  Thus, 
  \begin{align*}
    \frac{\rd}{\rd t}  \inprod{\ol{[\KA]_p} }{\ol{ [\KAxi]_q }} 
    &=
      \left( 
        \inprod{ \ol{ [\KAxi]_q } }{\ol{[\KB]_p}}
        - \inprod{ \ol{ [\KAxi]_q } }{ \ol{ [\KA]_p } }
      \right)
      \frac{[\Q_1]_{p, p} \sigma_p^2 }{N_\A N_\B d } \\
      &\qquad
      - \Theta\left( \frac{ \norm{[\KB]_p}^2  }{ N_\A N_\B d} \right)
        \frac{\sigma_\xi^2 }{N_\A N_\B d } 
        \inprod{ \ol{[\KB]_p} }{ \ol{[\KAxi]_q} } \\
      &\qquad
      \pm O\left(
        \frac{\sigma_{\max}^2 }{N_\A N_\B d } 
        d \kappa_0 
        \delta_{\xi, \perp} \left( \delta_{\xi, \perp} +  \sqrt{\delta_-} \right)
      \right). 
  \end{align*}
  Similarly, one can show that 
  \begin{align*}
    \frac{\rd}{\rd t}  \inprod{\ol{[\KB]_p} }{\ol{ [\KAxi]_q }} 
    &=
      \left( 
        \inprod{ \ol{ [\KAxi]_q } }{\ol{[\KA]_p}}
        - \inprod{ \ol{ [\KAxi]_q } }{ \ol{ [\KB]_p } }
      \right)
      \frac{[\Q_1]_{p, p} \sigma_p^2 }{N_\A N_\B d } \\
      &\qquad
      - \Theta\left( \frac{ \norm{[\KA]_p}^2  }{ N_\A N_\B d} \right)
        \frac{\sigma_\xi^2 }{N_\A N_\B d } 
        \inprod{ \ol{[\KA]_p} }{ \ol{[\KAxi]_q} } \\
      &\qquad
      \pm O\left(
        \frac{\sigma_{\max}^2 }{N_\A N_\B d } 
        d \kappa_0 
        \delta_{\xi, \perp} \left( \delta_{\xi, \perp} +  \sqrt{\delta_-} \right)
      \right). 
  \end{align*}
  For notational simplicity, define 
  \begin{align*}
    X = \inprod{\ol{[\KA]_p} }{\ol{ [\KAxi]_q }}, \quad 
    Y = \inprod{\ol{[\KB]_p} }{\ol{ [\KAxi]_q }}, \quad 
    C_1 = \frac{[\Q_1]_{p, p} \sigma_p^2 }{N_\A N_\B d}, \quad
    C_2 = \frac{\kappa_p^2}{N_\A N_\B d} \frac{\sigma_\xi^2}{N_\A N_\B d}.
  \end{align*}
  Then, we can write 
  \[
    \frac{\rd}{\rd t} \begin{bmatrix} X \\ Y \end{bmatrix}
    = 
      \begin{bmatrix}
        - C_1                 &  C_1 - \Theta(C_2) \\
        C_1 - \Theta(C_2) &  - C_1 
      \end{bmatrix}
      \begin{bmatrix} X \\ Y \end{bmatrix}
      \pm O\left(
        \frac{\sigma_{\max}^2 }{N_\A N_\B d } 
        d \kappa_0 
        \delta_{\xi, \perp} \left( \delta_{\xi, \perp} +  \sqrt{\delta_-} \right)
      \right). 
  \]
  The first matrix is negative semi-definite, whence
  \[
    \frac{\rd}{\rd t} \left( X^2 + Y^2 \right)
    \le O\left(
      \frac{\sigma_{\max}^2 }{N_\A N_\B d } 
      d \kappa_0 
      \delta_{\xi, \perp}^2 \left( \delta_{\xi, \perp} +  \sqrt{\delta_-} \right)
    \right).
  \]
  Since the roles of $\KAxi$ and $\KBxi$ are interchangeable, the same bound also holds for 
  $\inprod{\ol{[\KA]_p} }{\ol{ [\KBxi]_q }}$ and $\inprod{\ol{[\KA]_p} }{\ol{ [\KBxi]_q }}$. 
\end{proof}

\begin{proof}[Proof of Lemma~\ref{lemma: stage 2: orthogonality between noises}]
  We write 
  \begin{align*}
    \inprod{ \ol{[\KAxi]_p} }{ \frac{\rd}{\rd t} \ol{[\KBxi]_q} }
    &= \ol{[\KAxi]_p}\trans 
      \left( \Id - \ol{ [\KBxi]_q } \left( \ol{ [\KBxi]_q } \right)\trans \right)
      \left(
        \frac{ [\KA \Q_{1, \bxi_\B}\trans]_q }{ \norm{[\KBxi]_q} }
        + \frac{ [\KAxi \Q_2\trans]_q }{ \norm{[\KBxi]_q} }
      \right) 
      \frac{\sigma_\xi^2 }{N_\A N_\B d } \\
    &= \ol{[\KAxi]_p}\trans 
      \left( \Id - \ol{ [\KBxi]_q } \left( \ol{ [\KBxi]_q } \right)\trans \right)
      \frac{ [\KA \Q_{1, \bxi_\B}\trans]_q }{ \norm{[\KBxi]_q} }
      \frac{\sigma_\xi^2 }{N_\A N_\B d } \\
      &\qquad
      + \ol{[\KAxi]_p}\trans 
      \left( \Id - \ol{ [\KBxi]_q } \left( \ol{ [\KBxi]_q } \right)\trans \right)
      \frac{ [\KAxi \Q_2\trans]_q }{ \norm{[\KBxi]_q} }
      \frac{\sigma_\xi^2 }{N_\A N_\B d } \\
    &=: \Term_1\left( \inprod{ \ol{[\KAxi]_p} }{ \frac{\rd}{\rd t} \ol{[\KBxi]_q} } \right)
      + \Term_2\left( \inprod{ \ol{[\KAxi]_p} }{ \frac{\rd}{\rd t} \ol{[\KBxi]_q} } \right). 
  \end{align*}
  Then, we compute 
  \begin{align*}
    & \Term_1\left( \inprod{ \ol{[\KAxi]_p} }{ \frac{\rd}{\rd t} \ol{[\KBxi]_q} } \right) \\
    =\;& \sum_{k=1}^r 
      \left( 
        \inprod{\ol{[\KAxi]_p}}{\ol{[\KA]_k}}
        - \inprod{ \ol{[\KAxi]_p} }{ \ol{ [\KBxi]_q } }
          \inprod{ \ol{ [\KBxi]_q } }{ \ol{[\KA]_k} }
      \right)
      \frac{ \norm{[\KA]_k} [\Q_{1, \bxi_\B}]_{q, k} }{ \norm{[\KBxi]_q} }
      \frac{\sigma_\xi^2 }{N_\A N_\B d }.
  \end{align*}
  Note that, the first part of each summand is bounded by $O(\delta_{\xi, \perp})$. For the second part, 
  by Lemma~\ref{lemma: stage 2: estimations for Q1xi}
  we also have
  \begin{align*}
    \frac{ \norm{[\KA]_k} [\Q_{1, \bxi_\B}]_{q, k} }{ \norm{[\KBxi]_q} }
    &= \pm \frac{ \norm{[\KA]_k} }{ \norm{[\KBxi]_q} }
      \left(
        O(1)
        \frac{ \inprod{ [\KB]_p }{ [\KAxi]_q } }{ N_\A N_\B d}
        \pm O\left( d^3 \left( \delta_{\AB, \perp} + \delta_{N/S} \right) \delta_{\xi, \perp} \delta_{N/S} \right)
      \right) \\
    &= \pm O(\delta_{\xi, \perp}). 
  \end{align*}
  Hence, 
  \[
    \Term_1\left( \inprod{ \ol{[\KAxi]_p} }{ \frac{\rd}{\rd t} \ol{[\KBxi]_q} } \right) 
    = O\left( \frac{\sigma_\xi^2 }{N_\A N_\B d } d \delta_{\xi, \perp}^2 \right). 
  \]
  Now we consider $\Term_2$. By Lemma~\ref{lemma: stage 2: estimations for Q2}, We have 
  \begin{align*}
    & \Term_2\left( \inprod{ \ol{[\KAxi]_p} }{ \frac{\rd}{\rd t} \ol{[\KBxi]_q} } \right) \\
    =\;&  
      \sum_{k=1}^{d-r}
      \ol{[\KAxi]_p}\trans 
      \left( \Id - \ol{ [\KBxi]_q } \left( \ol{ [\KBxi]_q } \right)\trans \right)
      \ol{[\KAxi]_k}
      \frac{ \norm{[\KAxi]_k} [\Q_2]_{q, k} }{ \norm{[\KBxi]_q} }
      \frac{\sigma_\xi^2 }{N_\A N_\B d } \\
    =\;&  
      \pm 
      \sum_{k=1}^{d-r}
      O(1) [\Q_2]_{q, k} 
      \frac{\sigma_\xi^2 }{N_\A N_\B d } \\
    =\;&  
      \pm O\left( 
        \frac{\sigma_{\max}^2 }{N_\A N_\B d } 
        d^4 \left( \delta_{\AB, \perp} + \delta_{N/S} \right) \delta_{\xi, \perp} \delta_{N/S} 
      \right).  \\
  \end{align*}
  Combine this together and we obtain 
  \[
    \inprod{ \ol{[\KAxi]_p} }{ \frac{\rd}{\rd t} \ol{[\KBxi]_q} }
    = \pm O\left( \frac{\sigma_\xi^2 }{N_\A N_\B d } d \delta_{\xi, \perp}^2 \right). 
  \]
  Similarly, one can derive the same bound for $\inprod{ \frac{\rd}{\rd t} \ol{[\KAxi]_p} }{ \ol{[\KBxi]_q} }$ and
  complete the proof. 
\end{proof}

\subsection{Maintaining $\mathbf{K_A} \approx \mathbf{K_B}$}
\label{sec: stage 2: KA approx KB}

In this subsection, we show that $\inprod{\ol{[\KA]_p}}{\ol{[\KB]_p}} \approx 1$, $\norm{[\KA]_p} 
\approx \norm{[\KB]_p}$, and also $\norm{[\KAxi]_q} \approx \norm{[\KBxi]_q}$ throughout Stage~2. 

\begin{lemma} 
  \label{lemma: stage 2: difference in angle}
  In Stage~2, we have 
  \[
    \frac{\rd}{\rd t} \inprod{\ol{[\KA]_p}}{\ol{[\KB]_p}}
    = \Omega\left(
        \frac{\sigma_p^2 [\Q_1]_{p, p}}{N_\A N_\B d } 
      \right)
      \left( 
        1 - \inprod{ \ol{ [\KA]_p } }{ \ol{[\KB]_p} }
      \right)
      \pm O\left(
        \frac{\sigma_{\max}^2 }{N_\A N_\B d } 
        d \kappa_0
        \delta_{\AB, \perp}^2
      \right). 
  \]
\end{lemma}

\begin{lemma}
  \label{lemma: stage 2: difference in norm}
  Define $\rho_{\A/\B, p} := \norm{[\KA]_p}^2 / \norm{[\KB]_p}^2$ and $\rho_{\B/\A, p} 
  := \norm{[\KB]_p}^2 / \norm{[\KA]_p}^2$. In Stage~2, we have 
  \begin{align*}
    \frac{\rd}{\rd t} \left( \rho_{\A/\B, p} + \rho_{\B/\A, p} \right)
    &= 
      \frac{2 [\Q_1]_{p, p} \sigma_p^2 }{N_\A N_\B d } 
      \inprod{\ol{[\KA]_p}}{\ol{[\KB]_p}}
      \left(
        2 -  \rho_{\A/\B, p} -  \rho_{\B/\A, p}
      \right) 
      \\
      &\qquad
      \pm O\left( 
        \frac{ \sigma_{\max}^2 }{N_\A N_\B d} 
        \left( d \kappa_0 \delta_{\AB, \perp}^2 + \delta_-^2 \right)
      \right). 
  \end{align*}
\end{lemma}

\begin{lemma}
  \label{lemma: stage 2: difference in norm (noise)}
  For any $p, q \in [d - r]$, define $\rho_{\xi, \A/\A, p/q} = \norm{[\KAxi]_p}^2 / \norm{[\KAxi]_q}^2$ and 
  $\rho_{\xi, \A/\B, p/q} = \norm{[\KAxi]_p}^2 / \norm{[\KBxi]_q}^2$. In Stage~2, we have 
  \begin{align*}
    \frac{\rd}{\rd t} \rho_{\xi, \A/\A, p/q}
    &= \pm O\left( 
        \frac{\sigma_\xi^2}{ N_\A N_\B d}  
        d \delta_{\xi, \perp}^2
      \right), \\ 
    \frac{\rd}{\rd t} \left( \rho_{\xi, \A/\B, p/q} + \rho_{\xi, \B/\A, q/p} \right)
    &= \pm O\left( 
      \frac{\sigma_\xi^2}{ N_\A N_\B d}  
      \left( d \delta_{\xi, \perp}^2 + \delta_-^2 \right) 
    \right). 
  \end{align*}
\end{lemma}

\subsubsection*{Omitted proof of this subsection}

\begin{proof}[Proof of Lemma~\ref{lemma: stage 2: difference in angle}]
  First, we write 
  \begin{align*}
    \inprod{\frac{\rd}{\rd t} \ol{[\KA]_p}}{\ol{[\KB]_p}}
    &= 
      \ol{[\KB]_p}\trans
      \left( \Id - \ol{ [\KA]_p } \left( \ol{ [\KA]_p } \right)\trans  \right)
      \frac{[\KB \Q_1]_p }{ \norm{[\KA]_p} }
      \frac{\sigma_p^2 }{N_\A N_\B d } \\
      &\qquad
      + \ol{[\KB]_p}\trans
      \left( \Id - \ol{ [\KA]_p } \left( \ol{ [\KA]_p } \right)\trans  \right)
      \frac{[\KBxi \Q_{1, \xi_\B}]_p}{\norm{[\KA]_p} }
      \frac{\sigma_p^2 }{N_\A N_\B d } \\
    &=: \Term_1\left( \inprod{\frac{\rd}{\rd t} \ol{[\KA]_p}}{\ol{[\KB]_p}} \right)
      + \Term_2\left( \inprod{\frac{\rd}{\rd t} \ol{[\KA]_p}}{\ol{[\KB]_p}} \right). 
  \end{align*}
  For $\Term_1$, we compute 
  \begin{align*}
    \Term_1\left( \inprod{\frac{\rd}{\rd t} \ol{[\KA]_p}}{\ol{[\KB]_p}} \right)
    &= 
      \ol{[\KB]_p}\trans
      \left( \Id - \ol{ [\KA]_p } \left( \ol{ [\KA]_p } \right)\trans  \right)
      \ol{[\KB]_p}
      \frac{\norm{[\KB]_p} [\Q_1]_{p, p} }{ \norm{[\KA]_p} }
      \frac{\sigma_p^2 }{N_\A N_\B d } \\
      &\qquad
      + \sum_{k \ne p} 
      \ol{[\KB]_p}\trans
      \left( \Id - \ol{ [\KA]_p } \left( \ol{ [\KA]_p } \right)\trans  \right)
      \ol{[\KB]_k}
      \frac{\norm{[\KB]_k} [\Q_1]_{k, p} }{ \norm{[\KA]_p} }
      \frac{\sigma_p^2 }{N_\A N_\B d } \\
    &= 
      \left( 
        1 - \inprod{ \ol{ [\KA]_p } }{ \ol{[\KB]_p} }^2
      \right)
      \frac{\norm{[\KB]_p} [\Q_1]_{p, p} }{ \norm{[\KA]_p} }
      \frac{\sigma_p^2 }{N_\A N_\B d } \\
      &\qquad
      \pm O\left(
        \frac{\sigma_{\max}^2 }{N_\A N_\B d } 
        d \kappa_0
        \delta_{\AB, \perp}^2
      \right) . 
  \end{align*}
  For $\Term_2$, we compute 
  \begin{align*}
    \Term_2\left( \inprod{\frac{\rd}{\rd t} \ol{[\KA]_p}}{\ol{[\KB]_p}} \right)
    &= \sum_{k=1}^{d-r}
      \ol{[\KB]_p}\trans
      \left( \Id - \ol{ [\KA]_p } \left( \ol{ [\KA]_p } \right)\trans  \right)
      \frac{[\KBxi]_k [\Q_{1, \xi_\B}]_{k, p}}{\norm{[\KA]_p} }
      \frac{\sigma_p^2 }{N_\A N_\B d } \\
    &= \pm O\left(
        \frac{\sigma_{\max}^2 }{N_\A N_\B d } 
        d \delta_{N/S}^2 \delta_{\xi, \perp}^2
      \right). 
  \end{align*}
  Therefore, 
  \[
    \inprod{\frac{\rd}{\rd t} \ol{[\KA]_p}}{\ol{[\KB]_p}}
    = \left( 
        1 - \inprod{ \ol{ [\KA]_p } }{ \ol{[\KB]_p} }^2
      \right)
      \frac{\norm{[\KB]_p} [\Q_1]_{p, p} }{ \norm{[\KA]_p} }
      \frac{\sigma_p^2 }{N_\A N_\B d } 
      \pm O\left(
        \frac{\sigma_{\max}^2 }{N_\A N_\B d } 
        d \kappa_0
        \delta_{\AB, \perp}^2
      \right) . 
  \]
  Then, by symmetry, we have 
  \begin{align*}
    \frac{\rd}{\rd t} \inprod{\ol{[\KA]_p}}{\ol{[\KB]_p}}
    &= \left( 
        1 - \inprod{ \ol{ [\KA]_p } }{ \ol{[\KB]_p} }^2
      \right)
      [\Q_1]_{p, p}
      \left( 
        \frac{\norm{[\KB]_p}  }{ \norm{[\KA]_p} }
        + \frac{\norm{[\KA]_p}  }{ \norm{[\KB]_p} }
      \right)
      \frac{\sigma_p^2 }{N_\A N_\B d } \\ 
      &\qquad
      \pm O\left(
        \frac{\sigma_{\max}^2 }{N_\A N_\B d } 
        d \kappa_0
        \delta_{\AB, \perp}^2
      \right) \\ 
    &= \Omega\left(
        \frac{\sigma_p^2 [\Q_1]_{p, p}}{N_\A N_\B d } 
      \right)
      \left( 
        1 - \inprod{ \ol{ [\KA]_p } }{ \ol{[\KB]_p} }
      \right)
      \pm O\left(
        \frac{\sigma_{\max}^2 }{N_\A N_\B d } 
        d \kappa_0
        \delta_{\AB, \perp}^2
      \right). 
  \end{align*}
\end{proof}

\begin{proof}[Proof of Lemma~\ref{lemma: stage 2: difference in norm}]
  Similar to Stage~1, we define $\rho_{\A/\B, p} = \norm{[\KA]_p}^2 / \norm{[\KB]_p}^2$. 
  By Lemma~\ref{lemma: d norm and d bar}, we have
  \begin{align*}
    \frac{\rd}{\rd t} \rho_{\A/\B, p}
    &= \frac{ \frac{\rd}{\rd t} \norm{[\KA]_p}^2 }{ \norm{[\KB]_p}^2 }
      - \rho_{\A/\B, p} \frac{ \frac{\rd}{\rd t} \norm{[\KB]_p}^2 }{ \norm{[\KB]_p}^2 } \\
    &= 
      2 \frac{\inprod{[\KA]_p}{[\KB\Q_1]_p}}{N_\A N_\B d \norm{[\KB]_p}^2} \sigma_p^2
      + 2 \frac{\inprod{[\KA]_p}{[\KBxi\Q_{1, \xi_\B}]_p}}{N_\A N_\B d \norm{[\KB]_p}^2}  \sigma_p^2
      + 2 \frac{ \rho_{\A/\B, p} }{N_\A^2 d} Q_0 \sigma_p^2 \\
      &\qquad
      - \rho_{\A/\B, p} \left(
        2 \frac{ \inprod{[\KB]_p}{[\KA\Q_1\trans]_p}}{N_\A N_\B d \norm{[\KB]_p}^2} \sigma_p^2
        + 2 \frac{ \inprod{[\KB]_p}{[\KAxi\Q_{1, \xi_\A}]_p}}{N_\A N_\B d \norm{[\KB]_p}^2}  
          \sigma_p^2
        + 2 \frac{1}{N_\B^2 d} Q_0 \sigma_p^2
      \right) \\
    &= 
      2 \frac{\inprod{[\KA]_p}{[\KB\Q_1]_p}}{N_\A N_\B d \norm{[\KB]_p}^2} \sigma_p^2
      + 2 \frac{\inprod{[\KA]_p}{[\KBxi\Q_{1, \xi_\B}]_p}}{N_\A N_\B d \norm{[\KB]_p}^2}  \sigma_p^2
        \\
      &\qquad
      - \rho_{\A/\B, p} \left(
        2 \frac{ \inprod{[\KB]_p}{[\KA\Q_1\trans]_p}}{N_\A N_\B d \norm{[\KB]_p}^2} \sigma_p^2
        + 2 \frac{ \inprod{[\KB]_p}{[\KAxi\Q_{1, \xi_\A}]_p}}{N_\A N_\B d \norm{[\KB]_p}^2}  
          \sigma_p^2
      \right) \\
      &\qquad
      + 2 Q_0 \sigma_p^2 
        \left(
          \frac{1}{N_\A^2 d} - \frac{1}{N_\B^2 d}
        \right) 
        \rho_{\A/\B, p}. 
  \end{align*}
  Then, by Lemma~\ref{lemma: stage 2: estimations for Q1} and Lemma~\ref{lemma: stage 2: estimations for Q1xi},
  we have 
  \begin{align*}
    \inprod{[\KA]_p}{[\KB\Q_1]_p}
    &= \inprod{[\KA]_p}{[\KB]_p} [\Q_1]_{p, p}
      \pm O\left( d \kappa_p^2 \kappa_0 \delta_{\AB, \perp}^2 \right), \\
    \inprod{[\KB]_p}{[\KA\Q_1\trans]_p}
    &= \inprod{[\KA]_p}{[\KB]_p} [\Q_1]_{p, p}
      \pm O\left( d \kappa_p^2 \kappa_0 \delta_{\AB, \perp}^2 \right), \\
    \inprod{[\KA]_p}{[\KBxi \Q_{1, \bxi_\B}]_p}
    &= \pm O\left( d \kappa_p^2 \delta_{N/S}^2 \delta_{\xi, \perp} \right), \\
    \inprod{[\KB]_p}{[\KAxi \Q_{1, \bxi_\A}]_p}
    &= \pm O\left( d \kappa_p^2 \delta_{N/S}^2 \delta_{\xi, \perp} \right). 
  \end{align*}
  Thus, 
  \begin{align*}
    \frac{\rd}{\rd t} \rho_{\A/\B, p}
    &= 
      2 \frac{ [\Q_1]_{p, p} \sigma_p^2 }{N_\A N_\B d } 
      \frac{ \inprod{[\KA]_p}{[\KB]_p} }{ \norm{[\KB]_p}^2 }
      \left( 1 -  \rho_{\A/\B, p}  \right) \\
      &\qquad
      + 2 Q_0 \sigma_p^2 
      \left(
        \frac{1}{N_\A^2 d} - \frac{1}{N_\B^2 d}
      \right) 
      \rho_{\A/\B, p}
      \pm O\left( 
        \frac{ \sigma_{\max}^2 }{N_\A N_\B d} 
        d \kappa_0 \delta_{\AB, \perp}^2 
      \right). 
  \end{align*}
  Interchange the roles of $\A, \B$ and we get 
  \begin{align*}
    \frac{\rd}{\rd t} \rho_{\B/\A, p}
    &= 
      2 \frac{ [\Q_1]_{p, p} \sigma_p^2 }{N_\A N_\B d } 
      \frac{ \inprod{[\KA]_p}{[\KB]_p} }{ \norm{[\KA]_p}^2 }
      \left( 1 -  \rho_{\B/\A, p}  \right) \\
      &\qquad
      + 2 Q_0 \sigma_p^2 
      \left(
        \frac{1}{N_\B^2 d} - \frac{1}{N_\A^2 d}
      \right) 
      \rho_{\B/\A, p}
      \pm O\left( 
        \frac{ \sigma_{\max}^2 }{N_\A N_\B d} 
        d \kappa_0 \delta_{\AB, \perp}^2 
      \right). 
  \end{align*}
  Hence, 
  \begin{align*}
    \frac{\rd}{\rd t} \left( \rho_{\A/\B, p} + \rho_{\B/\A, p} \right)
    &= 
      2 \frac{ [\Q_1]_{p, p} \sigma_p^2 }{N_\A N_\B d } 
      \inprod{[\KA]_p}{[\KB]_p}
      \left(
        \frac{1 -  \rho_{\A/\B, p}}{ \norm{[\KB]_p}^2 }
        + \frac{1 -  \rho_{\B/\A, p}}{ \norm{[\KA]_p}^2 }
      \right) 
      \\
      &\qquad
      + 2 Q_0 \sigma_p^2 
        \left(
          \frac{1}{N_\A^2 d} - \frac{1}{N_\B^2 d}
        \right) 
        \left(
          \rho_{\A/\B, p}
          - \rho_{\B/\A, p}
        \right)
      \pm O\left( 
        \frac{ \sigma_{\max}^2 }{N_\A N_\B d} 
        d \kappa_0 \delta_{\AB, \perp}^2 
      \right) \\
    &= 
      \frac{2 [\Q_1]_{p, p} \sigma_p^2 }{N_\A N_\B d } 
      \inprod{\ol{[\KA]_p}}{\ol{[\KB]_p}}
      \left(
        2 -  \rho_{\A/\B, p} -  \rho_{\B/\A, p}
      \right) 
      \\
      &\qquad
      \pm O\left( 
        \frac{ \sigma_{\max}^2 }{N_\A N_\B d} 
        \left( d \kappa_0 \delta_{\AB, \perp}^2 + \delta_-^2 \right)
      \right). 
  \end{align*}
\end{proof}

\begin{proof}[Proof of Lemma~\ref{lemma: stage 2: difference in norm (noise)}]
  For notational simplicity, we drop the subscript $\xi$ in the proof. 
  Recall from Corollary~\ref{cor: stage 2: d norm} that 
  \[ 
    \frac{\rd}{\rd t} \norm{[\KAxi]_q}^2
    = \frac{2 \norm{[\KAxi]_q}^2}{N_\A^2 d} Q_0 \sigma_\xi^2
      \pm O\left( 
        \frac{\sigma_\xi^2  \norm{[\KAxi]_q}^2}{ N_\A N_\B d}  
        d \delta_{\xi, \perp}^2
      \right). 
  \] 
  Hence, for any $p, q \in [d - r]$, we have 
  \begin{align*}
    \frac{\rd}{\rd t} \rho_{\A/\A, p/q}
    &= \frac{\frac{\rd}{\rd t} \norm{[\KAxi]_p}^2}{\norm{[\KAxi]_q}^2}
      - \rho_{\A/\A, p/q} \frac{\frac{\rd}{\rd t} \norm{[\KAxi]_q}^2}{\norm{[\KAxi]_q}^2} \\
    &= \frac{2}{N_\A^2 d} Q_0 \sigma_\xi^2 \rho_{\A/\A, p/q}
      \pm O\left( 
        \frac{\sigma_\xi^2}{ N_\A N_\B d}  
        d \delta_{\xi, \perp}^2
      \right) \\ 
      &\qquad
      - \rho_{\A/\A, p/q} \left(
        \frac{2}{N_\A^2 d} Q_0 \sigma_\xi^2
        \pm O\left( 
          \frac{\sigma_\xi^2}{ N_\A N_\B d}  
          d \delta_{\xi, \perp}^2
        \right)
      \right) \\ 
    &= \pm O\left( 
        \frac{\sigma_\xi^2}{ N_\A N_\B d}  
        d \delta_{\xi, \perp}^2
      \right).
  \end{align*}
  Similarly, we have 
  \begin{align*}
    \frac{\rd}{\rd t} \rho_{\A/\B, p/q}
    &= \frac{\frac{\rd}{\rd t} \norm{[\KAxi]_p}^2}{\norm{[\KBxi]_q}^2}
      - \rho_{\A/\B, p/q} \frac{\frac{\rd}{\rd t} \norm{[\KBxi]_q}^2}{\norm{[\KBxi]_q}^2} \\
    &= \frac{2}{N_\A^2 d} Q_0 \sigma_\xi^2 \rho_{\A/\B, p/q}
      \pm O\left( 
        \frac{\sigma_\xi^2}{ N_\A N_\B d}  
        d \delta_{\xi, \perp}^2
      \right) \\ 
      &\qquad
      - \rho_{\A/\B, p/q} \left(
        \frac{2}{N_\B^2 d} Q_0 \sigma_\xi^2
        \pm O\left( 
          \frac{\sigma_\xi^2}{ N_\A N_\B d}  
          d \delta_{\xi, \perp}^2
        \right)
      \right) \\ 
    &= \left( \frac{2}{N_\A^2 d} - \frac{2}{N_\B^2 d} \right) Q_0 \sigma_\xi^2 \rho_{\A/\B, p/q}
      \pm O\left( 
        \frac{\sigma_\xi^2}{ N_\A N_\B d}  
        d \delta_{\xi, \perp}^2
      \right). 
  \end{align*}
  By symmetry, we also have 
  \[
    \frac{\rd}{\rd t} \rho_{\B/\A, q/p}
    = \left( \frac{2}{N_\B^2 d} - \frac{2}{N_\A^2 d} \right) Q_0 \sigma_\xi^2 \rho_{\B/\A, q/p}
      \pm O\left( 
        \frac{\sigma_\xi^2}{ N_\A N_\B d}  
        d \delta_{\xi, \perp}^2
      \right).
  \]
  Hence, 
  \begin{align*}
    \frac{\rd}{\rd t} \left( \rho_{\A/\B, p/q} + \rho_{\B/\A, q/p} \right)
    &= \left( \frac{2}{N_\A^2 d} - \frac{2}{N_\B^2 d} \right) Q_0 \sigma_\xi^2 
      \left( \rho_{\A/\B, p/q} - \rho_{\B/\A, q/p} \right) 
      \pm O\left( 
        \frac{\sigma_\xi^2}{ N_\A N_\B d}  
        d \delta_{\xi, \perp}^2
      \right) \\
    &= \pm O\left( 
        \frac{\sigma_\xi^2}{ N_\A N_\B d}  
        \left( d \delta_{\xi, \perp}^2 + \delta_-^2 \right) 
      \right). 
  \end{align*}
\end{proof}

\subsection{Controlling the noise-signal ratio}
\label{sec: stage 2: noise-signal ratio}

In this subsection, we show that the noise-signal ratio remains small throughout Stage~2.

\begin{lemma}
  \label{lemma: stage 2: noise signal ratio}
  Let $\norm{[\KA]_p}$ be the smallest among all $\braces{ \norm{[\KA]_k} }_{k \in [r]}$. For any 
  $q \in [r]$, in Stage~2, we have 
  \[ 
    \frac{\rd}{\rd t} \frac{\norm{[\KAxi]_q}^2}{\norm{[\KA]_p}^2}
    \le O\left( 
        \frac{\sigma_\xi^2 }{ N_\A N_\B d}  
        \left( d \delta_{\xi, \perp}^2 + d^2 \delta_{\AB, \perp}^2 + \delta_- + d^2 \delta_{N/S} \delta_{\xi, \perp} \right) 
      \right)
      \frac{\norm{[\KAxi]_q}^2}{\norm{[\KA]_p}^2}. 
  \] 
\end{lemma}
\begin{proof}
  Recall from Corollary~\ref{cor: stage 2: d norm} that 
  \begin{align*}
    \frac{\rd}{\rd t} \norm{[\KA]_p}^2
    &= \frac{2 \sigma_p^2 \norm{[\KA]_p} \norm{[\KB]_p}}{N_\A N_\B d} 
      \inprod{\ol{[\KA]_p}}{\ol{[\KB]_p}} [\Q_1]_{p, p} 
      + 2 \frac{\norm{[\KA]_p}^2}{N_\A^2 d} Q_0 \sigma_p^2 \\ 
      &\qquad
      \pm O\left(  
        \frac{\sigma_p^2 \kappa_p^2}{N_\A N_\B d} 
        \kappa_0 d
        \delta_{\AB, \perp}^2 
      \right) \\ 
    \frac{\rd}{\rd t} \norm{[\KAxi]_q}^2
    &= \frac{2 \norm{[\KAxi]_q}^2}{N_\A^2 d} Q_0 \sigma_\xi^2
      \pm O\left( 
        \frac{\sigma_\xi^2  \norm{[\KAxi]_q}^2}{ N_\A N_\B d}  
        d \delta_{\xi, \perp}^2
      \right). 
  \end{align*}
  Since the condition number of $\KA$ is bounded by $\sqrt{d}$, it suffices to consider the smallest 
  $\norm{[\KA]_p}$, for which we have 
  \begin{align*}
    \frac{\rd}{\rd t} \norm{[\KA]_p}^2
    &= \frac{2 \sigma_p^2 }{N_\A N_\B d} 
      [\Q_1]_{p, p} 
      \norm{[\KA]_p}^2
      + \frac{2 \sigma_p^2}{N_\A^2 d} Q_0 \norm{[\KA]_p}^2\\ 
      &\qquad
      \pm O\left(  
        \frac{\sigma_p^2 \kappa_p^2}{N_\A N_\B d} 
        \left( \kappa_0 d \delta_{\AB, \perp}^2 + \delta_- \right) 
      \right) \\ 
    &= \frac{2 \sigma_p^2 }{N_\A N_\B d} 
      [\Q_1]_{p, p} 
      \norm{[\KA]_p}^2
      - \frac{2 \sigma_p^2}{N_\A^2 d} 
        \left( \sum_{k=1}^r \frac{\kappa_k^2}{\norm{\bkappa}^2} [\Q_1]_{k, k} \right)
        \norm{[\KA]_p}^2\\ 
      &\qquad
      \pm O\left(  
        \frac{\sigma_p^2 \kappa_p^2}{N_\A N_\B d} 
        \left( d^2 \delta_{\AB, \perp}^2 + d^2 \delta_{N/S} \delta_{\xi, \perp} + \delta_- \right) 
      \right), 
  \end{align*}
  where the last line comes from Lemma~\ref{lemma: stage 2: estimations for Q0}. By Lemma~\ref{lemma: stage 1: 
  estimations for Q1}, $[\Q_1]_{p, p}$ is negative correlated with $\kappa_p^2$. As a result, we have 
  \[
    \frac{\rd}{\rd t} \norm{[\KA]_p}^2
    \ge - O\left(  
        \frac{\sigma_p^2 \kappa_p^2}{N_\A N_\B d} 
        \left( d^2 \delta_{\AB, \perp}^2 + d^2 \delta_{N/S} \delta_{\xi, \perp} + \delta_- \right) 
      \right). 
  \]
  For $\KAxi$, we simply have 
  \begin{align*}
    \frac{\rd}{\rd t} \norm{[\KAxi]_q}^2
    \le O\left( 
        \frac{\sigma_\xi^2 }{ N_\A N_\B d}  
        \left( d \delta_{\xi, \perp}^2 + d^2 \delta_{\AB, \perp}^2 + \delta_- + d^2 \delta_{N/S} \delta_{\xi, \perp} \right) 
      \right)
      \norm{[\KAxi]_q}^2. 
  \end{align*}
  Thus, 
  \begin{align*}
    \frac{\rd}{\rd t} \frac{\norm{[\KAxi]_q}^2}{\norm{[\KA]_p}^2}
    &= \frac{\frac{\rd}{\rd t} \norm{[\KAxi]_q}^2}{\norm{[\KA]_p}^2}
      - \frac{\norm{[\KAxi]_q}^2}{\norm{[\KA]_p}^2} \frac{\frac{\rd}{\rd t} \norm{[\KA]_p}^2}{\norm{[\KA]_p}^2} \\
    &\le O\left( 
        \frac{\sigma_\xi^2 }{ N_\A N_\B d}  
        \left( d \delta_{\xi, \perp}^2 + d^2 \delta_{\AB, \perp}^2 + \delta_- + d^2 \delta_{N/S} \delta_{\xi, \perp} \right) 
      \right)
      \frac{\norm{[\KAxi]_q}^2}{\norm{[\KA]_p}^2}. 
  \end{align*}
\end{proof}

\subsection{Estimating the convergence rate}
\label{sec: stage 2: convergence rate}

In this subsection, we estimate how fast the condition number will become close to $1$. 

\begin{lemma}
  \label{lemma: stage 2: condition number}
  Suppose that $\norm{[\KA]_p}$ is the largest and $\norm{[\KA]_q}$ the smallest among all 
  $\braces{ \norm{[\KA]_k} }_{k \in [r]}$. In Stage~2, we have 
  \begin{align*}
    \frac{\rd}{\rd t} \frac{\norm{[\KA]_p}^2}{\norm{[\KA]_q}^2}
    &\le - \frac{4 (1 - \tilde{S}) \sigma_{\min}^2}{N_\A N_\B d} 
      \left( T_p - T_q \right) 
      \frac{\norm{[\KA]_p}^2}{\norm{[\KA]_q}^2} \\
      &\qquad
      \pm O\left(  
        \frac{\sigma_p^2}{N_\A N_\B d} 
        \left( d^2 \delta_{\AB, \perp}^2 + \delta_- + d^2 \delta_{N/S} \delta_{\xi, \perp} \right) 
        \kappa_0^2 
      \right). 
  \end{align*}
\end{lemma}

\begin{corollary}[Convergence rate]
  \label{cor: stage 2: convergence rate}
  Suppose that $\norm{[\KA]_p}$ is the largest and $\norm{[\KA]_q}$ the smallest among all 
  $\braces{ \norm{[\KA]_k} }_{k \in [r]}$. For any constant $c > 1$, it takes at most $\poly(d)$ amount of 
  time for $\norm{[\KA]_p}^2 / \norm{[\KA]_q}^2$ to become smaller than $c$. 
\end{corollary}

\subsubsection*{Omitted proof of this subsection}

\begin{proof}[Proof of Lemma~\ref{lemma: stage 2: condition number}]
  By Corollary~\ref{cor: stage 2: d norm}, Lemma~\ref{lemma: stage 2: estimations for Q0} and 
  Lemma~\ref{lemma: stage 2: estimations for Q1}, we have 
  \begin{align*}
    \frac{\rd}{\rd t} \norm{[\KA]_p}^2
    &= \frac{2 \sigma_p^2  }{N_\A N_\B d} [\Q_1]_{p, p} \norm{[\KA]_p}^2
      + \frac{2 \sigma_p^2 }{N_\A^2 d} Q_0 \norm{[\KA]_p}^2 \\ 
      &\qquad
      \pm O\left(  
        \frac{\sigma_p^2 \kappa_p^2}{N_\A N_\B d} 
        \left( \kappa_0 d \delta_{\AB, \perp}^2 + \delta_- \right) 
      \right) \\
    &= \frac{4 (1 - \tilde{S}) \sigma_p^2  }{N_\A N_\B d} (1 - T_p)  \norm{[\KA]_p}^2
      + \frac{4 (1 - \tilde{S}) \sigma_p^2 }{N_\A^2 d} 
        \left( \sum_{k=1}^r \frac{\kappa_k^2}{\norm{\bkappa}^2} (1 - T_p) \right)
        \norm{[\KA]_p}^2 \\ 
      &\qquad
      \pm O\left(  
        \frac{\sigma_p^2 \kappa_p^2}{N_\A N_\B d} 
        \left( d^2 \delta_{\AB, \perp}^2 + \delta_- + d^2 \delta_{N/S} \delta_{\xi, \perp} \right) 
      \right) \\
    &= - \frac{4 (1 - \tilde{S}) \sigma_p^2  }{N_\A N_\B d} \left( T_p - \tilde{T} \right)  \norm{[\KA]_p}^2 
      \pm O\left(  
        \frac{\sigma_p^2 \kappa_p^2}{N_\A N_\B d} 
        \left( d^2 \delta_{\AB, \perp}^2 + \delta_- + d^2 \delta_{N/S} \delta_{\xi, \perp} \right) 
      \right), 
  \end{align*}
  where $\tilde{T} = \sum_{k=1}^r \frac{\kappa_k^2}{\norm{\bkappa}^2} T_p $. Then, we compute 
  \begin{align*}
    \frac{\rd}{\rd t} \frac{\norm{[\KA]_p}^2}{\norm{[\KA]_q}^2}
    &= \frac{\frac{\rd}{\rd t} \norm{[\KA]_p}^2}{\norm{[\KA]_q}^2}
      - \frac{\norm{[\KA]_p}^2}{\norm{[\KA]_q}^2} 
      \frac{\frac{\rd}{\rd t} \norm{[\KA]_q}^2}{\norm{[\KA]_q}^2} \\
    &= - \frac{4 (1 - \tilde{S}) }{N_\A N_\B d} 
      \left( 
        \sigma_p^2 \left( T_p - \tilde{T} \right)  
        - \sigma_q^2 \left( T_q - \tilde{T} \right)  
      \right) 
      \frac{\norm{[\KA]_p}^2}{\norm{[\KA]_q}^2} \\
      &\qquad
      \pm O\left(  
        \frac{\sigma_p^2}{N_\A N_\B d} 
        \kappa_0^2 \left( d^2 \delta_{\AB, \perp}^2 + \delta_- + d^2 \delta_{N/S} \delta_{\xi, \perp} \right) 
      \right). 
  \end{align*}
  Since $\norm{[\KA]_p}$ is the largest, $\norm{[\KA]_q}$ is the smallest, $T_p$ is positively correlated with 
  $\norm{[\KA]_p}$, and $\tilde{T}$ is a weighted average of $T_p$, we have 
  \[
    \sigma_p^2 \left( T_p - \tilde{T} \right) - \sigma_q^2 \left( T_q - \tilde{T} \right)  
    \ge \sigma_q^2 \left( T_p - T_q \right) .
  \]
  Thus, 
  \begin{align*}
    \frac{\rd}{\rd t} \frac{\norm{[\KA]_p}^2}{\norm{[\KA]_q}^2}
    &\le - \frac{4 (1 - \tilde{S}) \sigma_{\min}^2}{N_\A N_\B d} 
      \left( T_p - T_q \right) 
      \frac{\norm{[\KA]_p}^2}{\norm{[\KA]_q}^2} \\
      &\qquad
      \pm O\left(  
        \frac{\sigma_p^2}{N_\A N_\B d} 
        \left( d^2 \delta_{\AB, \perp}^2 + \delta_- + d^2 \delta_{N/S} \delta_{\xi, \perp} \right) 
        \kappa_0^2 
      \right). 
  \end{align*}
\end{proof}

\begin{proof}[Proof of Corollary~\ref{cor: stage 2: convergence rate}]
  Recall that 
  \[ 
    T_p 
    = \tanh\left( \frac{\hat\kappa_p^2}{N_\A N_\B d} \right)
    = \tanh\left( \frac{\norm{[\KA]_p}^2}{\norm{\KA}_F^2} \left( 1 \pm O( \delta_- + \delta_{N/S} ) \right) \right). 
  \] 
  Since $\kappa_0 \le \sqrt{d}$, we have $\norm{[\KA]_p}^2 / \norm{[\KA]_F}^2 \le 1/2$. Note that 
  $\tanh'(z) = 1 - \tanh^2(z) = \Omega(1)$ for any $z \le 1.1/2$. Therefore,  
  \[ 
    T_p - T_q
    \ge \Omega\left(  
          \frac{ \norm{[\KA]_p}^2 - \norm{[\KA]_q}^2}{\norm{\KA}_F^2} 
        \right)
        \pm O( \delta_- + \delta_{N/S} ) 
    \ge \Omega\left( \frac{1}{d} \right). 
  \] 
  Then, by Lemma~\ref{lemma: stage 2: condition number}, we have 
  \begin{align*}
    \frac{\rd}{\rd t} \frac{\norm{[\KA]_p}^2}{\norm{[\KA]_q}^2}
    &\le - \frac{4 (1 - \tilde{S}) \sigma_{\min}^2}{N_\A N_\B d} 
      \left( T_p - T_q \right) 
      \frac{\norm{[\KA]_p}^2}{\norm{[\KA]_q}^2} \\
      &\qquad
      \pm O\left(  
        \frac{\sigma_p^2}{N_\A N_\B d} 
        \left( d^2 \delta_{\AB, \perp}^2 + \delta_- + d^2 \delta_{N/S} \delta_{\xi, \perp} \right) 
        \kappa_0^2 
      \right) \\ 
    &\le - \Omega\left( \frac{ \sigma_{\min}^2}{N_\A N_\B d} \right) 
        \frac{\norm{[\KA]_p}^2}{\norm{[\KA]_q}^2}. 
  \end{align*}
  By the proof of Lemma~\ref{lemma: stage 2: condition number}, the largest $\norm{[\KA]_p}^2$ is non-increasing. 
  Hence, $N_\A N_\B d$ is upper bounded by some $\poly(d)$. Thus, it takes at most $\poly(d)$ for 
  $ \norm{[\KA]_p^2} / \norm{[\KA]_q} $ to become smaller than $c$. 
\end{proof}

\subsection{Proof of the main lemma of Stage 2}
\label{sec: stage 2: proof of the main lemma}

\begin{proof}[Proof of Lemma~\ref{lemma: stage 2: main}]
  The polynomial bound on the convergence time has been proved in Corollary~\ref{cor: stage 2: convergence rate}. 
  For the errors, recall from Lemma~\ref{lemma: stage 2: orthogonality between signals}, 
  Lemma~\ref{lemma: stage 2: orthogonality between signals and noises} and Lemma~\ref{lemma: stage 2: orthogonality 
  between noises} that 
  \begin{align*}
    \frac{\rd}{\rd t} \hat{\delta}_{\perp, p, q}
    &\le  
      O\left(
        \frac{
          \sigma_{\max}^2 
        }{N_\A N_\B d}  
        \kappa_0
        \left( d^2 \delta_{\AB, \perp}^2 + \delta_{\AB, \perp} \sqrt{\delta_-} + \delta_- \right)
        \sqrt{\hat{\delta}_{\perp, p, q}}
      \right), \\
    \frac{\rd}{\rd t} \hat{\delta}_{\xi, \perp, p, q}
    &\le O\left(
        \frac{\sigma_{\max}^2 }{N_\A N_\B d } 
        d \kappa_0 
        \delta_{\xi, \perp}^2 \left( \delta_{\xi, \perp} +  \sqrt{\delta_-} \right)
      \right), \\
    \frac{\rd}{\rd t} \inprod{ \ol{[\KAxi]_p} }{ \ol{[\KBxi]_q} }
    &= \pm O\left( \frac{\sigma_\xi^2 }{N_\A N_\B d } d \delta_{\xi, \perp}^2 \right). 
  \end{align*}
  Recall from Lemma~\ref{lemma: stage 2: difference in angle}, Lemma~\ref{lemma: stage 2: difference in norm},
  and Lemma~\ref{lemma: stage 2: difference in norm (noise)} that
  \begin{equation}
    \label{eq: stage 2: main lemma proof: delta-}
    \begin{aligned}
      \frac{\rd}{\rd t} \inprod{\ol{[\KA]_p}}{\ol{[\KB]_p}}
      &= \Omega\left(
          \frac{\sigma_p^2 [\Q_1]_{p, p}}{N_\A N_\B d } 
        \right)
        \left( 
          1 - \inprod{ \ol{ [\KA]_p } }{ \ol{[\KB]_p} }
        \right)
        \pm O\left(
          \frac{\sigma_{\max}^2 }{N_\A N_\B d } 
          d \kappa_0
          \delta_{\AB, \perp}^2
        \right), \\ 
      \frac{\rd}{\rd t} \left( \rho_{\A/\B, p} + \rho_{\B/\A, p} \right)
      &\le O\left( 
          \frac{ \sigma_{\max}^2 }{N_\A N_\B d} 
          \left( d \kappa_0 \delta_{\AB, \perp}^2 + \delta_-^2 \right)
        \right), \\ 
      \frac{\rd}{\rd t} \rho_{\xi, \A/\A, p/q}
      &\le O\left( 
          \frac{\sigma_\xi^2}{ N_\A N_\B d}  
          d \delta_{\xi, \perp}^2
        \right), \\ 
      \frac{\rd}{\rd t} \left( \rho_{\xi, \A/\B, p/q} + \rho_{\xi, \B/\A, q/p} \right)
      &\le O\left( 
        \frac{\sigma_\xi^2}{ N_\A N_\B d}  
        \left( d \delta_{\xi, \perp}^2 + \delta_-^2 \right) 
      \right).  
    \end{aligned}
  \end{equation}
  By Lemma~\ref{lemma: stage 2: noise signal ratio}, we have 
  \[ 
    \frac{\rd}{\rd t} \frac{\norm{[\KAxi]_q}^2}{\norm{[\KA]_p}^2}
    \le O\left( 
        \frac{\sigma_\xi^2 }{ N_\A N_\B d}  
        \left( d \delta_{\xi, \perp}^2 + d^2 \delta_{\AB, \perp}^2 + \delta_- + d^2 \delta_{N/S} \delta_{\xi, \perp} \right) 
      \right)
      \frac{\norm{[\KAxi]_q}^2}{\norm{[\KA]_p}^2}. 
  \] 
  Note that on the RHS of these equations, the only terms whose order may potentially be smaller than or equal to LHS
  are the $\delta_-$-reltaed terms. However, \eqref{eq: stage 2: main lemma proof: delta-}, we can make sure 
  $\delta_-$ is at most $\delta_{\AB, \perp}^{1.5}$. As a result, the orders of the RHS are all greater than the 
  orders of the LHS, which implies that these errors can at most double within $\poly(d)$ time if they are
  sufficiently small at the beginning of Stage~2. 
\end{proof}

\section{From Gradient Flow to Gradient Descent}
\label{sec: gf to gd}

Converting the above gradient flow argument to a gradient descent one is standard. All our 
estimations can tolerate an inverse polynomially large error. Since the all quantities of interest here are 
polynomially and inverse polynomially bounded, at each step of gradient descent, one can always make the 
GF-to-GD discretization error sufficiently (inverse polynomially) small by choose a sufficiently 
(inverse polynomially) small learning rate and generating sufficiently (polynomially) many samples. Since the 
times need for Stage~1 and Stage~2 are both polynomial, this also implies a polynomial sample complexity.

\end{document}